\documentclass[10pt,twocolumn,letterpaper]{article}

\usepackage{iccv}              %

\usepackage{comment}
\usepackage{bbm}
\usepackage[percent]{overpic}
\usepackage{float}

\usepackage{graphicx,multirow,array}
\usepackage{tikz}
\usetikzlibrary{decorations.fractals,spy}

\newcommand{\spyimg}[4]{%
	\begin{tikzpicture}[spy using outlines={yellow,magnification=3,size=0.95cm, connect spies}]
		\node[anchor=south west,inner sep=0] at (0,0) {\includegraphics[width=#1]{#2}};
		\spy on (#3) in node [left] at (#4);
	\end{tikzpicture}%
}

\newcommand{\gtimage}{I}
\newcommand{\predimage}{\hat{I}}
\newcommand{\visibility}{V}
\newcommand{\denseb}{B}
\newcommand{\densei}{T}
\newcommand{\densed}{D}

\newcommand{\colorfunc}{C}
\newcommand{\gsopacity}{o}

\newcommand{\embedding}{\ell}
\newcommand{\imageembeddingsupers}{(\mathcal{I})}
\newcommand{\gsembeddingsupers}{(\mathcal{G})}
\newcommand{\imageembedding}[0]{\embedding^{\imageembeddingsupers}}
\newcommand{\gsembedding}[0]{\embedding^{\gsembeddingsupers}}

\newcommand{\tabmetrichead}[0]{SSIM & PSNR & LPIPS & \#G & FPS}
\newcommand{\tabmetricheadwithoutfps}[0]{SSIM & PSNR & LPIPS & \#G}

\newcommand{\tabletitle}[1]{\textbf{#1}}
\newcommand{\ourcampusname}{\textit{JNU-ZH}}

\definecolor{iccvblue}{rgb}{0.21,0.49,0.74}
\usepackage[pagebackref,breaklinks,colorlinks,allcolors=iccvblue]{hyperref}

\def\paperID{16104} %
\def\confName{ICCV}
\def\confYear{2025}

\title{Robust and Efficient 3D Gaussian Splatting for Urban Scene Reconstruction}

\def\inst#1{\unskip$^{#1}$}

\author{Zhensheng Yuan\inst{1,2}
\quad
Haozhi Huang\inst{1}
\quad
Zhen Xiong\inst{1}
\quad
Di Wang\inst{1}\footnotemark[1]\thanks{Corresponding author}
\quad
Guanghua Yang\inst{1}\footnotemark[1] \\
{$^1$Jinan University
\quad
$^2$University of Macau} \\
{\small\texttt{
zhensheng@stu2022.jnu.edu.cn, hzhuang@jnu.edu.cn, acxz2000@stu2022.jnu.edu.cn,}} \\
{\small\texttt{diwang@jnu.edu.cn, ghyang@jnu.edu.cn}}}

\begin{document}
\maketitle

\begin{abstract}
We present a 
framework that enables fast reconstruction and real-time rendering of urban-scale scenes while maintaining robustness against appearance variations across multi-view captures.
Our approach begins with scene partitioning for parallel training, employing a visibility-based image selection strategy to optimize training efficiency.
\textbf{A controllable level-of-detail (LOD) strategy explicitly regulates Gaussian density under a user-defined budget, enabling efficient training and rendering while maintaining high visual fidelity.}
The appearance transformation module mitigates the negative effects
of appearance inconsistencies across images while enabling flexible adjustments.
Additionally, we utilize enhancement modules, such as depth regularization, scale regularization, and anti-aliasing, to improve reconstruction fidelity.
Experimental results demonstrate that our method effectively reconstructs urban-scale scenes and outperforms previous approaches in both efficiency and quality.
The source code is available at: \url{https://yzslab.github.io/REUrbanGS}.

\end{abstract}
    
\section{Introduction}
\label{sec:intro}

Urban-scale scene reconstruction plays a crucial role in various fields, such as autonomous driving \cite{li2019aads, ost2021neural}, urban planning \cite{musialski2013survey}, and digital twins \cite{deng2021systematic}, making it a highly active research area. %
The recently emerged 3D Gaussian Splatting (3DGS) \cite{kerbl20233d} represents scenes with explicit 3D elliptical Gaussians, enabling high-fidelity, photorealistic reconstruction from image collections while supporting real-time rendering. %

Compared to traditional MVS methods \cite{schonberger2016pixelwise} relying on depth estimation and mesh reconstruction, 3DGS optimizes explicit 3D Gaussians for higher rendering fidelity. %
Unlike implicit neural representations \cite{ost2021neural, mildenhall2021nerf} that require costly volume rendering, 3DGS leverages efficient rasterization for fast reconstruction and real-time rendering. %
However, its explicit representation introduces scalability challenges, as spatial complexity increases with scene size. For instance, reconstructing a 360$^{\circ}$ unbounded outdoor scene, like \textit{Bicycle} from MipNeRF360 \cite{barron2022mip} requires over 6 million Gaussians, with out-of-memory (OOM) errors occurring beyond 11 million Gaussians on a 24GB GPU. 
While larger memory can temporarily mitigate OOM issues, excessive Gaussians burden rasterizers and demand longer training, making real-time rendering impractical.

Urban-scale datasets often exhibit greater complexity with limited controlled acquisition conditions \cite{tancik2022block, martin2021nerf, chen2022hallucinated}. Variations in temporal factors (e.g. time of day, seasonal changes), weather patterns, and sensor configurations induce significant appearance discrepancies for identical objects. In addition, transient elements such as pedestrians and vehicles are frequently captured. 3DGS \cite{kerbl20233d}  tends to overfit these localized variations by introducing superfluous Gaussian primitives that align solely with a very limited number training viewpoints. However, these redundant Gaussians manifest as floating artifacts and structural inconsistencies when rendered from novel perspectives, severely compromising reconstruction fidelity.

To address these challenges, we propose a novel, efficient, and robust 3DGS method specifically designed for urban scene reconstruction. %
Building upon the original 3DGS framework, our approach incorporates multiple enhancements at different stages to achieve
efficient, high-quality reconstruction and real-time rendering.

\textbf{Promoting reconstruction efficiency:} We extend Block-NeRF \cite{tancik2022block} with 
a novel visibility-based image selection mechanism to minimize redundancy in data preprocessing (\Cref{sec:scene-and-data-division}). Additionally, the densification strategy is refined to concentrate computational resources within individual partitions, thereby avoiding redundant computation in irrelevant regions and significantly improving efficiency (\Cref{sec:inpartition_prioritized_densification}). To achieve real-time rendering under limited resources, we introduce a novel bottom-up LOD generation strategy. Each level is carefully controlled within a predefined budget and builds upon the previous one, effectively managing resource consumption during training. This strategy significantly improves reconstruction efficiency while maintaining superior quality compared to previous pruning or compression-based methods. The generated levels are dynamically selected during rendering using an LOD switching strategy, enabling real-time performance with limited resources (\Cref{sec:steerable-lod}).

\textbf{Improving reconstruction quality:} An appearance transform module is designed to learn and mitigate appearance variations of objects across different images, effectively neutralizing their negative impact on reconstruction. Once the reconstruction is complete, this module allows for flexible transformations of the scene’s appearance without negatively impacting rendering speed. Furthermore, we also propose scale and depth regularization to mitigate the generation of floaters and artifacts to further improving the reconstruction quality (\Cref{sec:quality-enhancements}).

Experimental results demonstrate that our method outperform existing methods in terms of reconstruction quality, resource efficiency, and rendering speed, enabling the reconstruction of arbitrarily large urban-scale scenes. 
The main contributions are summarized as follows:

\begin{itemize}
    \item We propose a novel visibility-based data division strategy and in-partition prioritized densification method, to achieve efficient urban-scale scene reconstruction.
    \item A controllable LOD generation strategy is designed to realize dynamic LOD selection and real-time rendering under limited resources.
    \item A fine-grained appearance transform module is developed to allow flexibly adjust the appearance without affecting the real-time performance, significantly enhancing the robustness to inter-image appearance variations.
\end{itemize}

\section{Related Works}
\label{sec:related_work}

\subsection{Novel View Synthesis}

\noindent\textbf{Neural Radiance Field (NeRF)}. NeRF \cite{mildenhall2021nerf} uses neural networks as an implicit representation of scenes to enable photorealistic novel view synthesis, has achieved remarkable success and garnered significant attention from researchers. %
Subsequent works have sought to enhance its quality \cite{barron2022mip, martin2021nerf, song2023sc, barron2023zip}, extent its application to dynamic scene \cite{park2021nerfies, pumarola2021d, song2023nerfplayer, park2021hypernerf}, surface reconstruction \cite{wang2021neus, yariv2021volume, li2023neuralangelo}, and so on.

However, the rendering speed is a major obstacle to the broader adoption of NeRF. Rendering a single 1080P image takes several minutes, which falls far short of real-time requirements.
Subsequent methods \cite{yu2021plenoctrees, fridovich2022plenoxels, reiser2021kilonerf, sun2022direct} have improved rendering efficiency. Notably, Instant-NGP \cite{muller2022instant} introduced multi-resolution hash encoding, achieving orders-of-magnitude improvements in efficiency. 
Real-world scenes typically feature richer content and intricate details, but these NeRF-based methods, constrained by model capacity, often learn only coarse, low-frequency information. %

\noindent\textbf{3D Gaussian Splatting}. 3DGS \cite{kerbl20233d} has recently emerged as a groundbreaking method. It explicitly represents scenes using 3D elliptical Gaussians and efficiently rasterizes them into images, enabling photorealistic reconstruction and real-time rendering. It significantly surpasses NeRF-based approaches in rendering efficiency and visual quality, attracting considerable research interest.
Numerous derivative works emerged that enhanced its reconstruction fidelity \cite{yu2024mip, sabour2024spotlesssplats, radl2024stopthepop, li2024dngaussian, kheradmand20253d}, extended it to dynamic scene \cite{luiten2024dynamic, wu20244d, li2024spacetime, yang2024deformable}, and explored its robustness under diverse conditions \cite{zhang2024gaussian, sabour2024spotlesssplats, dahmani2024swag}.
Nonetheless, explicit representations inherently suffer from high computational resource demands for large-scale scenes.
To address this, various methods \cite{niedermayr2024compressed, morgenstern2024compact, levoy2023light, chen2024hac} have been proposed to compress the models. Recently, Taming3DGS \cite{mallick2024taming} introduced a steerable densification strategy, enabling the control of 3DGS memory usage during the training phase while minimizing the negative impact in fidelity. Despite these advancements, the current focus of these methods remains on small-scale scenes, scaling 3DGS to truly urban-scale environments remains challenging due to rapidly increasing computational and memory demands.

\subsection{Large Scale Scene Reconstruction}

Several NeRF-based methods
have proposed approaches for reconstructing large-scale scenes.
Block-NeRF \cite{tancik2022block} and Mega-NeRF \cite{turki2022mega} apply a manual divide-and-conquer strategy, segmenting scenes into multiple regions represented by separate MLPs.
Switch-NeRF \cite{zhenxing2022switch} replaces manual partitioning with a learnable gating network.
BungeeNeRF \cite{xiangli2022bungeenerf} progressively reconstructs urban scenes at multiple detail levels by dynamically adding modules.
Grid-NeRF \cite{xu2023grid} employs a dual-branch structure with coarse-to-fine refinement based on multi-resolution hash encoding.
GP-NeRF \cite{zhang2025efficient} combines 3D hash-grids and 2D plane features for higher accuracy and efficiency.
However, these methods continue to face significant challenges in rendering speed and reconstruction fidelity, limiting their practical applicability in large-scale urban environments.

Several recent works have extended 3DGS to large-scale scenes.
Grendel-GS \cite{zhao2024scaling} addresses resource constraints by leveraging multiple GPUs, yet scaling hardware with scene complexity remains impractical.
VastGaussian \cite{lin2024vastgaussian} adopts a divide-and-conquer approach combined with CNN-based color transformations to handle appearance variations; however, it fails to address the real-time rendering challenge, and image-space transformations often cause unstable optimization and unnatural color artifacts.
Hierarchical-3DGS \cite{kerbl2024hierarchical} and CityGaussian \cite{liu2024citygaussian} take a step further by introducing LOD strategies for real-time rendering of large-scale scenes. However, they lack mechanisms to control resources during training, relying instead on post-training compression and extensive fine-tuning, which inevitably compromises quality in large-scale scenes due to higher compression requirements.

Different from previous methods, our approach ensures strict resource control during training, eliminating the need for post-training compression. Furthermore, our novel appearance transform module enables fine-grained adjustments at the Gaussian level, enhancing robustness and flexibility while maintaining real-time performance.%

\section{Method}
\label{sec:method}

\subsection{Preliminary}
\label{sec:preliminary}

3DGS \cite{kerbl20233d} represents a scene using a set of explicit 3D Gaussians. Each 3D Gaussian is defined by: 
\begin{equation}
    G(x)=e^{-\frac{1}{2}(x-\mu)^{T}\Sigma^{-1}(x-\mu)}
\end{equation}
where $\mu$ is the center and $\Sigma$ is the covariance matrix that can be decomposed into a rotation $\mathbf{R} \in SO(3)$ and a scale $\mathbf{S} \in \mathbb{R}^{3}$. Each Gaussian is assigned a color $c$ and an opacity $o$ to represent scene appearance.
When rendering an image, the 3D Gaussians are first projected onto the image plane, forming 2D Gaussians $G^{'}(x)$. These are then processed in depth order, to compute each pixel color using $\alpha$-blending:
\begin{equation}
    \colorfunc(x_{p}) = \sum_{i \in N}{c_{i}\gsopacity_{i}G^{'}_{i}(x_{p})}\prod_{j=1}^{i-1}(1-\gsopacity_{j}G^{'}_{j}(x_{p}))
\end{equation}
where $x_{p}$ represents a pixel, $\gsopacity_{i}$ and $c_{i}$ denote the opacity and color of the $i$-th Gaussian respectively, and $N$ represents the set of Gaussians covering the pixel $x_{p}$.

During the training process, 3DGS periodically performs a densification operation on those Gaussians that exhibit relatively high mean gradients in the image space, thereby improving their ability to capture the underlying geometry. As described in \cite{kerbl20233d}, the loss function of 3DGS includes the L1 and D-SSIM metrics, computed between the rendered image $\predimage$ and its corresponding ground-truth image $\gtimage$:
\begin{equation}
    \label{eq:3dgs-loss}
    \mathcal{L} = (1-\lambda_{\textrm{}})\mathcal{L}_{1}(\gtimage, \predimage) + \lambda_{\textrm{}}\mathcal{L}_{\textrm{D-SSIM}}(\gtimage, \predimage)
\end{equation}
where $\lambda_{\textrm{}}$ is a hyperparameter, which is set to 0.2 in \cite{kerbl20233d}.

By optimizing the attributes of the Gaussians and carrying out densification to minimize this loss, 3DGS ultimately fulfills its goal of reconstructing the target scene.

\subsection{Scene and Data Division}
\label{sec:scene-and-data-division}
\begin{figure*}[t]
    \centering
    \includegraphics[width=\linewidth]{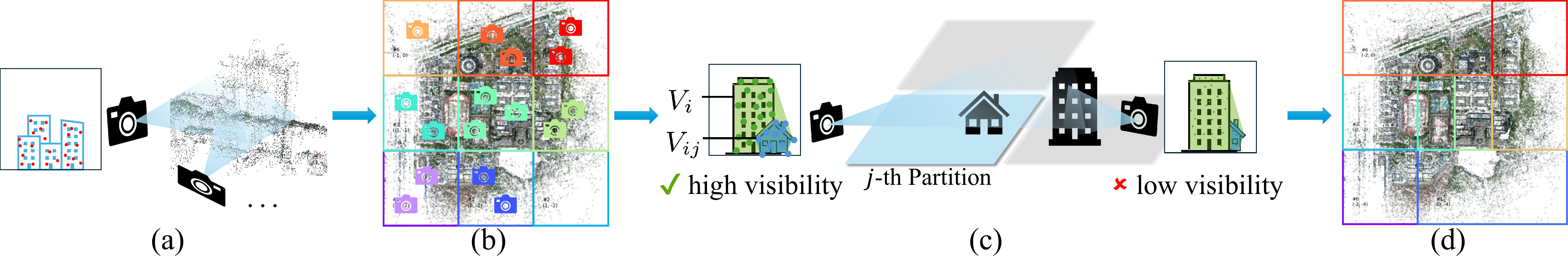}
    \caption{\tabletitle{The process of scene and data division.} (a) Obtain the 3D point cloud and its corresponding 2D feature points through estimating camera poses by SfM. (b) Determine training cameras based on their spatial locations after partitioning the scene into smaller regions. (c) Cameras outside the partitions are assigned based on visibility. The $\visibility_{i}$ and $\visibility_{ij}$ correspond to the green and blue regions in the image, respectively. Only cameras with high visibility are utilized for training. (d) Adjust the partition sizes to achieve a more balanced workload.}
    \label{fig:division-pipeline}
\end{figure*}

We partition the scene horizontally and then assign training images to them. Initially, images are assigned based on their locations: an image $\gtimage_{i}$ is assigned to the $j$-th partition if its location falls within the partition's bounding box. For images located outside the bounding box, we calculate a point-based visibility with respect to the partition to determine whether they should be assigned to it. This strategy is illustrated in \Cref{fig:division-pipeline}. 

\subsubsection{Point-based Visibility}
\label{sec:point-based-visibility}

Visibility is calculated using the 3D point cloud and its association with 2D feature points, both generated by Structure from Motion (SfM).
For an unselected image $\gtimage_{i}$, the 3D point cloud of the scene is projected onto its image plane, and compute its convex hull area $\visibility_\text{i}$. 
Subsequently, we leverage the relationship between the 2D feature points of the $\gtimage_{i}$ and 3D points to extract those within the $j$-th partition. These extracted feature points are then used to calculate the convex hull area $\visibility_{ij}$. Therefore, visibility is calculated by $\visibility_{ij} / \visibility_{i}$.

The feature points are inherently occlusion-aware, ensuring that only visible regions are considered. This effectively prevents redundant image selection, achieving higher quality with the same number of training iterations.

\subsubsection{Partition Rebalancing}

In practice, the central partition typically contains more high-visibility images than the edge partitions, causing an unbalanced workload. To address this, we adjust the partition sizes after visibility-based data division: partitions with too few images are merged with the smallest neighboring partition, while those with too many images are subdivided. This process repeats iteratively until the image distribution is reasonably uniform.

\subsection{In-Partition Prioritized Densification}
\label{sec:inpartition_prioritized_densification}
\begin{figure}
    \centering
    \begin{overpic}[trim={0 0 2.3cm 0},clip,width=0.5\linewidth]{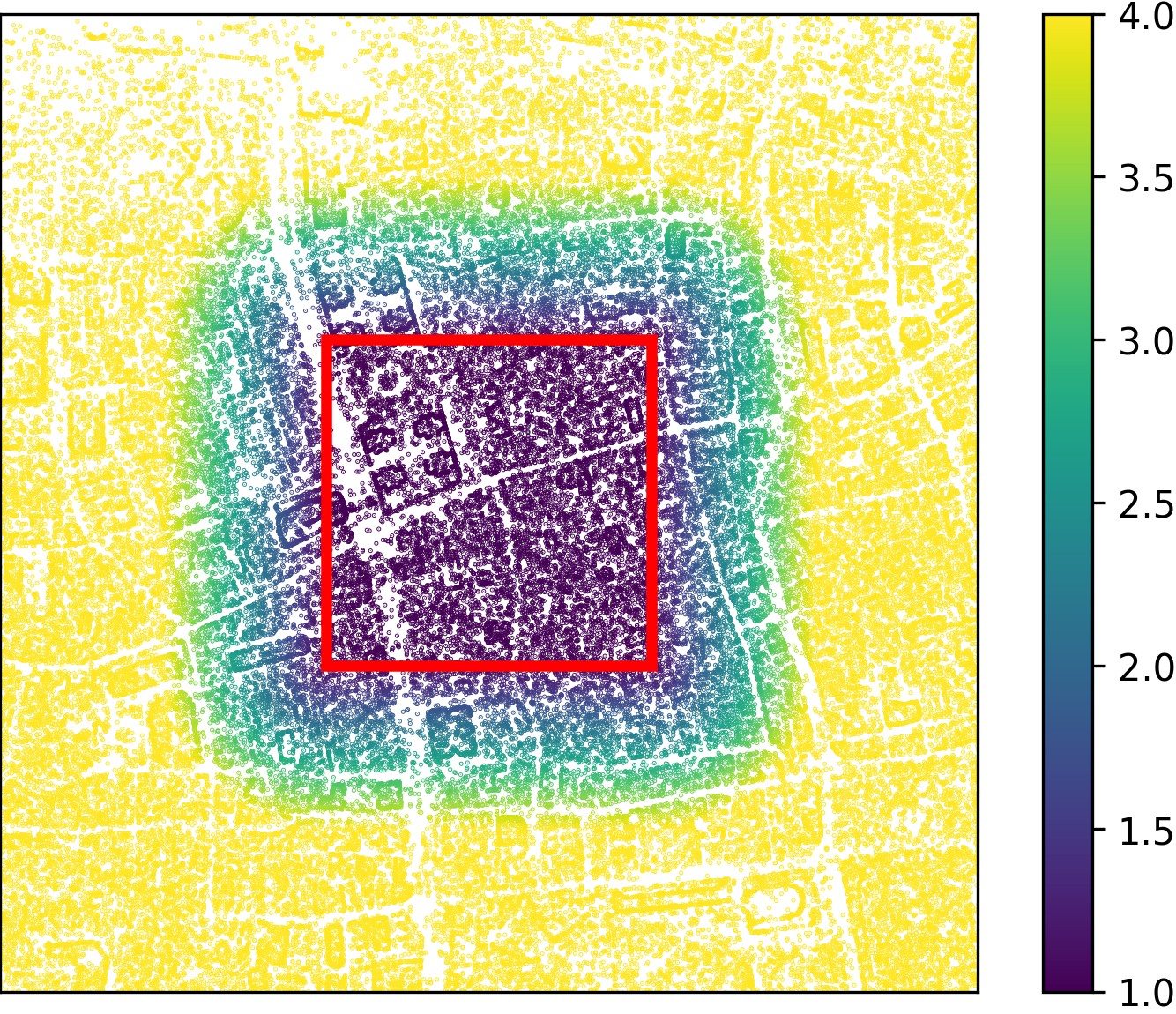}
        \put(99, 88){$\hat{\tau}_\textrm{max}$}
        \put(99, 0){$\hat{\tau}_\textrm{min}$}
    \end{overpic}
    \caption{\tabletitle{In-partition prioritized densification.} The red rectangle is the partition bounding box, and points represent Gaussians. Point colors indicate gradient thresholds.}
    \label{fig:ff-dense}
\end{figure}

Excessive resource allocation to areas outside the partition is unnecessary during training. However, simply increasing the gradient threshold in these areas may lead to Gaussians within the partition shifting outward to compensate for under-reconstruction, thereby degrading the quality of partition boundaries. To solve this problem, as shown in \Cref{fig:ff-dense} we propose a distance-related threshold for each Gaussian:
\begin{equation}
    \tau_{i} = \hat{\tau}_{\textrm{min}}\left(\frac{
    \min(d_{i}, \hat{d}_{\textrm{max}})
    }{
    \hat{d}_{\textrm{max}}
    } \cdot (\eta - 1) + 1 \right)
\end{equation}
Where the $d_{i}$ is the the distance between the $i$-th Gaussian and the partition, $\hat{d}_{\textrm{max}}$ is the distance at which the maximum threshold $\hat{\tau}_{\textrm{max}} = \hat{\tau}_{\textrm{min}} \cdot \eta\ (\eta \ge 1)$ is applied. The $i$-th Gaussian will only be considered for densification if and only if its mean gradient satisfies $\bar{\Delta}_{G_{i}} > \tau_{i}$.

This strategy effectively reduces resource consumption in out-of-partition regions, accelerating training while preserving final reconstruction quality.

\subsection{Controllable Level-of-detail}
\begin{figure}
    \begin{subfigure}{0.57\linewidth}
        \includegraphics[width=\columnwidth]{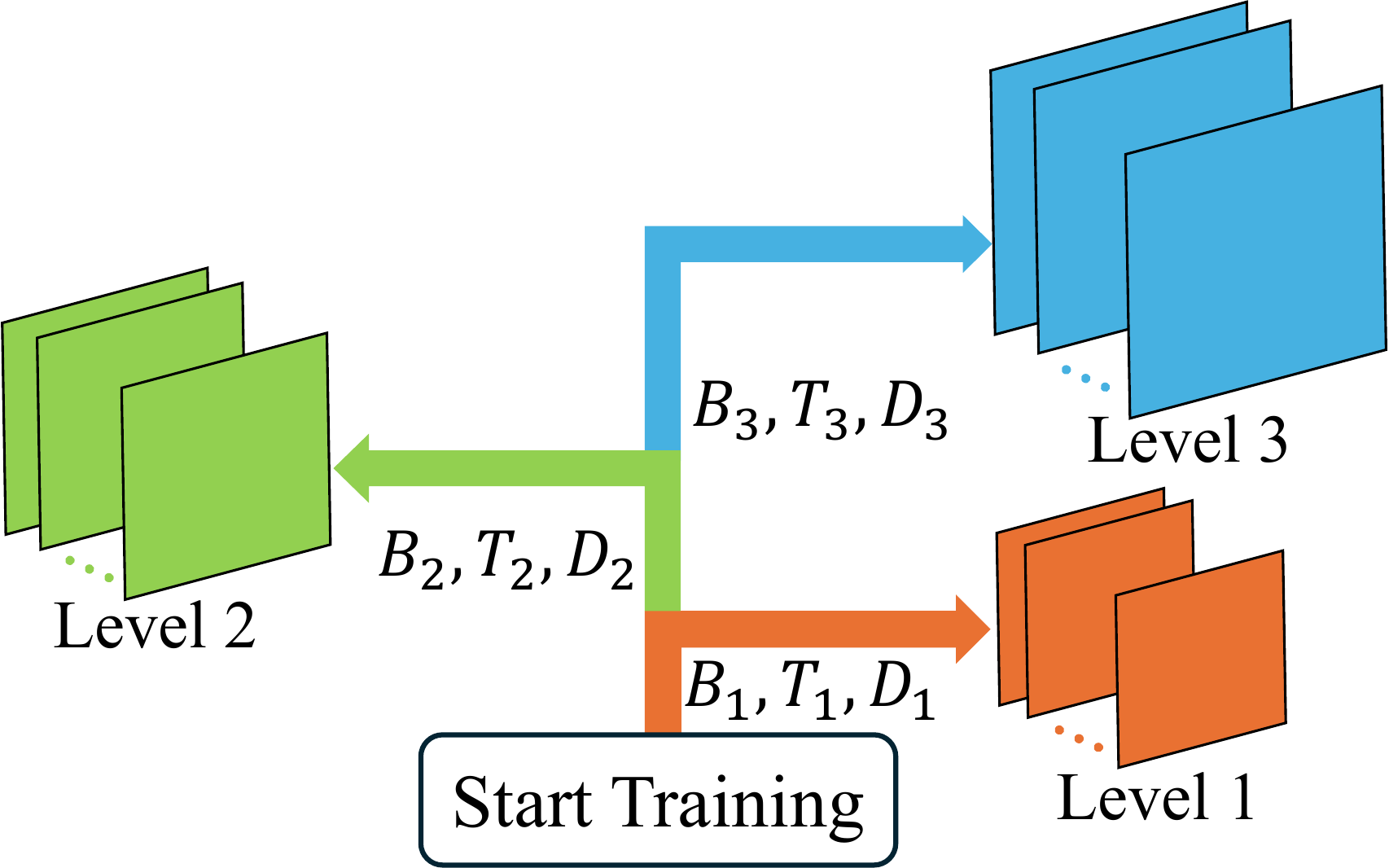}
        \caption{Detail level generation.}
        \label{fig:lod-generation}
    \end{subfigure}
    \hfill
    \begin{subfigure}{0.37\linewidth}
        \includegraphics[width=\columnwidth]{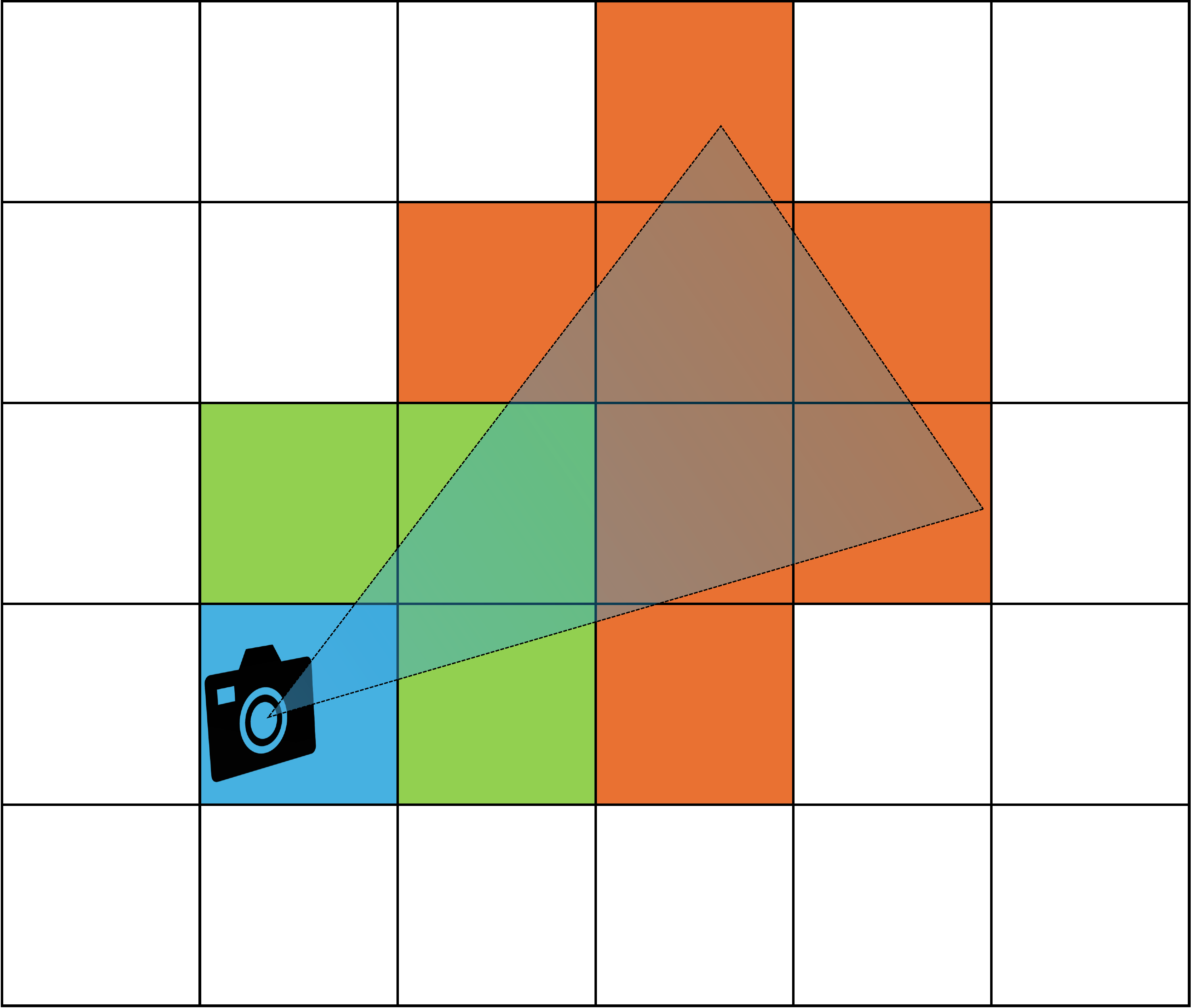}
        \caption{Detail level Selection.}
        \label{fig:lod-switching}
    \end{subfigure}
    \caption{\tabletitle{Controllable LOD generation and detail level selection.} (a) During training, detail levels are progressively generated in a bottom-up manner, guided by resource budgets $\denseb$, densification intervals $\densei$, and downsampling factors $\densed$. (b) During rendering, detail levels are dynamically selected based on the partition-camera distance, assigning higher levels to closer partitions and lower levels to distant ones. Invisible partitions are culled.}
    \label{fig:lod}
\end{figure}

\label{sec:steerable-lod}

The original 3DGS densification strategy \cite{kerbl20233d} lacks resource constraints, making it impractical for urban-scale scenes. To address this, we extend the steerable densification strategy of \cite{mallick2024taming}, generating multiple levels of detail in a bottom-up manner while strictly enforcing predefined resource limits. During rendering, appropriate levels are dynamically selected, ensuring efficient real-time rendering.

\subsubsection{Controllable Detail Level Generation}
The number of LOD levels $l \in \mathbb{Z}^{+}$, the budget $\denseb$, densify interval $\densei$ and image downsample factor $\densed$ for each level are defined as:
\begin{equation}
\{\denseb_{i} \in \mathbb{Z}^{+} \mid \denseb_1 < \denseb_2 < \dots < \denseb_l\}
\end{equation}
\begin{equation}
\{\densei_{i} \in \mathbb{Z}^{+} \mid \densei_1 > \densei_{2} > \dots > \densei_l\}
\end{equation}
\begin{equation}
\{\densed_i \in (0, 1] \mid \densed_1 < \densed_2 < \dots < \densed_l = 1\}
\end{equation}
where the budget $B$ is the parameter of the densification strategy of \cite{mallick2024taming}.
These parameters implies that when training at lower detail levels, a lower budget, longer densification intervals, and lower-resolution images will be utilized. 
This avoids unnecessary focus on high-frequency details when reconstructing lower levels.

As shown in \Cref{fig:lod-generation}, for each partition, the reconstruction begins with the 1st level.
For the $i$-th level, 
upon completion of training, a checkpoint is created, and the budget, interval and downsample factor are changed to $\denseb_{i+1}$, $\densei_{i+1}$ and $\densed_{i+1}$ to facilitate the generation of the next level, until all levels are progressively generated.

This process is entirely end-to-end, eliminating the need for extensive post-processing steps common in compression strategy. Experiments show that our method achieves higher quality than compression-based method while enabling faster completion by utilizing low-resolution images and a smaller budget for lower levels.

\subsubsection{Detail Level Selection}

During rendering, we adopt the strategy proposed by CityGaussian \cite{liu2024citygaussian}, performing detail level selection at the partition level, as illustrated in the \Cref{fig:lod-switching}. However, LOD selection incurs additional computational overhead. To further improve rendering efficiency, we also adopt the tile-based culling purposed by StopThePop \cite{radl2024stopthepop} to disregard Gaussians with a low contribution in tiles.

\subsection{Quality Enhancements}
\label{sec:quality-enhancements}
To further improve the reconstruction quality, a series of methods are introduced to overcome the key limitations of 3DGS in this subsection.
\Cref{sec:appearance-transform-model} proposes the appearance transform module to ensure robust adaptation to appearance variations in images. 
To mitigate local overfitting and eliminate artifacts, \Cref{sec:scale-regularization} and \Cref{sec:depth-regularization} propose two regularization techniques.
\Cref{sec:aa-absgs} adopts an anti-aliasing technique and enhances the fidelity of fine details.
\Cref{sec:transient-object-removal} presents a method to remove transient objects.

\subsubsection{Appearance Transform Module}
\label{sec:appearance-transform-model}
\begin{figure}
\centering
\includegraphics[trim={2.0cm 0.45cm 2.7cm 0.25cm},clip,width=\linewidth]{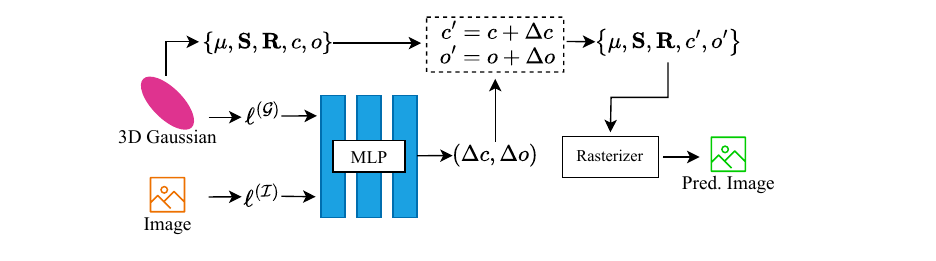}
\caption{\tabletitle{Illustration of the appearance transform.} For each image and 3D Gaussian, $\gsembedding$ represents the Gaussian embedding and $\imageembedding$ represents the image embedding, respectively. By the lightweight MLP, $\Delta c$ and $\Delta o$ can be predicted to adjust the color $c$ and opacity $o$ of the 3D Gaussian.}
\label{fig:appearance}
\end{figure}

3DGS \cite{kerbl20233d} tends to generate floaters to overfit appearance variations across training images, degrading reconstruction quality. Existing methods like NeRF-W \cite{martin2021nerf} and SWAG \cite{dahmani2024swag} mitigate appearance inconsistencies by assigning embeddings only to images, limiting flexibility and leading to suboptimal results.
We propose a fine-grained appearance transform module that assigns embeddings to both individual images and each 3D Gaussian independently. As shown in \Cref{fig:appearance}, these embeddings are processed by a lightweight MLP to predict per-Gaussian color and opacity offsets, enabling precise and adaptable appearance adjustments. This approach enhances robustness against inconsistencies, eliminates floating artifacts, and supports post-reconstruction appearance editing. Once computed, the offsets are reusable for subsequent frames as long as the image embedding remains unchanged, ensuring no additional rendering cost.

\textbf{Similarity regularization:} Adjacent Gaussians typically show similar variations. To leverage this property and avoid overfitting, we introduce a similarity regularization to encourage more similar embeddings among neighboring Gaussians. Since normalized embeddings are used, cosine similarity is adopted to compute the loss:
\begin{equation}
\mathcal{L}^{\textrm{sim}}_{i,j} = w_{i, j}\left(
1 - \frac{
\gsembedding_{i} \cdot \gsembedding_{j}
}{
\|\gsembedding_{i}\| \|\gsembedding_{j}\|
}
\right)
\end{equation}
\begin{equation}
\label{eq:sim}
\mathcal{L}_{\textrm{sim}} = 
\frac{1}{|M|\binom{k}{2}} \sum_{i\in M} \sum_{\substack{j, l \in \textrm{knn}_{i;k} \\ j < l}} \mathcal{L}^{\textrm{sim}}_{j,l}
\end{equation}
where $\gsembedding_{i}$ and $\gsembedding_{j}$ are the embeddings of the $i$-th and $j$-th Gaussian, respectively. $M$ is the set of Gaussians sampled randomly for regularization, $\textrm{knn}_{i;k}$ denotes K-nearest neighbors of $i$, $w_{i,j}$ is the decay factor \cite{luiten2024dynamic} and determined by the distance between the two Gaussians:
\begin{equation}
w_{i,j} = \exp{\left(
-\lambda_{w} \|
\mu_{i} - \mu_{j}
\|
\right)}
\end{equation}
where $\lambda _{w}$ controls the decay rate, chosen based on the scale of the scene.

\textbf{Opacity offset regularization:} Based on real-world experience, most appearance transformations do not involve changes in transparency. To prevent the model from unnecessarily overusing transparency to fit color variations, we introduce an additional regularization term for the opacity offset, restricting transparency changes to only a few Gaussian components:
\begin{equation}
    \mathcal{L}_{\Delta o} = \frac{1}{N}\sum^{N}_{i=1}{\Delta o_{i}}.
\end{equation}

\subsubsection{Scale Regularization}
\label{sec:scale-regularization}

During optimization, we frequently observed scale-anomalous Gaussians, such as those exceeding the scene size or forming highly anisotropic shapes. These anomalies cause severe artifacts during camera rotation due to ordering inversion \cite{radl2024stopthepop}. To address this, we introduce a scale regularization with two components: a maximum constraint to prevent excessive growth and a ratio constraint to maintain reasonable proportions. This regularization effectively suppresses artifacts caused by scale anomalies and improves the stability and consistency of the rendering results.

The maximum constraint limits the upper bound of Gaussian scales to prevent them from growing to unreasonable values: 
\begin{equation}
\mathcal{L}_{\textrm{ms}}=
\frac{
\sum_{i}{
\mathbbm{1}\left\{\mathbf{S}_{i} >  s_\textrm{max}\right\}
}
\cdot
\mathbf{S}_{i}
}{
\sum_{i}{
\mathbbm{1}\left\{\mathbf{S}_{i} >  s_\textrm{max}\right\}
} + \delta}
\end{equation}
where the $\mathbbm{1}$ is an indicator function that takes the value 1 when the condition is true and 0 otherwise, $\mathbf{S}_{i}$ is the scales of the $i$-th Gaussian, $s_\textrm{max}$ is the maximum acceptable scale, $\delta$ prevents division by zero.

The ratio constraint enforces a limit on the ratio between the first and second largest scales of a Gaussian, avoiding highly anisotropic shapes:
\begin{equation}
r_{i} = \frac{\max(\mathbf{S}_{i})}{\textrm{median}({\mathbf{S}_{i}})}
\end{equation}
\begin{equation}
\mathcal{L}_\textrm{r} =
\frac{
\sum_{i}{
\mathbbm{1}
\left\{
r_{i} > r_\textrm{max}
\right\}
} \cdot r_{i}
}{\sum_{i}{
\mathbbm{1}
\left\{
r_{i} > r_\textrm{max}
\right\}
} + \delta}
\end{equation}
where $r_{\textrm{max}}$ is the maximum acceptable ratio.

\subsubsection{Depth Regularization}
\label{sec:depth-regularization}

Inspired by the DNGaussian \cite{li2024dngaussian}, we utilize Depth Anything V2 \cite{yang2025depth} to predict fine-grained depth maps from RGB images and align them to actual depths using the SfM point cloud. During training, we alternate between \cite{li2024dngaussian}'s hard depth and soft depth regularization. 
The depth loss $\mathcal{L}_{d}$ is the L1 error between the rendered and estimated depths.
This effectively mitigates the floating artifacts, significantly enhancing visual realism.

\subsubsection{Anti-aliasing and Detail Enhancement}
\label{sec:aa-absgs}
To improve rendering quality, we integrate anti-aliasing from Mip-Splatting \cite{yu2024mip} and adopt AbsGS \cite{ye2024absgs} to enhance fine details.
These adaptations effectively reduce artifacts and boost visual fidelity.

\subsubsection{Transient Objects Removal}
\label{sec:transient-object-removal}
In practical data collection, the presence of transient objects (e.g., pedestrians, vehicles) is unavoidable. These transient objects introduce visual artifacts into the reconstructed scene, adversely affecting the final reconstruction quality. We utilize an open-world object detection model \cite{ren2024grounding} to identify the 2D bounding boxes of these transient objects and employ them as prompts for a semantic segmentation model \cite{ravi2024sam} to generate fine-grained masks.

\subsection{Loss of Individual Partition Training}
The loss for partition training consists of five components: 
\begin{equation}
\mathcal{L'} = \mathcal{L}
 + 
\lambda_{\textrm{sim}}
\mathcal{L}_{\textrm{sim}}
+\lambda_{\Delta o}\mathcal{L}_{\Delta o}
+\lambda_{\textrm{d}}\mathcal{L}_{\textrm{d}}
+
\lambda_{\textrm{s}}(
\mathcal{L}_{\textrm{ms}}
+
\mathcal{L}_\textrm{r}
)
\end{equation}
Where $\mathcal{L}$ is \Cref{eq:3dgs-loss}. The $\lambda_{\textrm{sim}} = 0.2$, $\lambda_{\Delta o} = 0.05$ and $\lambda_{\textrm{s}} = 0.05$. The initial value of $\lambda_{\textrm{d}}$ is $0.5$, with an exponential decay scheduler reduces it to a final value of $0.01$.

\begin{table*}[htbp]
	\begin{center}
		\resizebox{\linewidth}{!}{
			\begin{tabular}{l|rrrrr|rrrrr|rrrrr}
				\toprule
				Scene   &   \multicolumn{5}{c|}{\emph{Rubble}} & \multicolumn{5}{c|}{\emph{\ourcampusname{}}} & \multicolumn{5}{c}{\emph{BigCity}} \\
				\midrule
				Metrics &  \tabmetrichead{} &
				\tabmetrichead{} &
				\tabmetrichead{} \\
				\midrule
                    Switch-NeRF & 0.544 & 23.05 & 0.508 & -- & $<$0.1 & 0.574 & 21.96 & 0.587 & -- & $<$0.1 & 0.469 & 20.39 & 0.645 & -- & $<$0.1 \\
                    CityGaussian (no LOD) & \underline{0.813} & \underline{25.77} & \textbf{0.228} & \underline{9.60} & 57.0 & \underline{0.776} & \underline{22.57} & 0.282 & \underline{11.76} & 34.6 & 0.825 & \underline{24.57} & \underline{0.240} & \underline{58.47} & \underline{23.1} \\
                    Hierarchical-3DGS (no LOD) & 0.768 & 23.76 & 0.275 & 11.13 & 49.6 & 0.772 & 21.23 & \underline{0.260} & 17.42 & 29.8 & -- & -- & -- & -- & -- \\
                    3DGS & 0.796 & 25.72 & 0.304 & \textbf{7.10} & \textbf{108.9} & 0.763 & 22.02 & 0.350 & \textbf{3.66} & \textbf{152.7} & \underline{0.830} & 24.52 & 0.279 & \textbf{33.21} & 22.7 \\
                    Ours (no LOD) & \textbf{0.826} & \textbf{27.29} & \textbf{0.228} & 13.52 & \underline{78.2} & \textbf{0.822} & \textbf{25.85} & \textbf{0.232} & 25.58 & \underline{41.2} & \textbf{0.847} & \textbf{26.62} & \textbf{0.219} & 75.15 & \textbf{23.7} \\
                    \midrule
                    CityGaussian & \underline{0.785} & \underline{24.90} & \underline{0.256} & \textbf{2.95} & \textbf{105.2} & \underline{0.770} & \underline{22.33} & 0.286 & \textbf{3.27} & \textbf{69.2} & 0.712 & 22.24 & 0.344 & \textbf{3.04} & \textbf{122.5} \\
                    Hierarchical-3DGS & 0.741 & 23.38 & 0.300 & 7.23 & 57.4 & 0.760 & 21.12 & \underline{0.274} & 11.21 & 34.6 & \underline{0.775} & \underline{23.17} & \underline{0.289} & 17.09 & 3.2 \\
                    Ours & \textbf{0.814} & \textbf{27.03} & \textbf{0.245} & \underline{3.60} & \underline{99.7} & \textbf{0.816} & \textbf{25.71} & \textbf{0.240} & \underline{6.65} & \underline{63.9} & \textbf{0.838} & \textbf{26.41} & \textbf{0.231} & \underline{6.84} & \underline{73.0} \\
                    \bottomrule
			\end{tabular}
		}
		\caption{\tabletitle{Quantitative results on three large scene datasets.} We report SSIM$\uparrow$, PSNR$\uparrow$, LPIPS$\downarrow$, the number of Gaussians (\#G, in $10^6$)$\downarrow$ and FPS$\uparrow$ on test views.
        The \textbf{best} and \underline{second best} results are highlighted. 
        All missing results are denoted by a ``--". 
        }
		\label{tab:compare}
		\vspace{-3mm}
	\end{center}
	\centering
\end{table*}

\begin{figure*}
   \centering
   \setlength{\tabcolsep}{1pt}
\renewcommand{\arraystretch}{0.5}
\begin{tabular}{ccccc}
{\small Ground Truth} & {\small 3DGS} & {\small CityGaussian} & {\small Hierarchial-3DGS} & {\small Ours} \\

    \spyimg{0.19\textwidth}{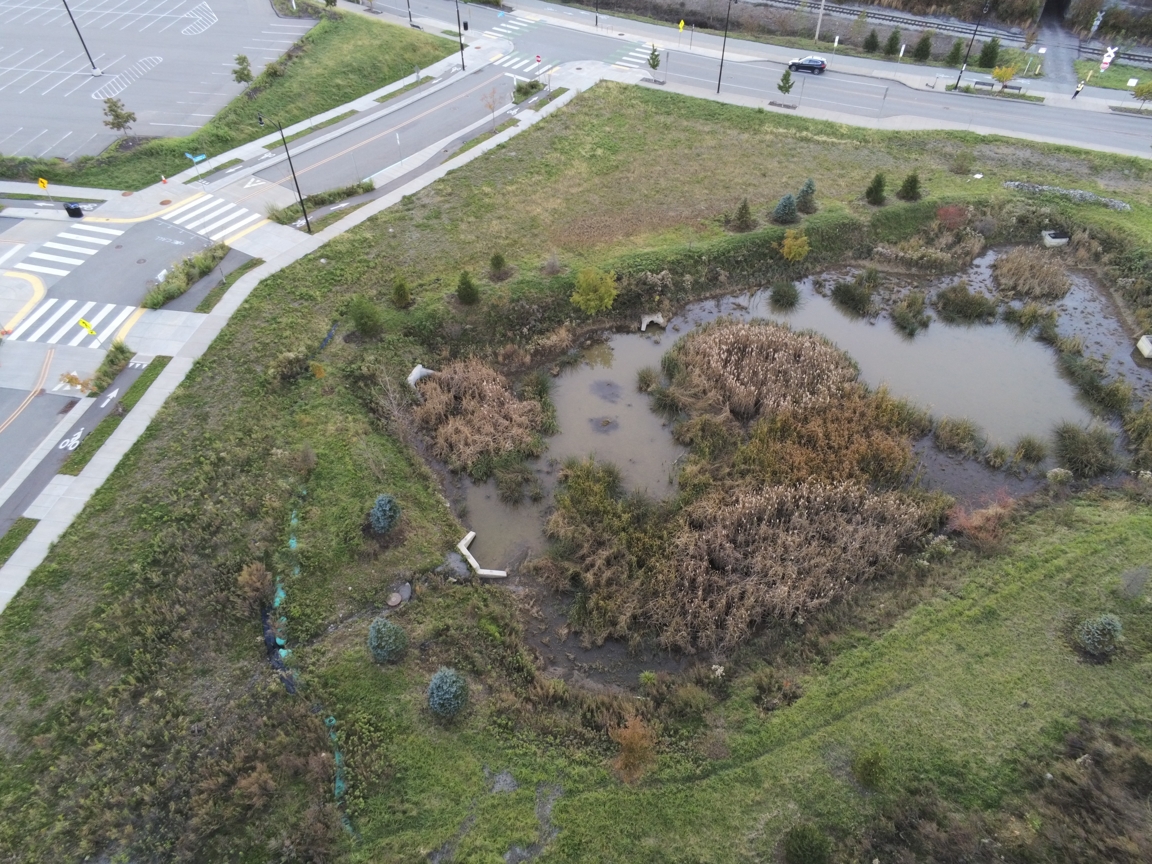}{0.57,2.05}{3,0.7} &
    \spyimg{0.19\textwidth}{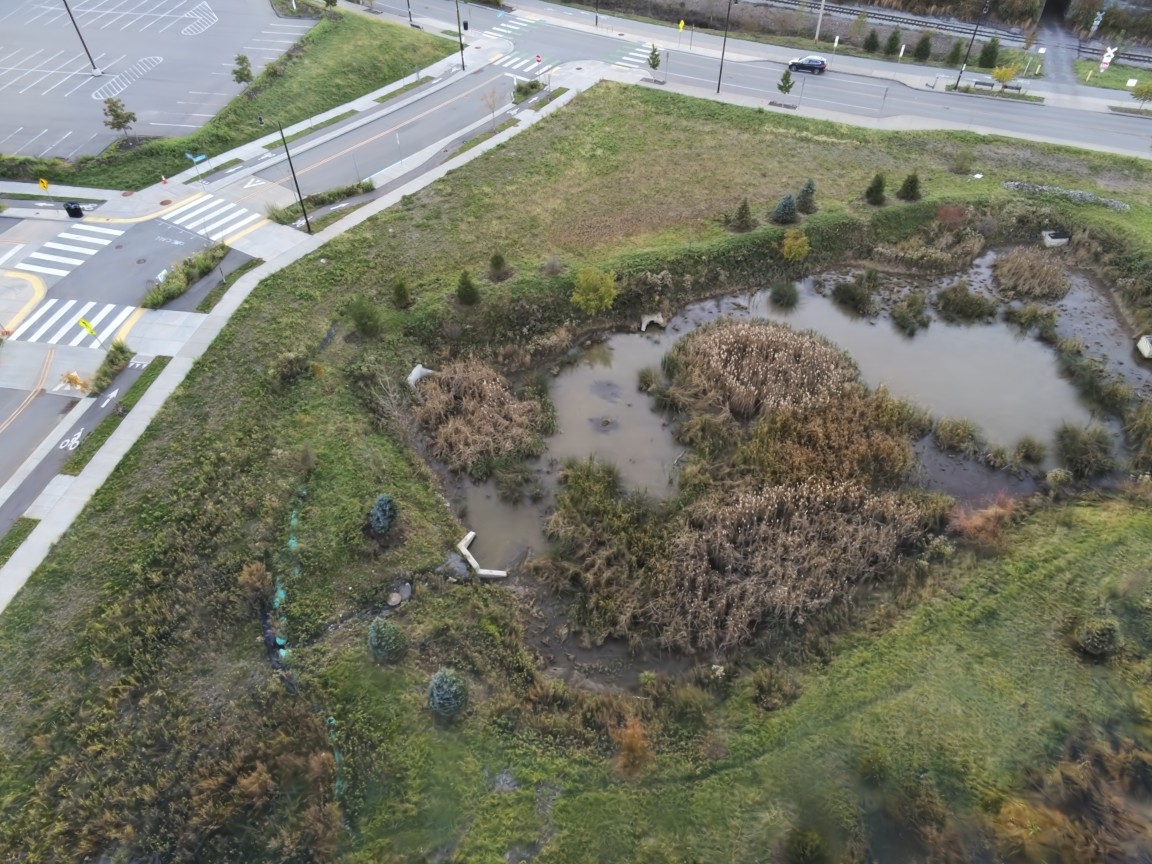}{0.57,2.05}{3,0.7} &
    \spyimg{0.19\textwidth}{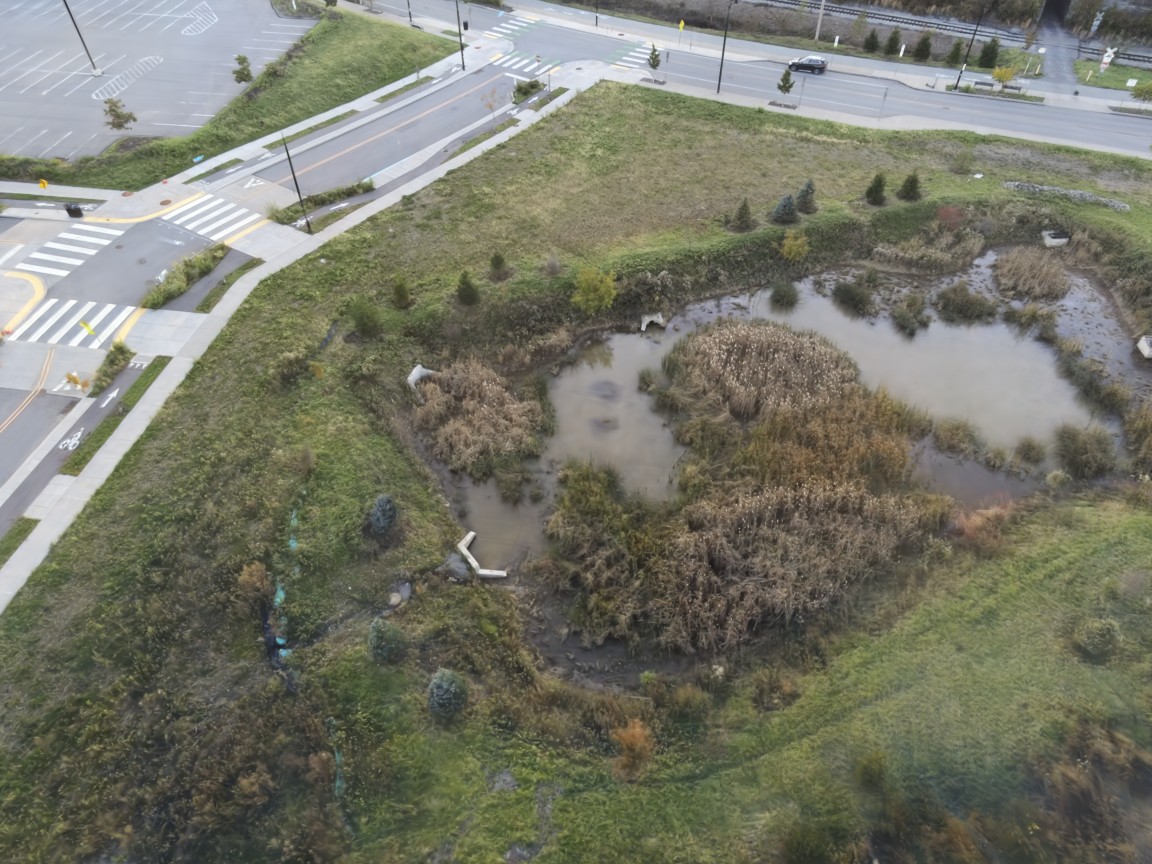}{0.57,2.05}{3,0.7} &
    \spyimg{0.19\textwidth}{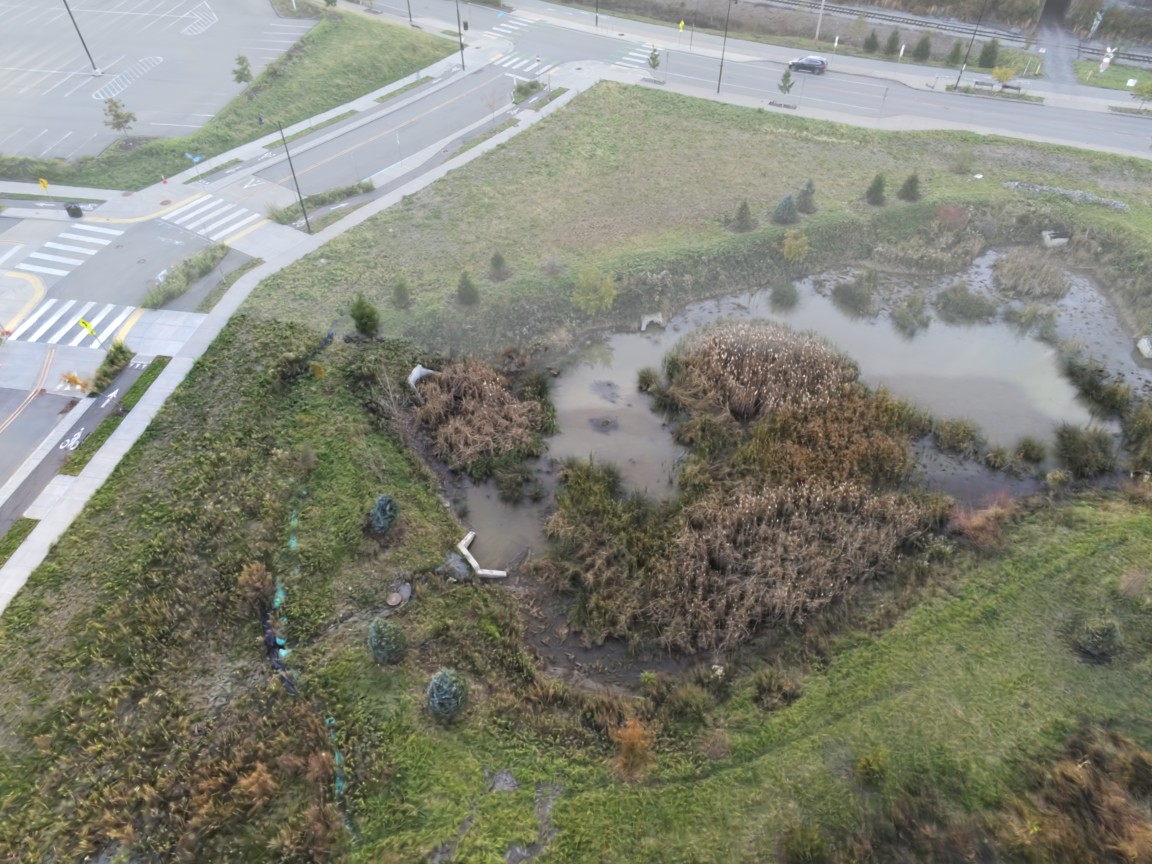}{0.57,2.05}{3,0.7} &
    \spyimg{0.19\textwidth}{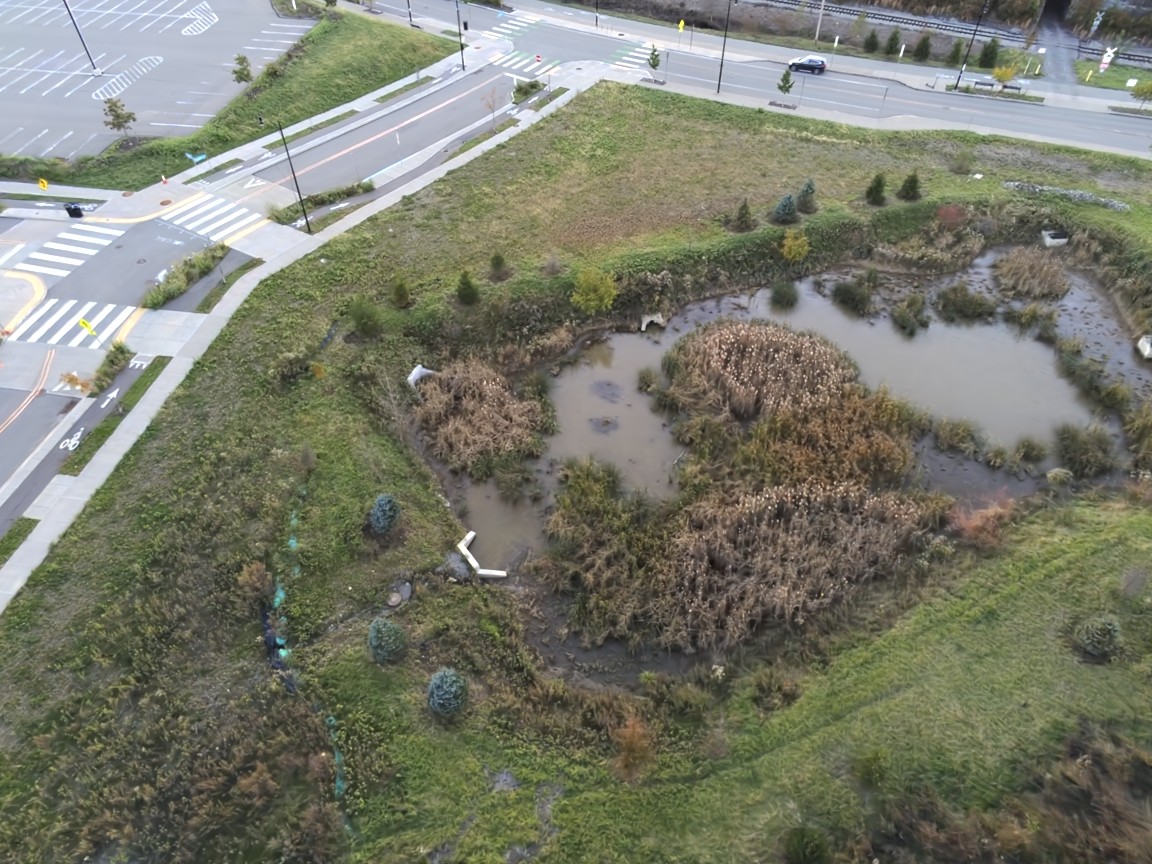}{0.57,2.05}{3,0.7} \\

    \spyimg{0.19\textwidth}{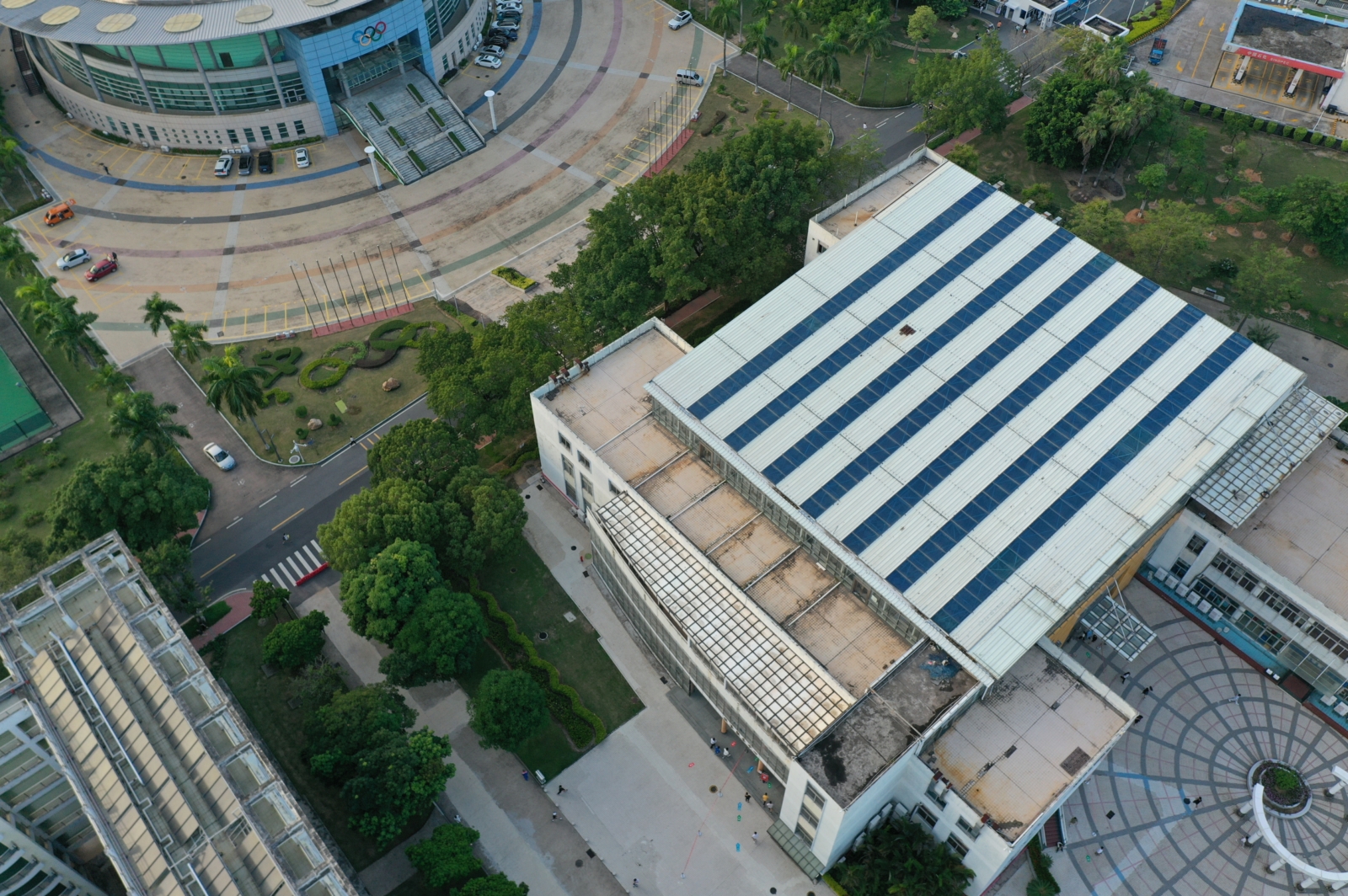}{1.52,2.0}{3,0.7} &
    \spyimg{0.19\textwidth}{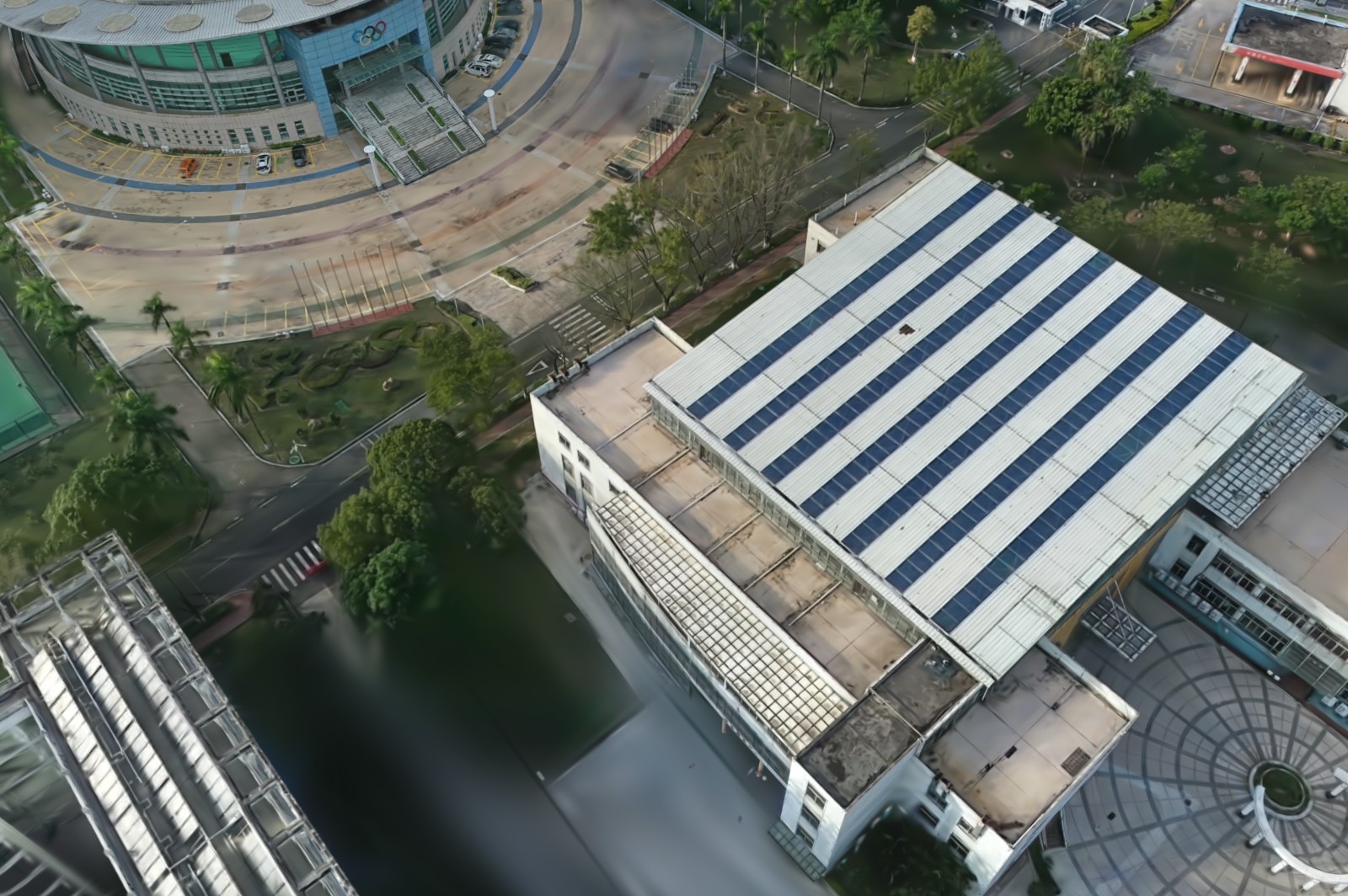}{1.52,2.0}{3,0.7} &
    \spyimg{0.19\textwidth}{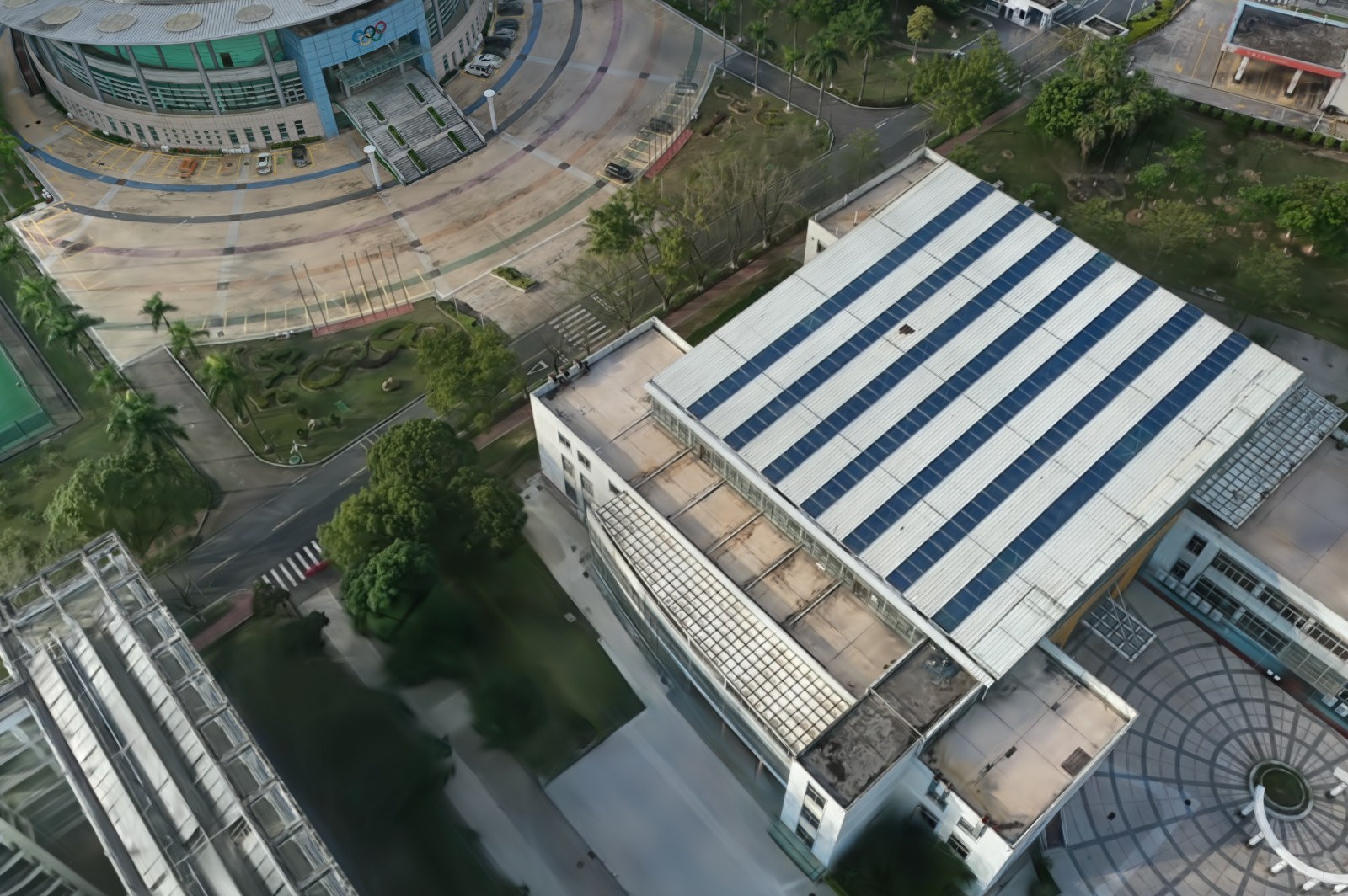}{1.52,2.0}{3,0.7} &
    \spyimg{0.19\textwidth}{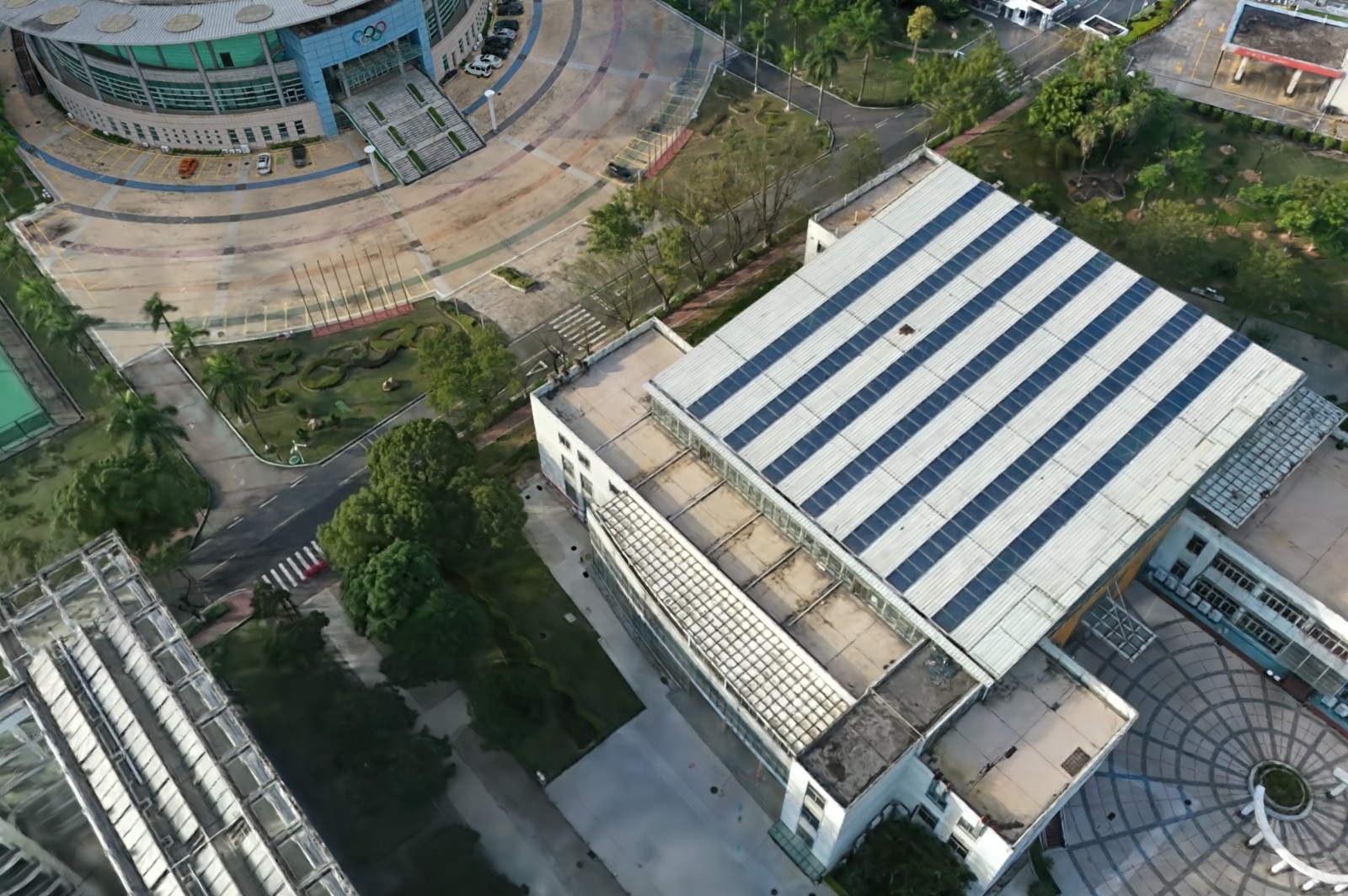}{1.52,2.0}{3,0.7} &
    \spyimg{0.19\textwidth}{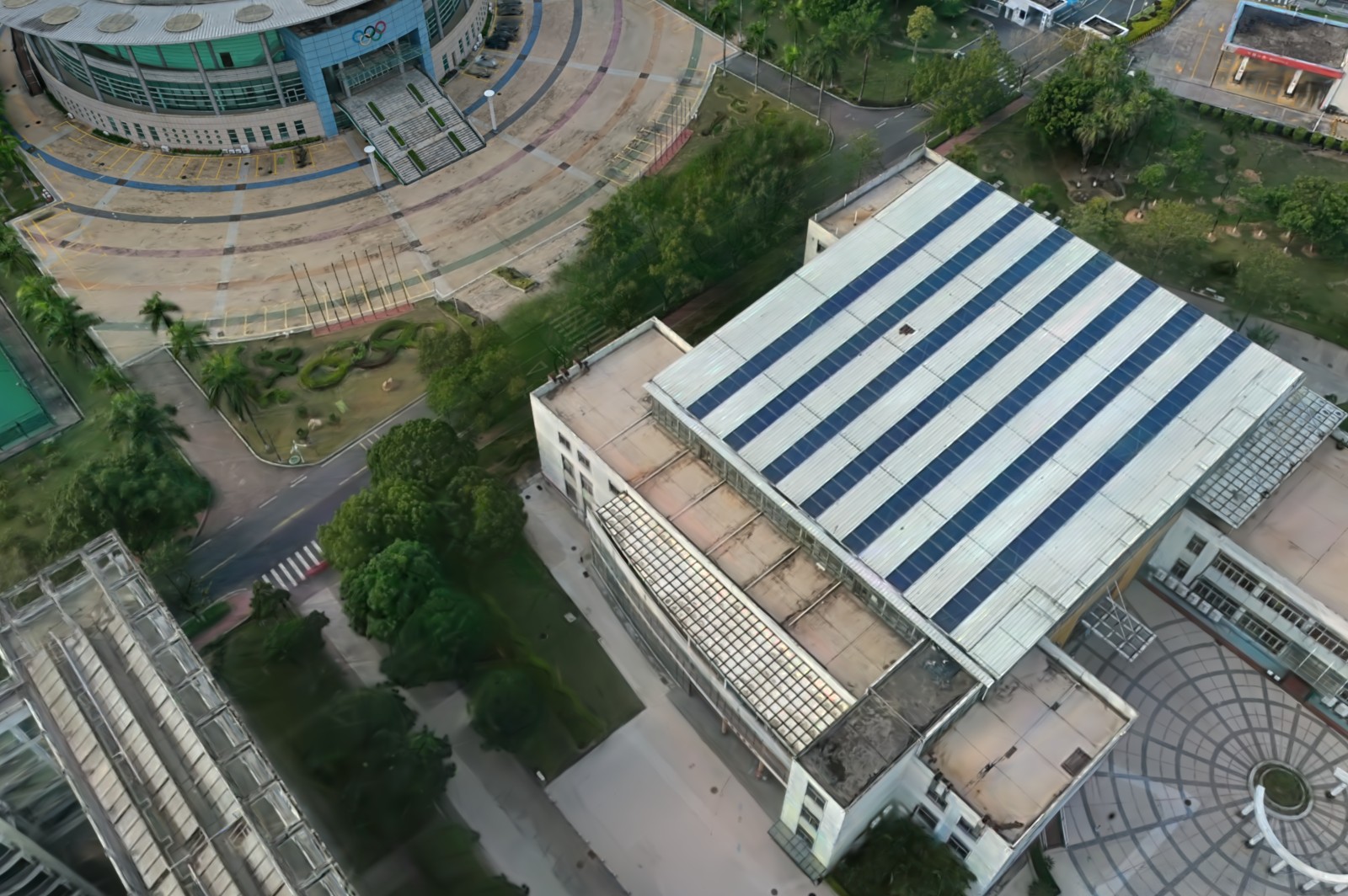}{1.52,2.0}{3,0.7} \\

    \spyimg{0.19\textwidth}{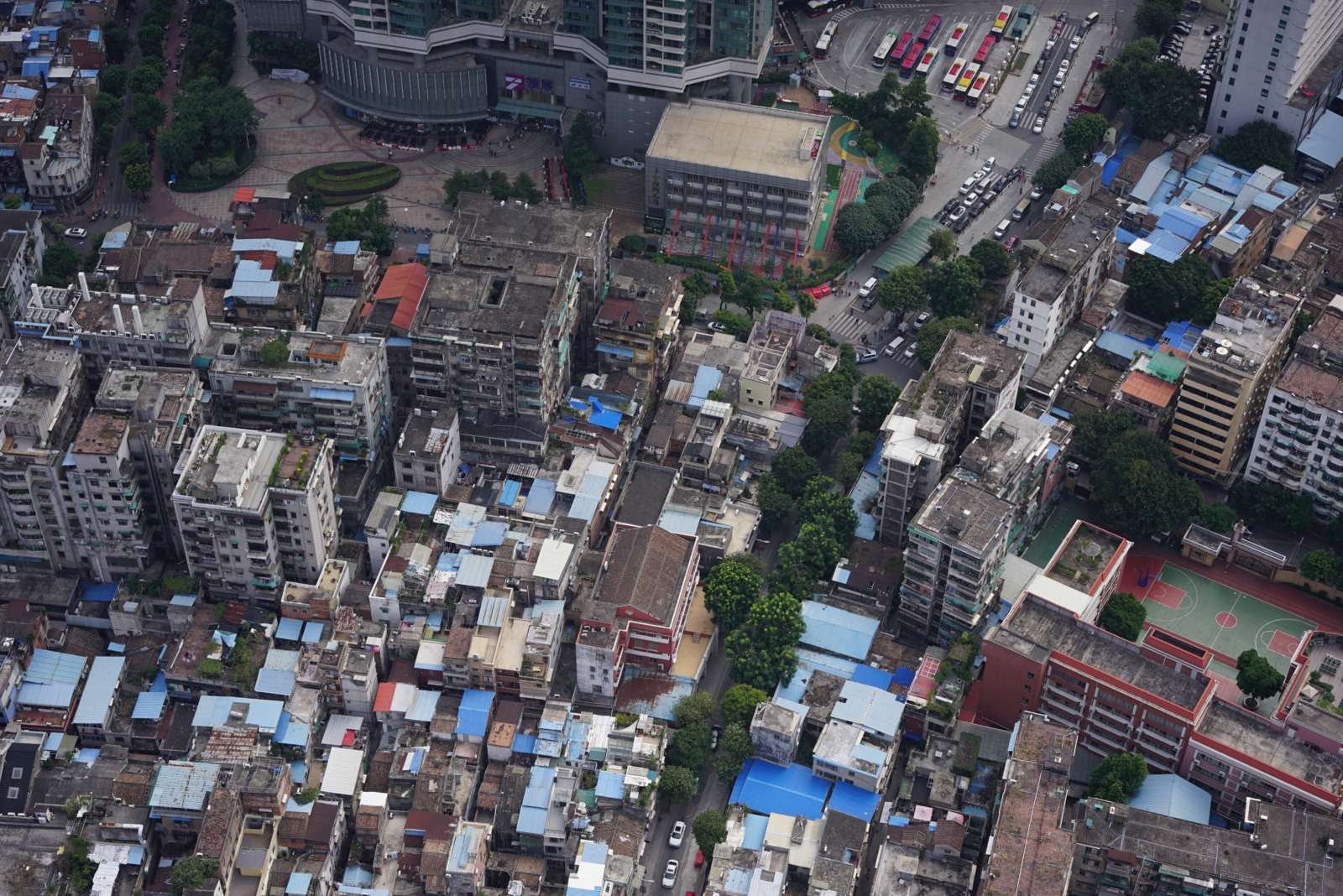}{1.52,1.9}{3,0.7} &
    \spyimg{0.19\textwidth}{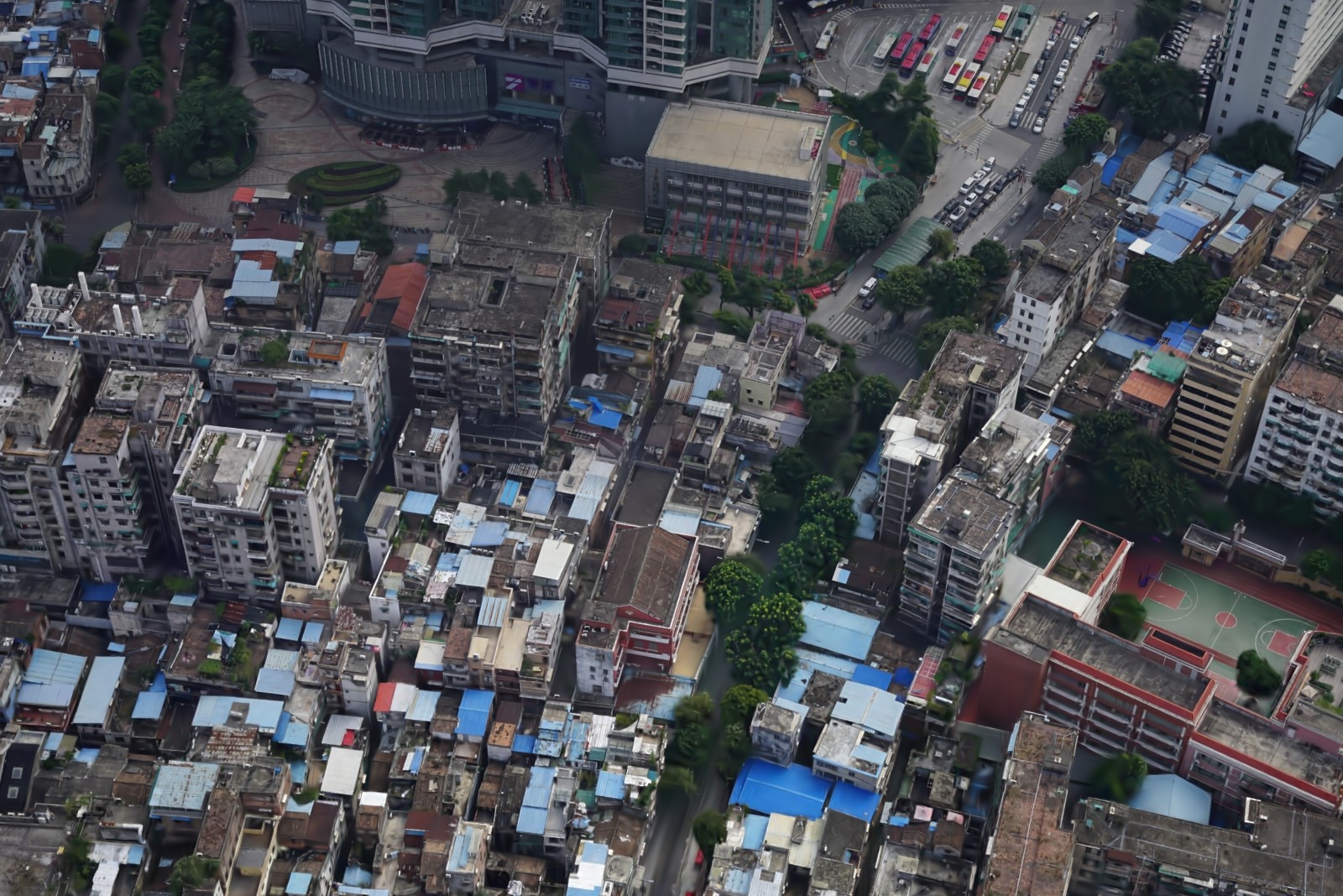}{1.52,1.9}{3,0.7} &
    \spyimg{0.19\textwidth}{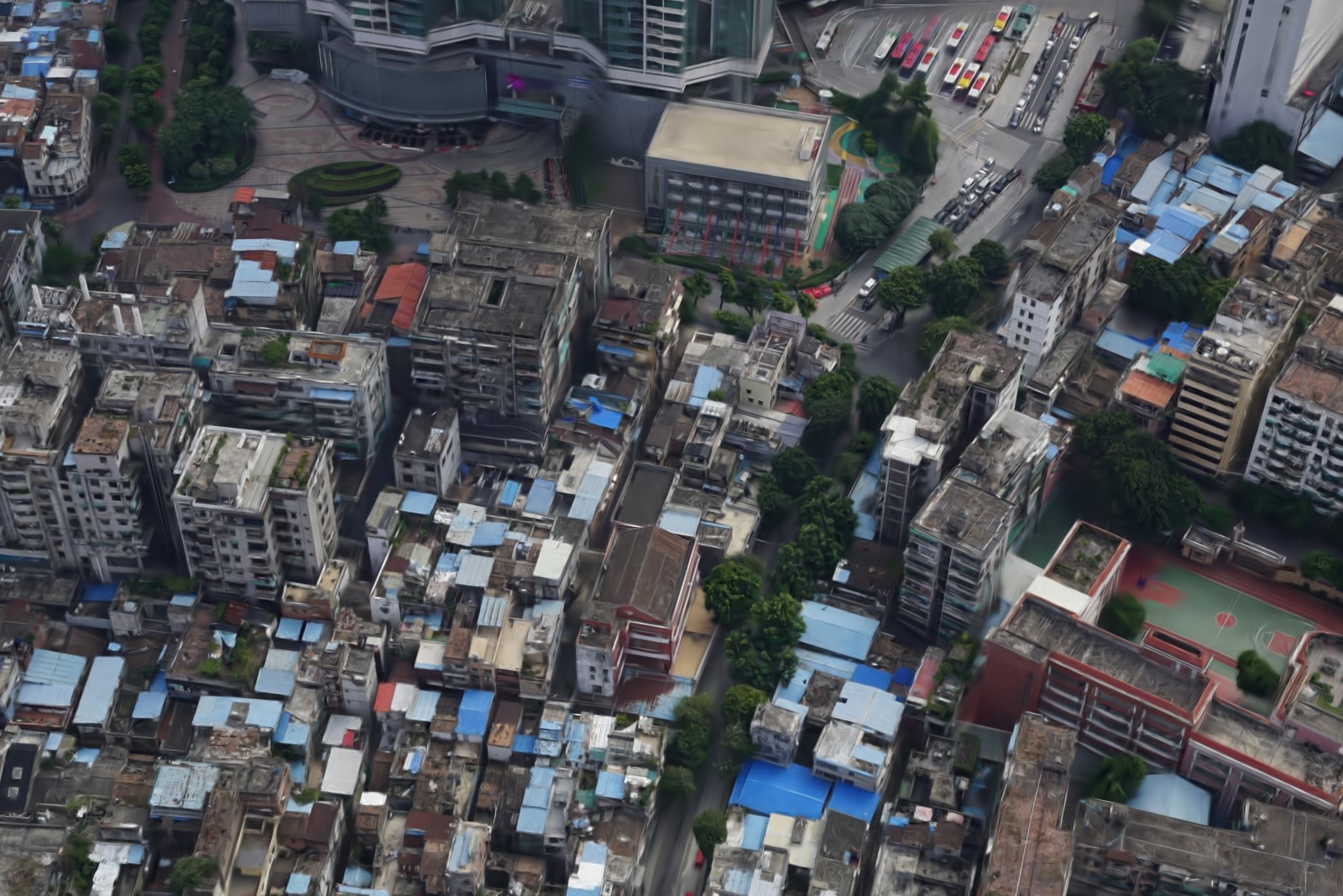}{1.52,1.9}{3,0.7} &
    \spyimg{0.19\textwidth}{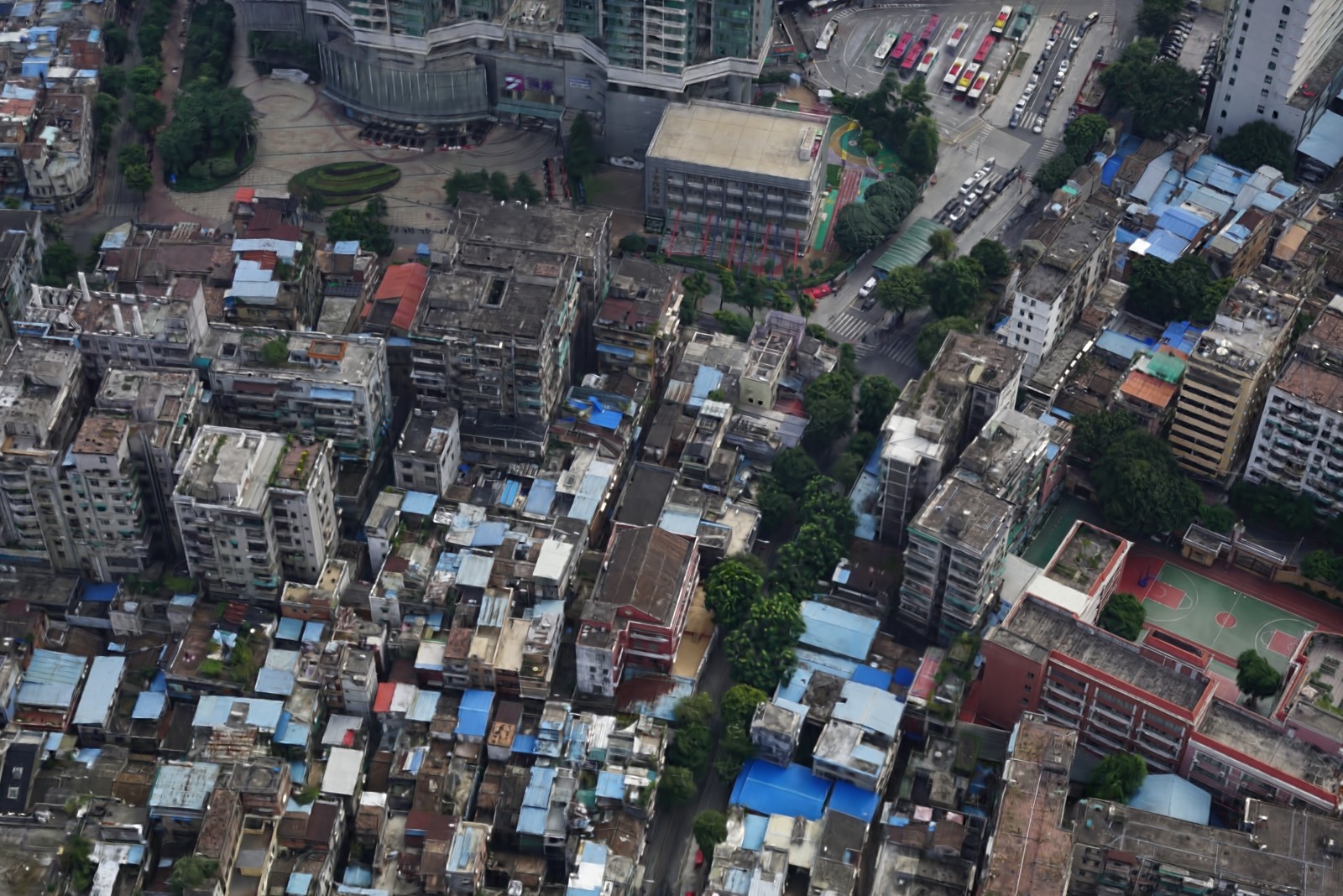}{1.52,1.9}{3,0.7} &
    \spyimg{0.19\textwidth}{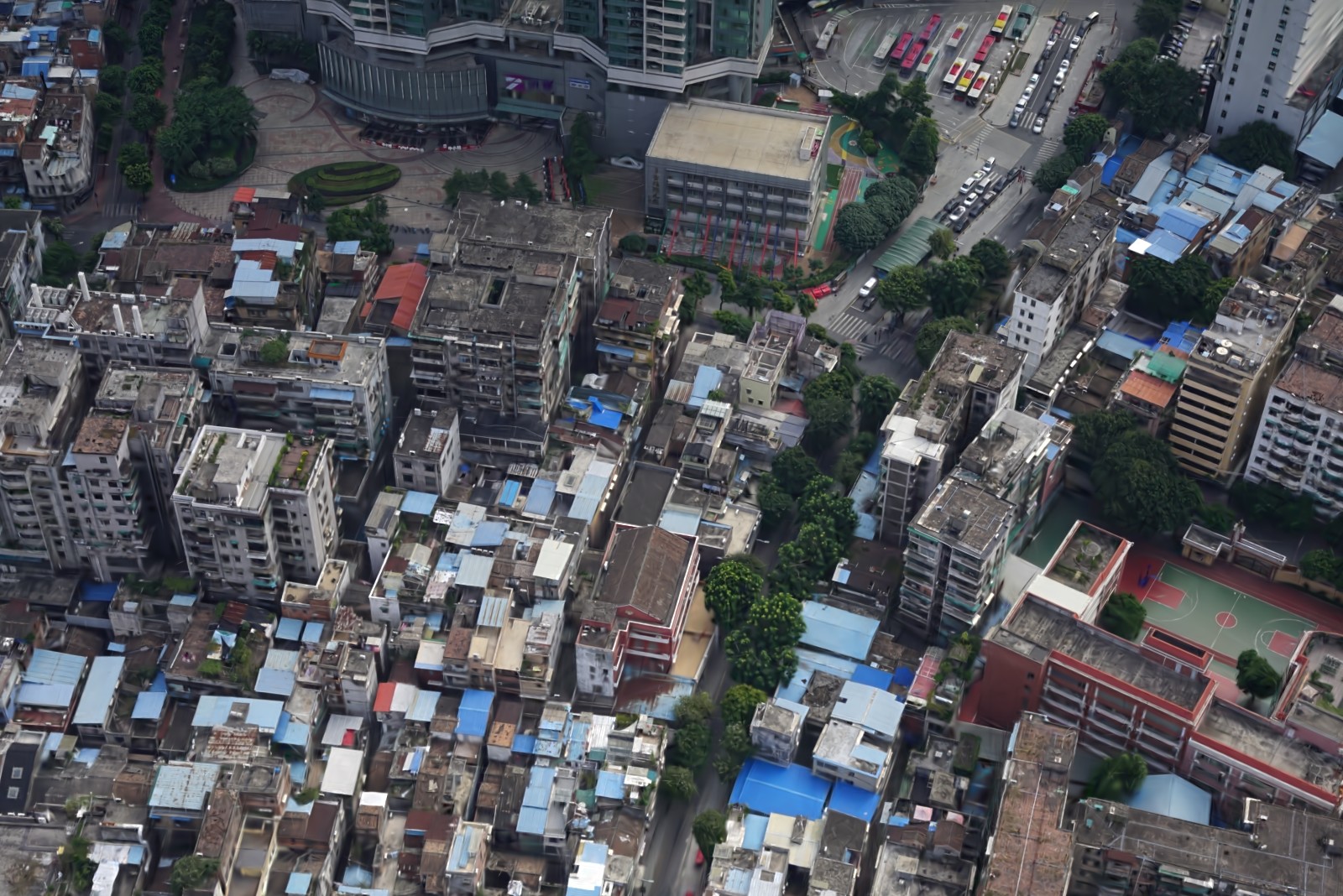}{1.52,1.9}{3,0.7} \\

\end{tabular}
    \caption{\tabletitle{Visualization results.} All methods (excluding 3DGS) render in LOD mode. Ours demonstrates better detail preservation and fewer artifacts.}
    \label{fig:qualitative-comparisons}
    \vspace{-3mm}
\end{figure*}

\section{Experiments}

\subsection{Experimental Setup}

\noindent\textbf{Datasets.} 
Our method is evaluated on three aerial photography scenes, including the \textit{Rubble} scene from Mega-NeRF \cite{turki2022mega} and two self-collected scenes, \ourcampusname{} and \textit{BigCity}. 
Notably, we also conducted validation using \textit{Building} scene from Mega-NeRF \cite{turki2022mega} as well as \textit{Residences}, \textit{Sci-Art} and \textit{Campus} scenes from UrbanScene3D \cite{lin2022capturing}, with results provided in the supplementary materials \ref{subsec:additional_quantitative_comparison}.
The \ourcampusname{} was collected over three months using two different cameras, consisting of approximately 5,000 images and covering an area of about 300,000 $\textrm{m}^2$. This scene exhibits diverse appearance variations due to seasonal changes, weather conditions, and lighting differences. The \textit{BigCity} was captured using five cameras, comprising around 10,000 images and spanning approximately 5 $\textrm{km}^2$. It features a complex road network and a dense distribution of diverse buildings. We use three detail levels for these scenes.

\noindent{}\textbf{Metrics.} We conduct quantitative comparisons of the rendered images using three metrics: PSNR, SSIM, and VGGNet-based LPIPS. To evaluate rendering performance, we measured the FPS and the average number of Gaussians (denoted as \#G, in \(10^6\)) required for rendering the test set at the same resolution, where \#G directly reflects VRAM consumption. 
All methods were evaluated using a single NVIDIA A100-80G GPU.

\subsection{Results}
Our proposed method is compared against four existing methods: Switch-NeRF \cite{zhenxing2022switch}, CityGaussian \cite{liu2024citygaussian}, Hierarchical-3DGS \cite{kerbl2024hierarchical}, and 3DGS \cite{kerbl20233d}.
Quantitative results are summarized in \Cref{tab:compare}.
The first section of the table compares our method, with the LOD mode disabled, against other methods without LOD mode or with it disabled.
For quality-related metrics (SSIM, PSNR, and LPIPS), the results indicate that our method outperforms others. The only exception is the \textit{Rubble} scene, where the LPIPS score matches that of CityGaussian.
The second section of the table evaluates our method with LOD mode enabled. Compared to other LOD-enabled methods, our method consistently outperforms previous approaches across all three quality-related metrics.
Moreover, in most cases, the results with LOD mode enabled surpass the non-LOD results of other methods. This underscores our method’s ability to achieve high-fidelity reconstructions of urban-scale scenes.

Regarding efficiency-related metrics-\#G and FPS, while the \#G in our method is not the smallest, it remains within a reasonable range and supports real-time rendering within 24 GB of memory. Notably, \#G can be further reduced by lowering the budget $B$ of the LOD generation. Meanwhile, the FPS does not experience a significant decline and consistently ranks as either the best or second-best, making real-time rendering entirely feasible. By comparing the results of our method with and without LOD mode, it becomes evident that the number of Gaussians is significantly reduced, leading to a substantial increase in FPS. Meanwhile, the quality experiences only minimal degradation. This can be attributed to our bottom-up detail level generation strategy, allows better preservation of fine-grained geometric and texture details, which is especially beneficial for challenging urban-scale scenes. 
Notably, other methods exhibit a lower \#G in the \textit{BigCity} scene, mainly due to our adjustment of their hyperparameters to ensure execution within 80 GB memory. Further details are provided in Supplementary Material \ref{sec:hparams-of-other-methods}.
The visualization results are shown in \Cref{fig:qualitative-comparisons}, demonstrating that our method achieves superior detail recovery and exhibits a greater ability to eliminate artifacts.

\subsection{LOD Generation}
We conducted experiments on \textit{Rubble} to analyze the impact of different values of the budget $B$ for detail level generation. \Cref{tab:comp-budget} presents the results. As observed, reducing $B$ decreases the number of Gaussians, thereby lowering resource consumption, but at the cost of reconstruction quality. However, increasing $B$ to beyond a certain threshold does not necessarily improve quality, because $B$ only imposes an upper limit, and the scene may not require as many Gaussians as the upper bound allows.

\begin{table}[h!]
    \begin{center}
        \resizebox{0.87\linewidth}{!}{
            \begin{tabular}{l|rrrrr|rrrrr}
                \toprule
                Budget ($\times 100$) &  \tabmetrichead{} \\
                \midrule
                $(1024, 2048, 4096)$ & 0.771 & 26.13 & 0.297 & \textbf{1.61} & \textbf{126.9} \\
                $(2048, 4096, 8192)$ & 0.797 & 26.65 & 0.265 & \underline{2.69} & \underline{108.8} \\
                $(4096, 8192, 16384)$ & \underline{0.814} & \underline{27.03} & \underline{0.245} & 3.60 & 99.7 \\
                $(8192, 16384, 32768)$ & \textbf{0.816} & \textbf{27.11} & \textbf{0.242} & 3.80 & 96.4 \\
                \bottomrule
            \end{tabular}
        }
    \caption{\tabletitle{Quantitative evaluation of budget $B$ for detail level generation.} Adjusting the budget effectively controls resource consumption, but also impacts the quality.}
    \label{tab:comp-budget}
    \vspace{-3mm}
    \end{center}
\end{table}

\subsection{Ablation Study}

\newcommand{\includeablationimage}[1]{\raisebox{-.5\height}{\includegraphics[height=2.2cm]{#1}}}
\newcommand{\ablationspyimg}[4]{%
	\begin{tikzpicture}[spy using outlines={yellow,magnification=3,size=1.3cm, connect spies}]
		\node[anchor=south west,inner sep=0] at (0,0) {\includegraphics[height=2.2cm]{#2}};
		\spy on (#3) in node [left] at (#4);
	\end{tikzpicture}%
}
\newcommand{\includeablationspyimage}[1]{\raisebox{-.5\height}{#1}}
\newcommand{\ablationsubfigurelabel}[1]{
\begin{subfigure}{0.\linewidth}
\caption{}
\label{fig:#1}
\end{subfigure}
}

\begin{figure}[h!]
    \centering
    \resizebox{1.0\linewidth}{!}{
    \begin{tabular}{c@{\raggedright\hspace{1.2pt}}c@{\raggedright\hspace{1.2pt}}c@{\raggedright\hspace{1.2pt}}c@{\raggedright\hspace{1.2pt}}}
        & {\small Appearance} & {\small Depth Reg.} & {\small Scale Reg.}\\
       \rotatebox[origin=c]{90}{Without} & 
       \includeablationspyimage{\ablationspyimg{1.0\textwidth}{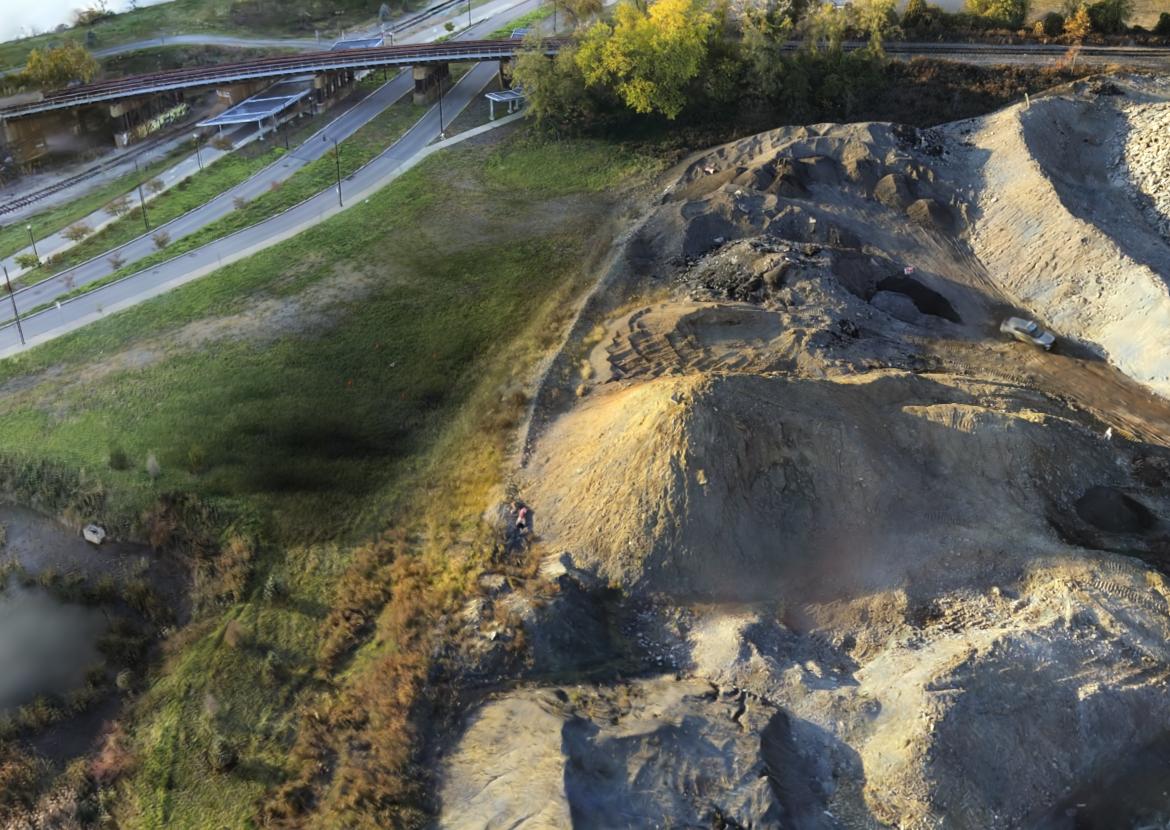}{0.8,1.05}{3,0.7}} & 
       \includeablationimage{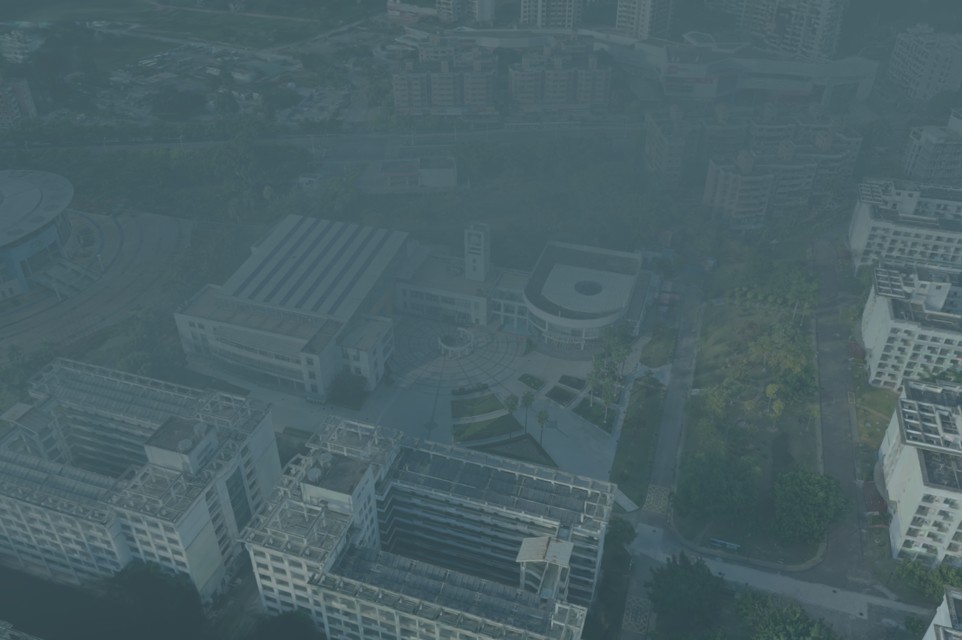} & 
       \includeablationspyimage{\begin{tikzpicture}[spy using outlines={red,magnification=2,size=0.9cm, connect spies}]
		\node[anchor=south west,inner sep=0] at (0,0) {\includegraphics[height=2.2cm]{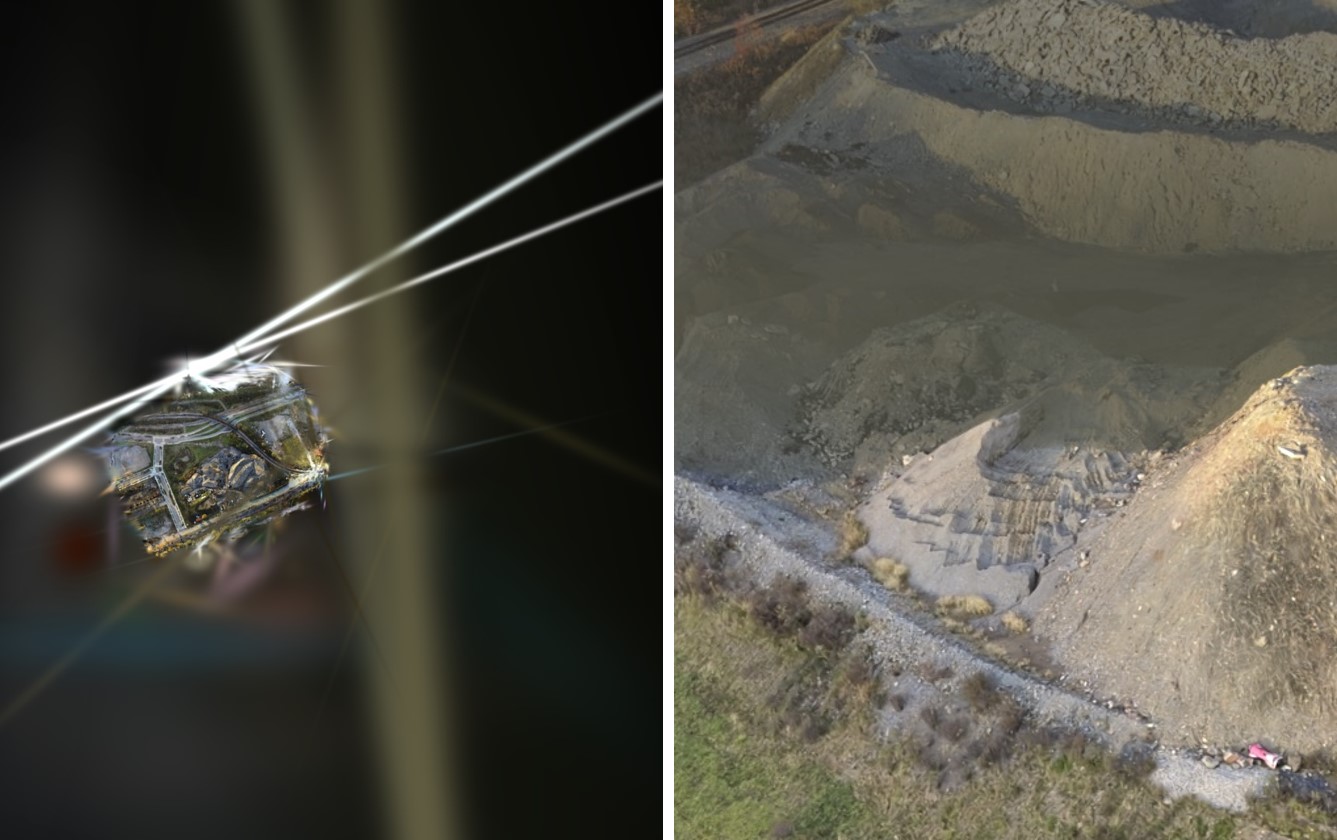}};
            \spy on (0.3,1.1) in node [left] at (2.1,1.7);
            \spy on (2.9,1.1) in node [left] at (2.5,0.5);
	\end{tikzpicture}} \\
       
       \rule{0pt}{33pt} 
       
       \rotatebox[origin=c]{90}{With} & 
       \includeablationspyimage{\ablationspyimg{1.0\textwidth}{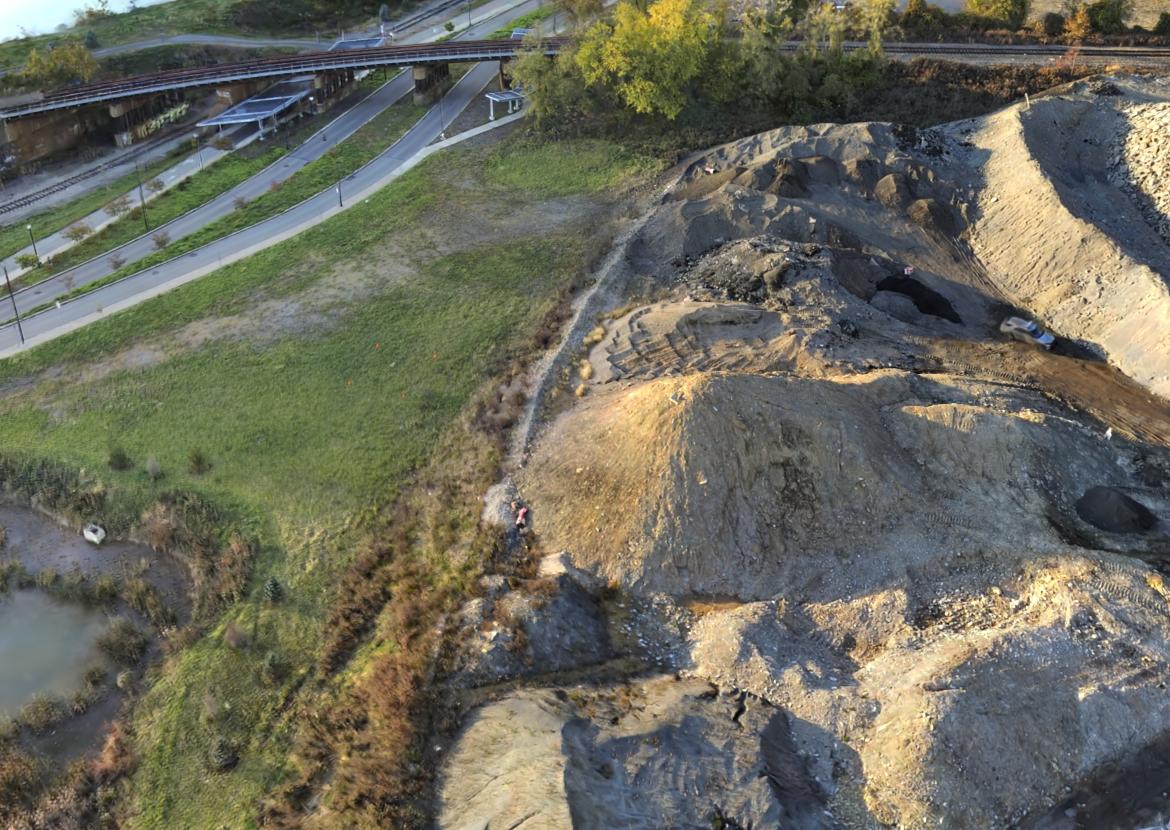}{0.8,1.05}{3,0.7}} & 
       \includeablationimage{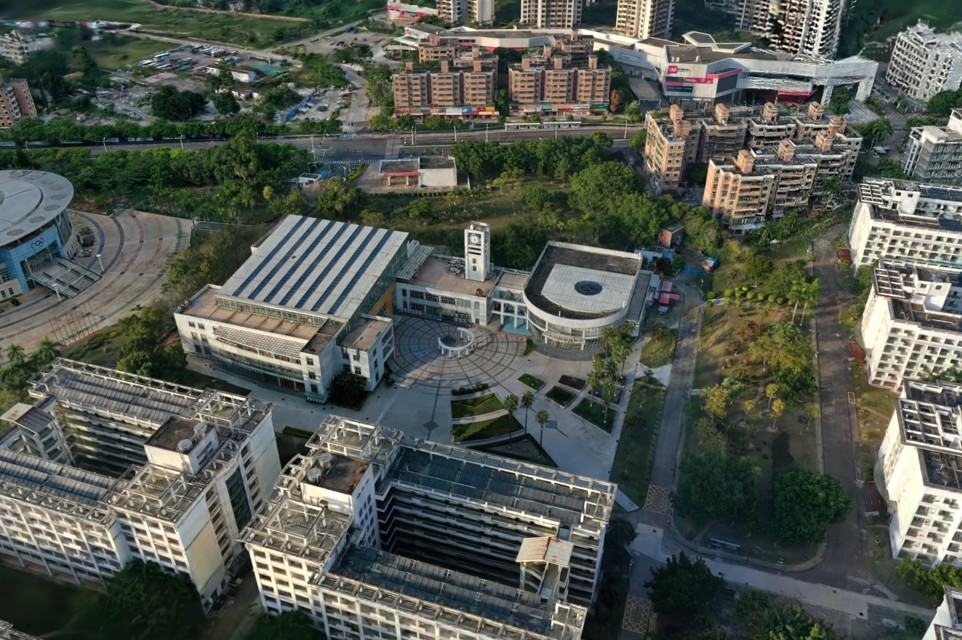} & 
       \includeablationspyimage{\begin{tikzpicture}[spy using outlines={red,magnification=2,size=0.9cm, connect spies}]
		\node[anchor=south west,inner sep=0] at (0,0) {\includegraphics[height=2.2cm]{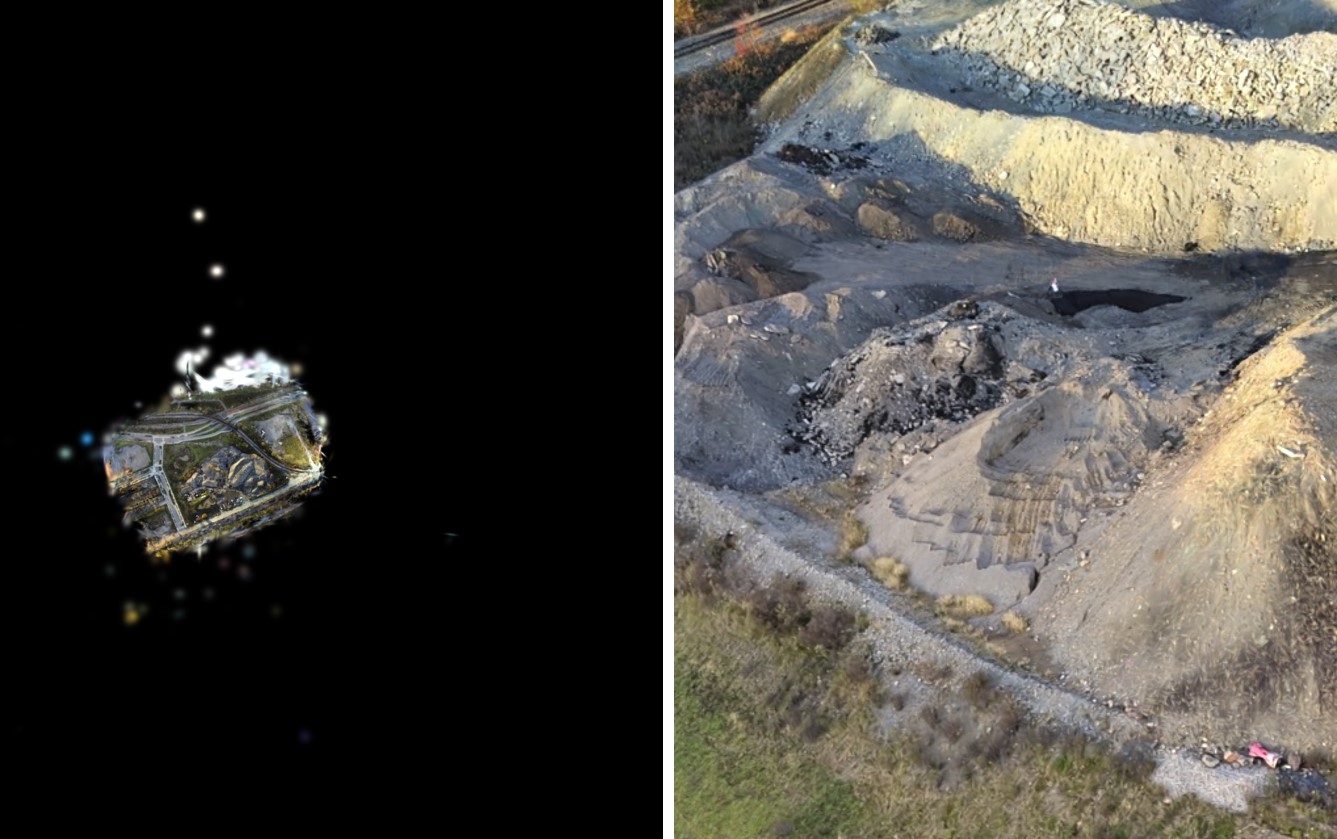}};
            \spy on (0.3,1.1) in node [left] at (2.1,1.7);
            \spy on (2.9,1.1) in node [left] at (2.5,0.5);
	\end{tikzpicture}} \\

       & \ablationsubfigurelabel{ablation-appearance} & \ablationsubfigurelabel{ablation-depth} & \ablationsubfigurelabel{ablation-scale} \\
    \end{tabular}
    }
    \vspace{-3mm}
    \caption{\tabletitle{Visualization results of ablation study.} Our proposed components effectively suppress the artifacts.}
    \label{fig:ablation-summary}
\end{figure}

\begin{table*}[htbp]
	\begin{center}
		\resizebox{\linewidth}{!}{
			\begin{tabular}{l|ccccc|ccccc|ccccc}
				\toprule
				Scene   &   \multicolumn{5}{c|}{\emph{Rubble}}  & \multicolumn{5}{c|}{\emph{\ourcampusname{}}} &   \multicolumn{5}{c}{\emph{BigCity}} \\
				\midrule
				Metrics & \tabmetrichead{} &
				\tabmetrichead{} &
				\tabmetrichead{} \\
				\midrule
                    w/o vis. & 0.803 & 26.95 & 0.256 & 4.01 & 94.3 & 0.809 & 25.14 & 0.242 & 8.53 & 58.6 & 0.826 & 25.74 & 0.238 & 6.92 & \underline{73.7} \\
                    w/o appearance & 0.771 & 25.17 & 0.284 & 4.65 & 82.6 & 0.780 & 22.57 & 0.246 & 8.30 & 54.8 & 0.831 & 25.08 & 0.233 & 6.89 & \textbf{79.5} \\
                    w/o depth reg. & \textbf{0.817} & 26.99 & \textbf{0.240} & \textbf{3.46} & \underline{98.2} & 0.815 & 25.54 & 0.241 & \textbf{6.56} & \underline{63.7} & \textbf{0.845} & \underline{26.40} & \textbf{0.221} & \underline{6.83} & 70.0 \\
                    w/o scale reg. & 0.814 & 26.93 & 0.241 & 3.65 & 89.4 & \textbf{0.824} & \textbf{25.81} & \textbf{0.230} & 7.07 & 56.5 & 0.829 & 25.42 & 0.235 & 6.85 & 68.4 \\
                    w/o in-prior. & \textbf{0.817} & \textbf{27.22} & \textbf{0.240} & 3.90 & 97.5 & 0.815 & \underline{25.72} & 0.241 & 7.90 & 61.2 & \underline{0.840} & 26.33 & \underline{0.228} & \textbf{6.80} & 73.0 \\
                    \midrule
                    full & 0.814 & \underline{27.03} & 0.245 & \underline{3.60} & \textbf{99.7} & \underline{0.816} & 25.71 & \underline{0.240} & \underline{6.65} & \textbf{63.9} & 0.838 & \textbf{26.41} & 0.231 & 6.84 & 73.0 \\
                \bottomrule
			\end{tabular}
		}
		\caption{
        \tabletitle{Qualitative ablations.}
        The results correspond to the removal of point-based visibility, appearance transform module, depth regularization, scale regularization, and in-partition prioritized densification, respectively.
        }
		\label{tab:ablation-compare}
		\vspace{-3mm}
	\end{center}
	\centering
\end{table*}

We conduct ablation experiments
to evaluate the impact of different components of our proposed method. 

\begin{table}[h!]
    \begin{center}
        \resizebox{0.62\linewidth}{!}{
        \begin{tabular}{c|ccc}
            \toprule
            Scene & \textit{Rubble} & \textit{\ourcampusname{}} & \textit{BigCity} \\
            \midrule
             w/o vis. & 1.70$\times$ & 1.46$\times$ & 1.28$\times$ \\
             full & \multicolumn{3}{c}{1$\times$} \\
            \bottomrule
        \end{tabular}
        }
    \caption{\tabletitle{Impact of the point-based visibility on the number of training cameras.} The component effectively reduces camera redundancy.}
    \label{tab:related-training-cameras}
    \vspace{-3mm}
    \end{center}
\end{table}

\noindent\textbf{Point-based Visibility.} The 1st row of \Cref{tab:ablation-compare} presents the results without point-based visibility, meaning that cameras are assigmend to partitions solely based on spatial locations. We follow the setup of Block-NeRF \cite{tancik2022block} and expand the bounding box of each partition by 50\% to define the range for selecting training cameras based on their locations. However, it indicates that a purely position-based selection leads to a suboptimal result. As shown in \Cref{tab:related-training-cameras}, our camera selection strategy can significantly reduce redundant cameras, thus achieving better results under the same number of training iterations.

\noindent\textbf{Appearance Transform Module}. 
The 2nd row of \Cref{tab:ablation-compare} demonstrates the effect of omitting the appearance transform model. This model significantly improves all three quality metrics across all scenes. Additionally, it sometimes noticeably reduces the number of Gaussians, as it prevents 3DGS from introducing extra Gaussians to overfit appearance variations. This, in turn, eliminates artifacts, as shown in \Cref{fig:ablation-appearance}.

\noindent\textbf{Depth Regularization}. The 3rd row of \Cref{tab:ablation-compare} presents the results without depth regularization. Although it does not have a significant impact on metrics, it remains a crucial component. Without it, the model tends to produce discrete floaters that overfit to certain training cameras. As shown in \Cref{fig:ablation-depth}, these floaters often obstruct the scene when rendering from viewpoints that differ significantly from the training cameras, ultimately affecting the visual experience.

\noindent\textbf{Scale Regularization}. As shown in the 4th row of \Cref{tab:ablation-compare}, scale regularization may lead to a slight decline in quality sometimes, because it prevents from generating scale-abnormal Gaussians that fit variations in appearance. But without it will lead to severe artifacts, as shown in \Cref{fig:ablation-scale}.

\begin{table}[btph]
    \resizebox{\linewidth}{!}{
    \centering
    \begin{tabular}{l|cc|cc|cc}
        \toprule
        Scene & \multicolumn{2}{c|}{\textit{Rubble}} & \multicolumn{2}{c|}{\textit{\ourcampusname{}}} & \multicolumn{2}{c}{\textit{BigCity}} \\
        \midrule
        Metrics & Time & \#G & Time & \#G & Time & \#G \\
        \midrule
         w/o in-prior. & 1.49$\times$ & 1.86$\times$ & 1.64$\times$ & 2.05$\times$ & 1.30$\times$ & 1.37$\times$ \\
         full & \multicolumn{6}{c}{1$\times$} \\
        \bottomrule
    \end{tabular}
    }
    \caption{\tabletitle{Impact of in-partition prioritized densification on training time and the number of Gaussians.} It effectively reduces the number of Gaussians while accelerating training.}
    \label{tab:fprior-training-time}
    \vspace{-3mm}
\end{table}

\noindent\textbf{In-Partition Prioritized Densification}. The 5th row of \Cref{tab:ablation-compare} shows that the our densification strategy has only a minimal impact on overall quality, particularly with respect to the SSIM and LPIPS metrics, where enabling or disabling it yields only negligible differences. However, as illustrated in \Cref{tab:fprior-training-time}, it markedly accelerates the training process.

\section{Conclusion}

We propose a robust and efficient 3D Gaussian Splatting method tailored for urban-scale scene reconstruction. Our scene partitioning and visibility-based image selection enable scalable reconstruction within limited resources. The controllable LOD strategy provides precise resource regulation and real-time rendering. Additionally, our fine-grained appearance transform module and scale regularization significantly enhance visual consistency and flexibility. Extensive experiments demonstrate our method’s superior reconstruction quality, efficiency, and practical applicability to large-scale urban scenes.

Our method currently relies on accurate camera poses obtained from external SfM methods. Inaccurate or noisy poses can negatively impact reconstruction quality, especially in large-scale urban scenes. Enhancing robustness to pose inaccuracies is thus an important future direction. Additionally, our LOD switching mechanism currently is not incremental, increasing storage and computational overhead. Future work could explore incremental switching mechanisms for smoother transitions and improved resource efficiency.

\vspace{-4mm}\paragraph{Acknowledgments.}
This work was supported in part by the Science and Technology Development Fund, Macau SAR (Grants No. 0087/2022/AFJ and No. 001/2024/SKL), in part by the National Natural Science Foundation of China (Grant No. 62261160650), in part by the Research Committee of University of Macau (Grant No. MYRG-GRG2023-00116-FST-UMDF), and in part by the  the Fundamental Research Funds for the Central Universities (Grant No. 21625360).

{
    \small
    \bibliographystyle{ieeenat_fullname}
    \bibliography{main}
}

\clearpage
\setcounter{page}{1}
\maketitlesupplementary

\appendix

\section{Our Datasets}
\begin{figure*}[htbp]
    \begin{center}
        \begin{subfigure}[b]{0.49\textwidth}
            \includegraphics[width=1.0\textwidth]{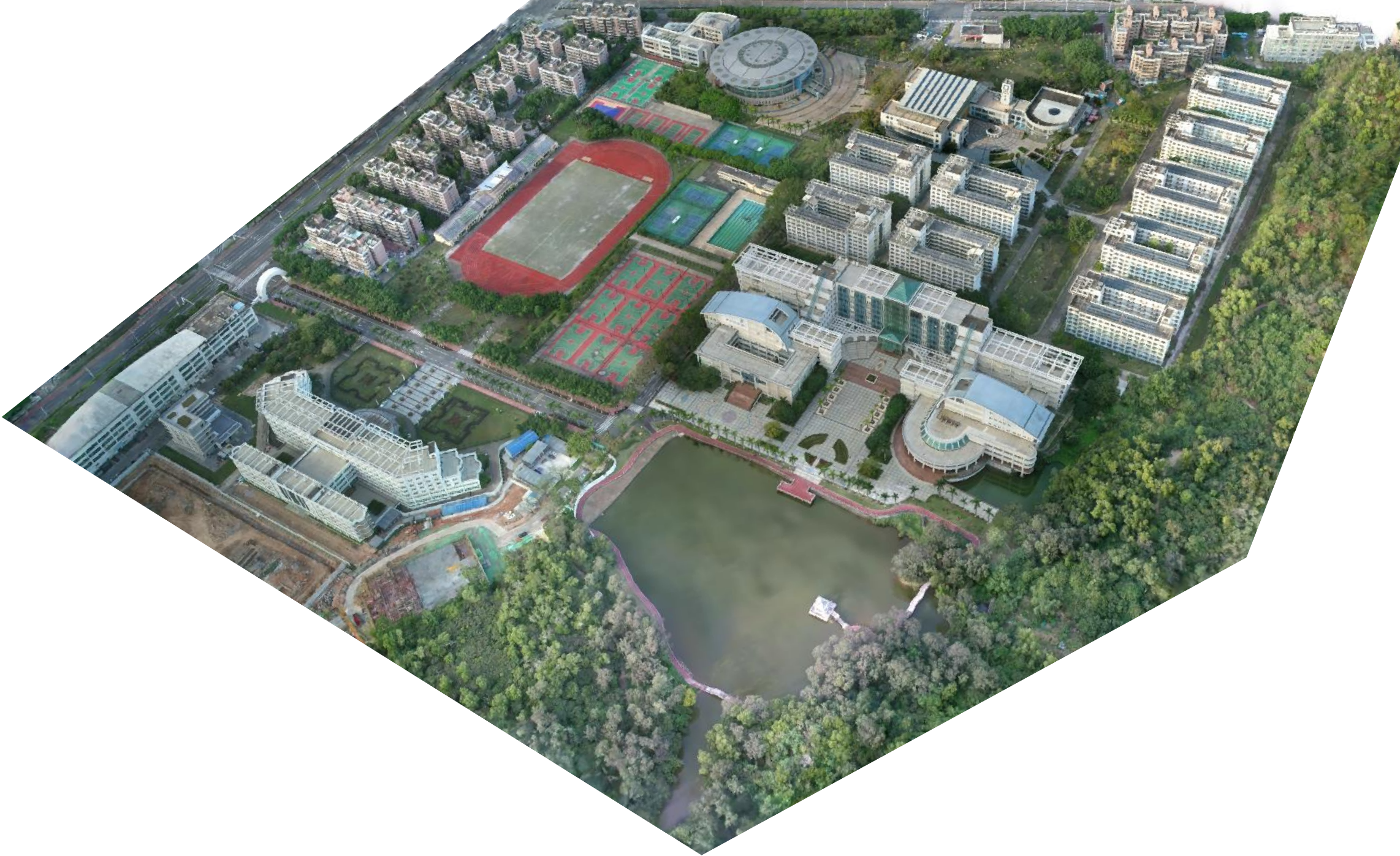}
            \caption{\ourcampusname{}}
        \end{subfigure}
        \begin{subfigure}[b]{0.49\textwidth}
            \includegraphics[width=1.0\textwidth]{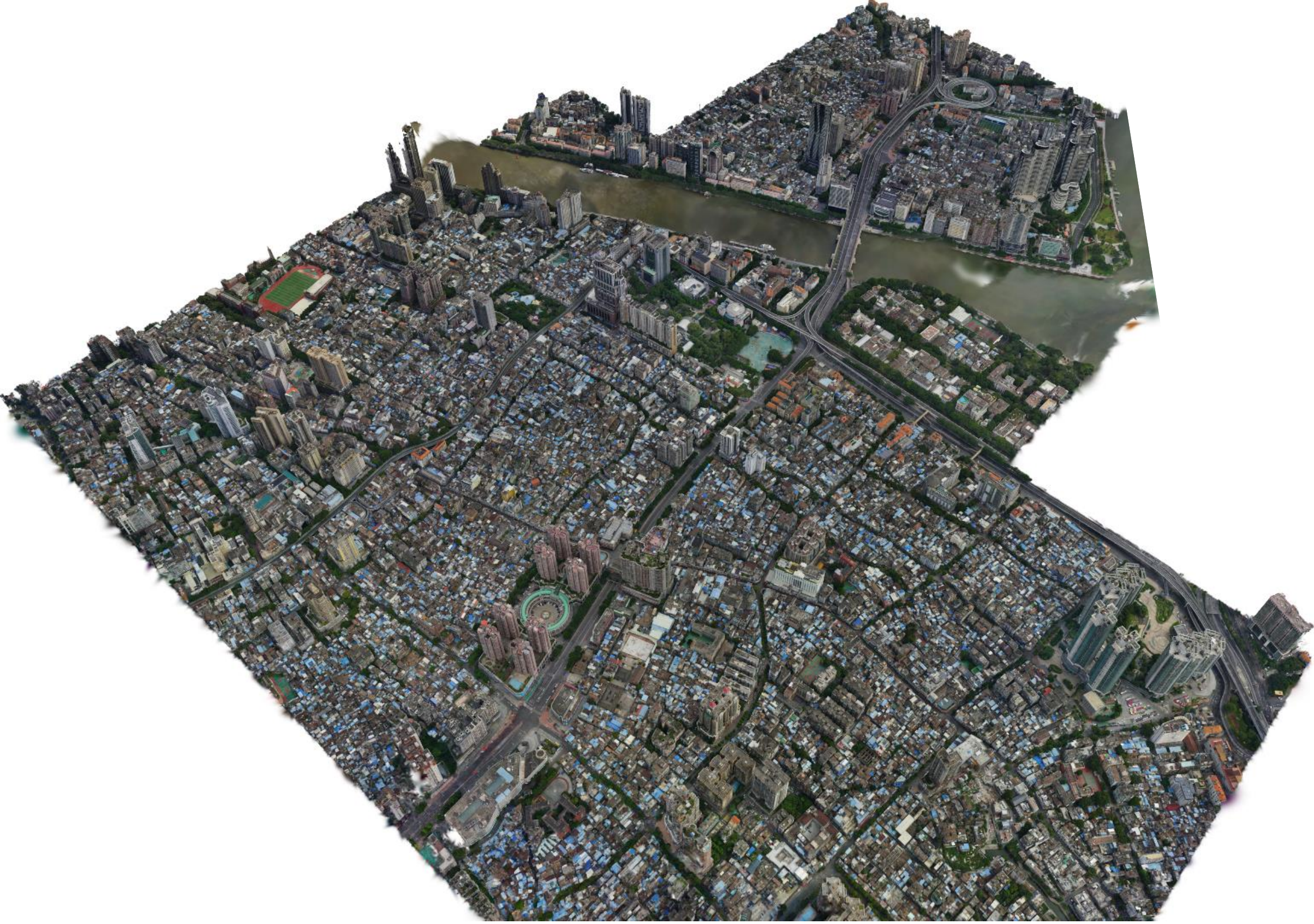}
            \caption{\textit{BigCity}}
        \end{subfigure}
    \end{center}
    \caption{\tabletitle{Our datasets: \ourcampusname{} and \textit{BigCity}.}}
    \label{fig:our-datasets}
\end{figure*}

The \ourcampusname{} and \textit{BigCity} scenes were collected by our team using drones, and their contents are shown in \Cref{fig:our-datasets}.
We employ COLMAP's hierarchical SfM \cite{schonberger2016structure} to perform sparse reconstruction for both scenes. After finishing reconstruction, we use COLMAP's geo-registration to align the reconstructed model with GPS coordinates. Subsequently, we compute the Euclidean distance between the estimated camera positions and their corresponding GPS coordinates. Outliers with excessively large distances are discarded, as they typically result from inaccurate pose estimations. This filtering process helps mitigate the negative impact of erroneous data. Finally, we downsample the images to a maximum edge length of 1600 pixels for experimentation. When partitioning, the sizes used for these two scenes are 180m and 400m, respectively.

\section{Implementation Details of Our Method}
\label{sec:imp-details}
We implemented our method based on gsplat \cite{ye2024gsplat}, which offers higher computational and memory efficiency compared to the \cite{kerbl20233d}.

The visibility threshold for dataset division is $1/6$. 
We use three detail levels for all scenes. 
The first and second levels each last a base of 15,000 iterations, with densification enabled. The third level runs for a base of 30,000 iterations, where densification is applied in the first half and the second half is solely dedicated to optimizing properties.
\Cref{tab:lod-generation-hyperparameter} presents the hyperparameters for detail level generation in different scenes.

\begin{table}[htbp]
    \begin{center}
    \resizebox{\linewidth}{!}{
    \begin{tabular}{l|ccc}
        \toprule
       Scenes & $\left(\denseb_{1}, \denseb_{2}, \denseb_{3}\right) \times 100$ & $(\densei_{1}, \densei_{2}, \densei_{3})$ & $(\densed_{1}, \densed_{2}, \densed_{3})$ \\
       \midrule
       \textit{Rubble} & $(4096, 8192, 16384)$ & \multirow{3}{*}{$(300, 200, 100)$} & $\left(1/2, 1/3, 1\right)$ \\
       \ourcampusname{} & $(4096, 8192, 20480)$ & & $\left(1/4, 1/2, 1\right)$ \\
       \textit{BigCity} & $(2048, 8192, 20480)$ & & $\left(1/4, 1/2, 1\right)$ \\
       \bottomrule
    \end{tabular}
    }
    \caption{\tabletitle{The hyperparameters for the detail level generation.}}
    \label{tab:lod-generation-hyperparameter}
    \vspace{-3mm}
    \end{center}
\end{table}

In practice, these iteration counts and densification interval $\densei$ are adjusted proportionally based on the number of images $N$ in each partition, scaled by a factor of $\max(N/600, 1)$.

In the appearance transform model, the MLP consist of 1 hidden layers with 32 channels, followed by a ReLU activation. The output layer followed by a sigmoid activation. The Gaussian embeddings is 16-dimensional, while the image embedding is 32-dimensional. The initial learning rate for the MLP and embeddings is set to 0.01, and an exponential decay scheduler reduces it to a final value of 0.00025. Every 50 iterations, we sample 20,480 Gaussians and select $k=16$ nearest neighbors to perform similarity regularization, minimizing the computational overhead.

In the scale regularization, the value of $s_{\textrm{max}}$ is set to a value corresponding to the typical size of most buildings in the scene, and $r_{\textrm{max}}=10$.

In the in-partition prioritized densification, The value of $\hat{d}_{\textrm{max}}$ is identical to the partition size. The maximum gradient threshold factor is $\eta=4$. The minimum threshold is $\tau_\textrm{min} = 0.0002$ for the 1st and 2nd levels, and is $0.6$ for the 3rd level with AbsGS \cite{ye2024absgs} enabled.

\section{Hyperparameters of Other Methods}
\label{sec:hparams-of-other-methods}
For the 3DGS, large-scale scenes generally require more iterations for sufficient optimization. Therefore, the training process was extended to 50 epochs, with densification enabled during the first 25 epochs. We also set the densification interval to $1/6$ of an epoch, ensuring a consistent number of densifications across all scenes. When the number of input images is 600, these adjustments yield consistent hyperparameter with the original settings.

For Switch-NeRF, we utilized the official open-source implementation with its provided hyperparameters. When conducting experiments on our own scenes, we proportionally increased the number of training iterations based on the number of input images.

For the remaining methods, we utilized their official open-source implementations and use a similar number of partitions to reconstruct all the scenes.
When evaluating the LOD mode of Hierarchical-3DGS, we used a granularity setting value of 6 pixels.

Due to the large scale and intricate details of the \textit{BigCity} scene, none of the previous 3DGS-based methods can complete the experiment within an 80GB memory limit. Therefore, we made additional adjustments to the hyperparameters for these methods. For 3DGS, we double the densification gradient threshold. For CityGaussian, during the coarse training stage, we tripled the densification cycle and doubled the densification gradient threshold compared to the original settings. During the pruning stage, we increased the pruning ratios from the default 40\%, 50\%, and 60\% to 70\%, 80\%, and 90\%. For Hierarchical-3DGS, the excessive number of Gaussians made it impossible to evaluate the non-LOD mode. When evaluating its metrics under the LOD mode, we doubled the granularity settings from 6 pixels to 12 pixels. In contrast, our method can complete all steps, except for the non-LOD mode, with memory usage not exceeding 24GB.

\section{Appearance Transform Module}
\subsection{Metric Calculation}
Given that we have the appearance transform model, which optimizes only the embeddings of training set images, we followed a strategy similar to NeRF-W \cite{martin2021nerf} to evaluate the test set images. Specifically, when computing the metrics for test images, we first optimize the image embedding $\imageembedding$ using the left half of the image and compute the metrics using the right half. 
Each partition is transformed using the embedding of the test image optimized within that partition, ensuring appearance consistency. 
Then, we optimize the embedding from scratch using the right half and computed the metrics with the left half. Finally, the average of the results from both rounds was taken as the final metric value for the entire image. This approach prevents information leakage and ensures fairness in the evaluation process. In practice, we further smooth transitions between partitions via weighted averaging.
\subsection{Appearance Transformation}

\begin{figure*}[htbp]
    \begin{center}
        \begin{subfigure}{0.39\linewidth}
            \begin{tikzpicture}
                \node [anchor=south west, inner sep=0] (main) at (0,0) {\includegraphics[width=1.\linewidth]{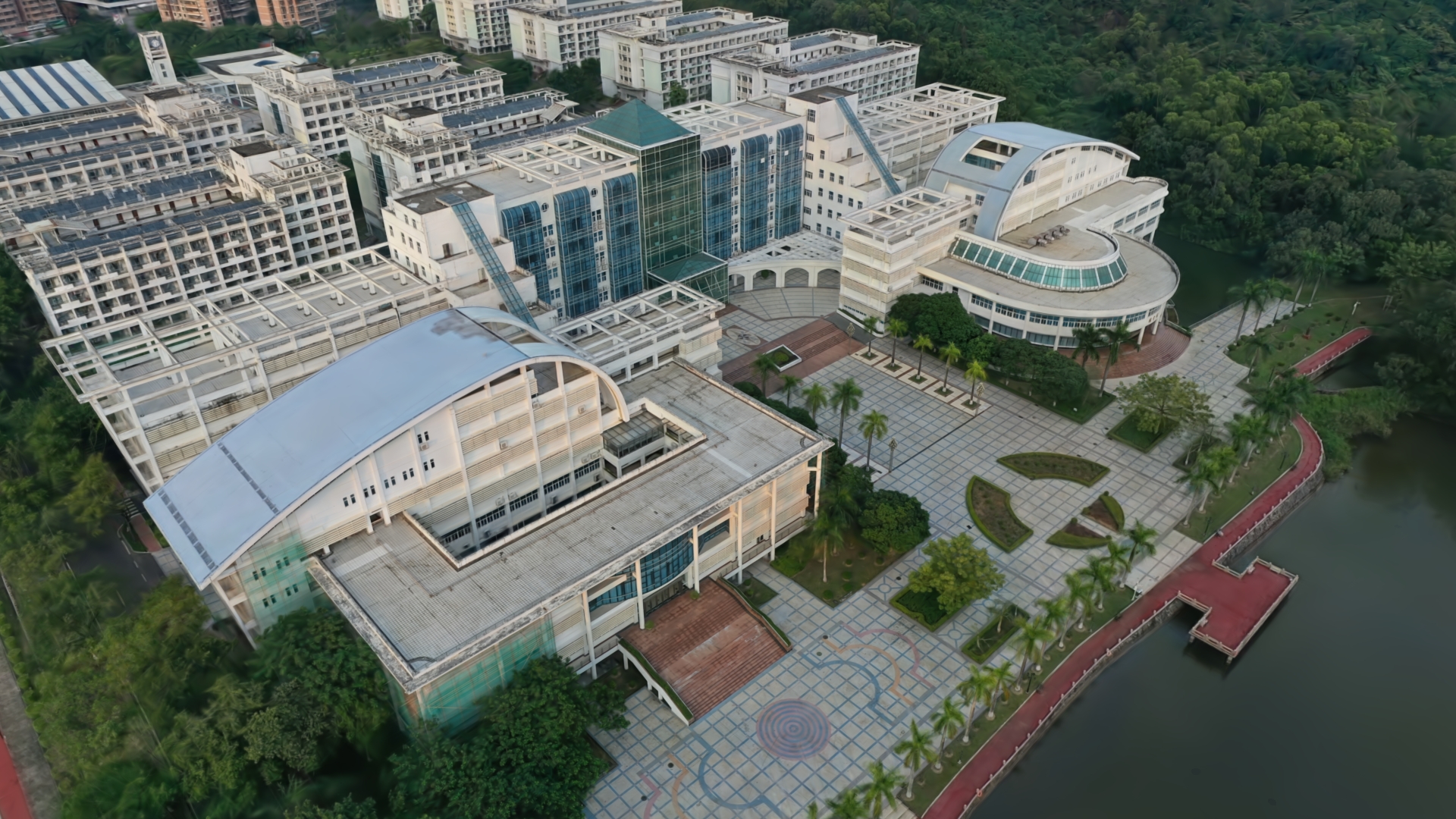}};
                
                \node [anchor=south east, xshift=3.6, yshift=-3.6] at (main.south east) 
                {
                    \begin{tikzpicture}
                        \node[inner sep=0] (img) {\includegraphics[width=0.4\textwidth]{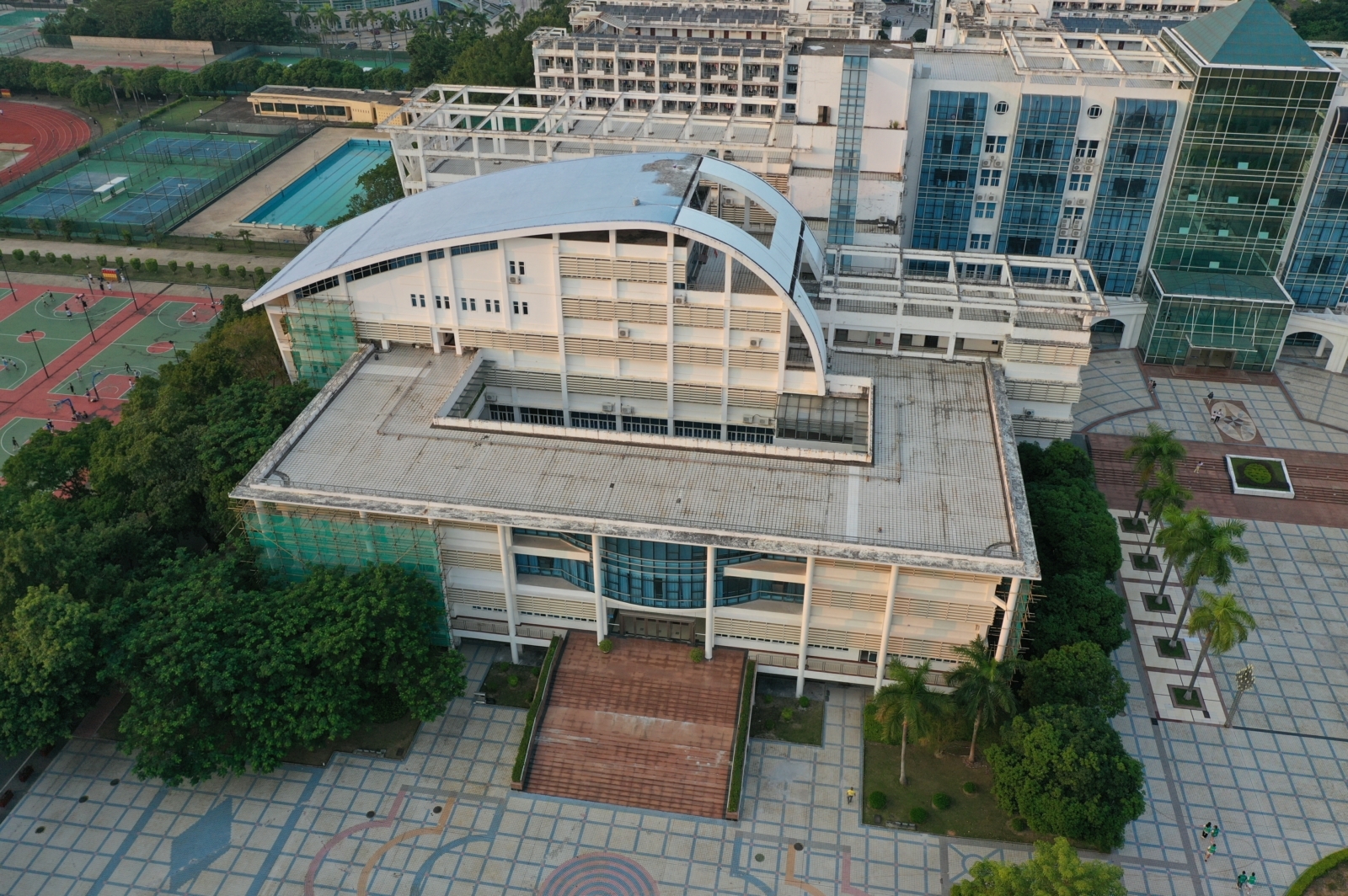}};
                        \draw[red, thick] (img.south west) rectangle (img.north east);
                    \end{tikzpicture}
                };
            \end{tikzpicture}
            \caption{Under maintenance, illuminated by the sunset.}
        \end{subfigure}
        \begin{subfigure}{0.39\linewidth}
            \begin{tikzpicture}
                \node [anchor=south west, inner sep=0] (main) at (0,0) {\includegraphics[width=1.\linewidth]{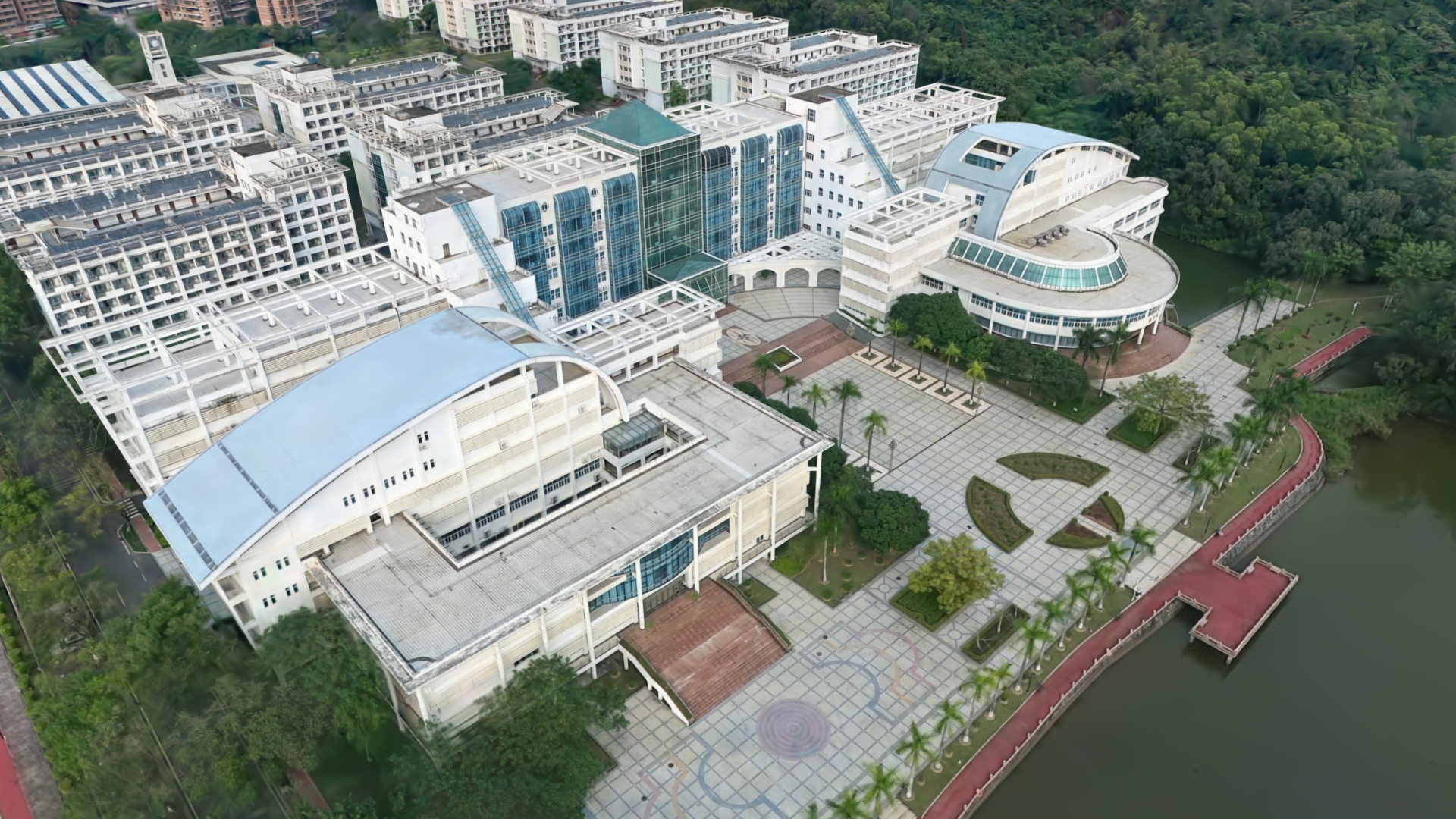}};
                
                \node [anchor=south east, xshift=3.6, yshift=-3.6] at (main.south east) 
                {
                    \begin{tikzpicture}
                        \node[inner sep=0] (img) {\includegraphics[trim={0 15cm 30cm 5cm},clip,width=0.4\textwidth]{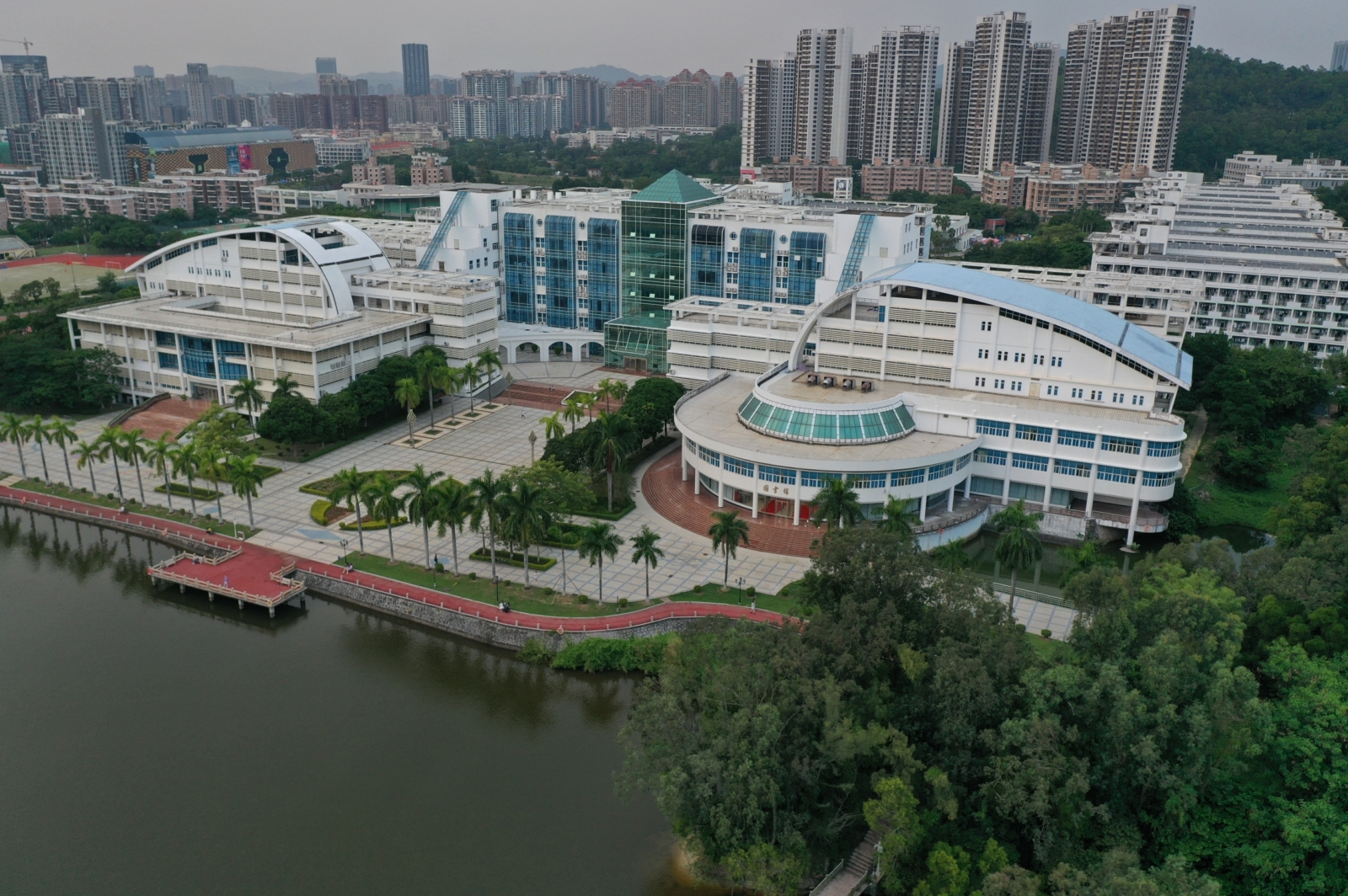}};
                        \draw[red, thick] (img.south west) rectangle (img.north east);
                    \end{tikzpicture}
                };
            \end{tikzpicture}
            \caption{Maintenance completed, illuminated by overcast light.}
        \end{subfigure}
    \caption{\tabletitle{Synthesis the two corresponding states from a new viewpoint based on the embedding vector provided by the reference image (bottom right).}}
    \label{fig:exp-building-appearance-transform}
    \end{center}
\end{figure*}
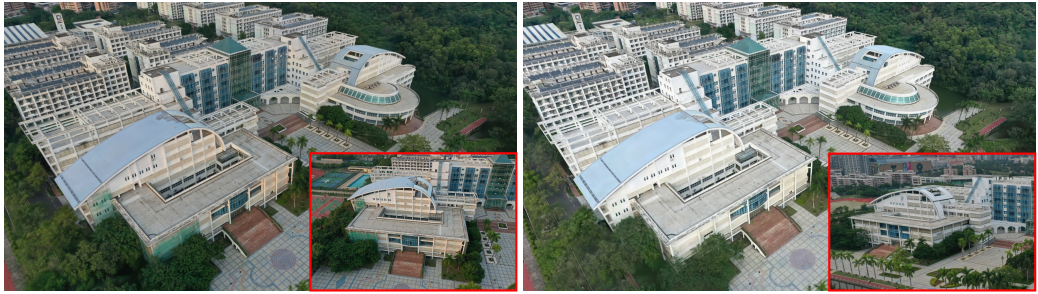

After reconstruction, our method enables scene appearance transformation. Using the image embedding $\imageembedding$ of a training image, we can synthesize novel views that match its appearance. As shown in \Cref{fig:exp-building-appearance-transform}, this enables transforming between different states of a building in the \ourcampusname{} scene.

\section{Additional Experiments}
\subsection{Training Time Comparison}
\Cref{tab:training-time} presents the training times of all 3DGS-based methods. Except for 3DGS, all results were obtained under parallel training setups. The results show that our method is also competitive in terms of training efficiency, consistently ranking among the best or second-best. We are not consistently the fastest due to the additional overhead introduced by the Appearance Transform Module, anti-aliasing, and various regularization mechanisms. Nonetheless, maintaining such competitive efficiency despite these added components demonstrates the effectiveness of our optimization strategies.

It is worth noting that the \textit{Rubble} and \ourcampusname{} are relatively small, where parallel training provides limited benefits. In contrast, the \textit{BigCity} is significantly larger, and the parallel setup leads to a substantial speedup.
\begin{table}[htbp]
	\begin{center}
		\resizebox{0.75\linewidth}{!}{
			\begin{tabular}{l|rrr}
				\toprule
				Scenes &  \textit{Rubble} & \ourcampusname{} & \textit{BigCity} \\
				\midrule
                    3DGS & 1.45 & 3.21 & 67.39 \\
                    CityGaussian & 2.33 & 2.49 & 4.01 \\
                    Hierarchical-3DGS & \textbf{1.00} & \underline{2.18} & \textbf{1.71} \\
                    Ours & \underline{1.30} & \textbf{2.14} & \underline{2.01} \\
                    \bottomrule
			\end{tabular}
		}
		\caption{\tabletitle{Comparison of training time (in hours).} Except for 3DGS, the results of all other methods were obtained under parallel training mode. VastGaussian is not included as it is not open-sourced.}
		\label{tab:training-time}
		\vspace{-3mm}
	\end{center}
	\centering
\end{table}

\subsection{Additional Quantitative Comparison}
\label{subsec:additional_quantitative_comparison}

\Cref{tab:additional-quan-comp} presents experimental results for the \textit{Building} \cite{turki2022mega}, \textit{Residence}, \textit{Sci-Art} and \textit{Campus} \cite{lin2022capturing} scenes. 
The camera poses are provided by Mega-NeRF. Overall, our method demonstrates a clear advantage in nearly all quality-related metrics.
Although our method is not the most optimal in terms of resource consumption and rendering speed, it remains within a reasonable range and is close to the best-performing approach. It fully ensures real-time rendering. It is worth noting that our method can further reduce resource consumption by lowering the budget $B$.
\Cref{fig:additional-qualitative-comparisons} presents the visualization results for both scenes, demonstrating that our method achieves higher detail preservation and fewer artifacts.
\begin{table*}[htbp]
    \begin{center}
    \resizebox{1.0\linewidth}{!}{
    \begin{tabular}{l|rrrrr|rrrrr|rrrrr|rrrrr}
    \toprule
    Scene   &   \multicolumn{5}{c|}{\emph{Building}} & \multicolumn{5}{c|}{\emph{Residence}} & \multicolumn{5}{c|}{\emph{Sci-Art}} & \multicolumn{5}{c}{\emph{Campus}} \\
    \midrule
    Metrics &  \tabmetrichead{} &
    \tabmetrichead{} &
    \tabmetrichead{} &
    \tabmetrichead{} \\
    \midrule

    Switch-NeRF & 0.579 & 21.54 & 0.474 & -- & $<$0.1 & 0.654 & 22.57 & 0.457 & -- & $<$0.1 & 0.795 & 26.52 & 0.360 & -- & $<$0.1 & 0.541 & 23.62 & 0.609 & -- & $<$0.1 \\
    VastGaussian & \underline{0.804} & \underline{23.50} & -- & -- & -- & \textbf{0.852} & \underline{24.25} & -- & -- & -- & \textbf{0.885} & \underline{26.81} & -- & -- & -- & \textbf{0.816} & \underline{26.00} & -- & -- & -- \\
    CityGaussian (no LOD) & 0.784 & 21.96 & \underline{0.243} & \underline{13.30} & 37.6 & 0.813 & 22.00 & \underline{0.211} & \underline{10.80} & 41.0 & 0.837 & 21.39 & \underline{0.230} & \underline{3.80} & 82.3 & 0.666 & 19.61 & 0.403 & \underline{16.41} & \underline{35.1} \\
    Hierarchical-3DGS (no LOD) & 0.720 & 20.55 & 0.270 & 14.79 & 36.4 & 0.753 & 19.85 & 0.230 & 13.68 & 39.9 & 0.792 & 19.85 & 0.273 & 9.13 & 31.5 & 0.741 & 22.66 & \underline{0.297} & 29.32 & 13.9 \\
    3DGS & 0.787 & 22.42 & 0.282 & \textbf{13.02} & \textbf{64.2} & 0.807 & 21.96 & 0.256 & \textbf{7.06} & \textbf{102.2} & 0.833 & 21.26 & 0.285 & \textbf{2.28} & \textbf{172.4} & 0.718 & 19.83 & 0.370 & \textbf{10.55} & 18.5 \\
    Ours (no LOD) & \textbf{0.808} & \textbf{24.12} & \textbf{0.219} & 18.22 & \underline{62.5} & \underline{0.845} & \textbf{24.93} & \textbf{0.201} & 14.15 & \underline{71.5} & \underline{0.876} & \textbf{27.78} & \textbf{0.190} & 8.39 & \underline{86.9} & \underline{0.788} & \textbf{26.63} & \textbf{0.278} & 42.42 & \textbf{38.5} \\
    \midrule
    CityGaussian & \underline{0.769} & \underline{21.75} & \underline{0.257} & \textbf{3.49} & \textbf{83.6} & \underline{0.805} & \underline{21.90} & \textbf{0.217} & \textbf{3.13} & \underline{65.7} & \underline{0.833} & \underline{21.34} & \underline{0.232} & \textbf{1.77} & \textbf{113.4} & \multicolumn{5}{c}{N/A(encountered a bug)} \\
    Hierarchical-3DGS & 0.695 & 20.18 & 0.296 & 6.59 & 46.3 & 0.741 & 19.70 & 0.243 & 10.01 & 44.2 & 0.788 & 19.82 & 0.278 & 6.67 & 36.0 & \underline{0.724} & \underline{22.43} & \underline{0.316} & \underline{10.55} & \underline{18.5} \\
    Ours & \textbf{0.799} & \textbf{24.03} & \textbf{0.233} & \underline{5.18} & \underline{82.0} & \textbf{0.818} & \textbf{24.32} & \underline{0.232} & \underline{4.09} & \textbf{90.9} & \textbf{0.859} & \textbf{27.09} & \textbf{0.208} & \underline{2.79} & \underline{99.4} & \textbf{0.778} & \textbf{26.41} & \textbf{0.293} & \textbf{5.68} & \textbf{93.7} \\

    \bottomrule
    \end{tabular}
    }
    \caption{\tabletitle{Quantitative evaluation on \textit{Building}, \textit{Residence}, \textit{Sci-Art} and \textit{Campus}.} The results for VastGaussian are only partially available as it is not open-sourced and can only be obtained from its paper. All missing results are denoted by a ``--".}
    \label{tab:additional-quan-comp}
    \end{center}
\end{table*}

\newcommand{\spyimgrebuttal}[4]{%
	\begin{tikzpicture}[spy using outlines={yellow,magnification=5,size=1.3cm, connect spies}]
		\node[anchor=south west,inner sep=0] at (0,0) {\includegraphics[width=#1]{#2}};
		\spy on (#3) in node [left] at (#4);
	\end{tikzpicture}%
}

\begin{figure*}[htbp]
   \centering
   \setlength{\tabcolsep}{1pt}
\renewcommand{\arraystretch}{0.5}
\begin{tabular}{ccccc}
{\small Ground Truth} & {\small 3DGS} & {\small CityGaussian} & {\small Hierarchial-3DGS} & {\small Ours} \\

    \spyimg{0.19\textwidth}{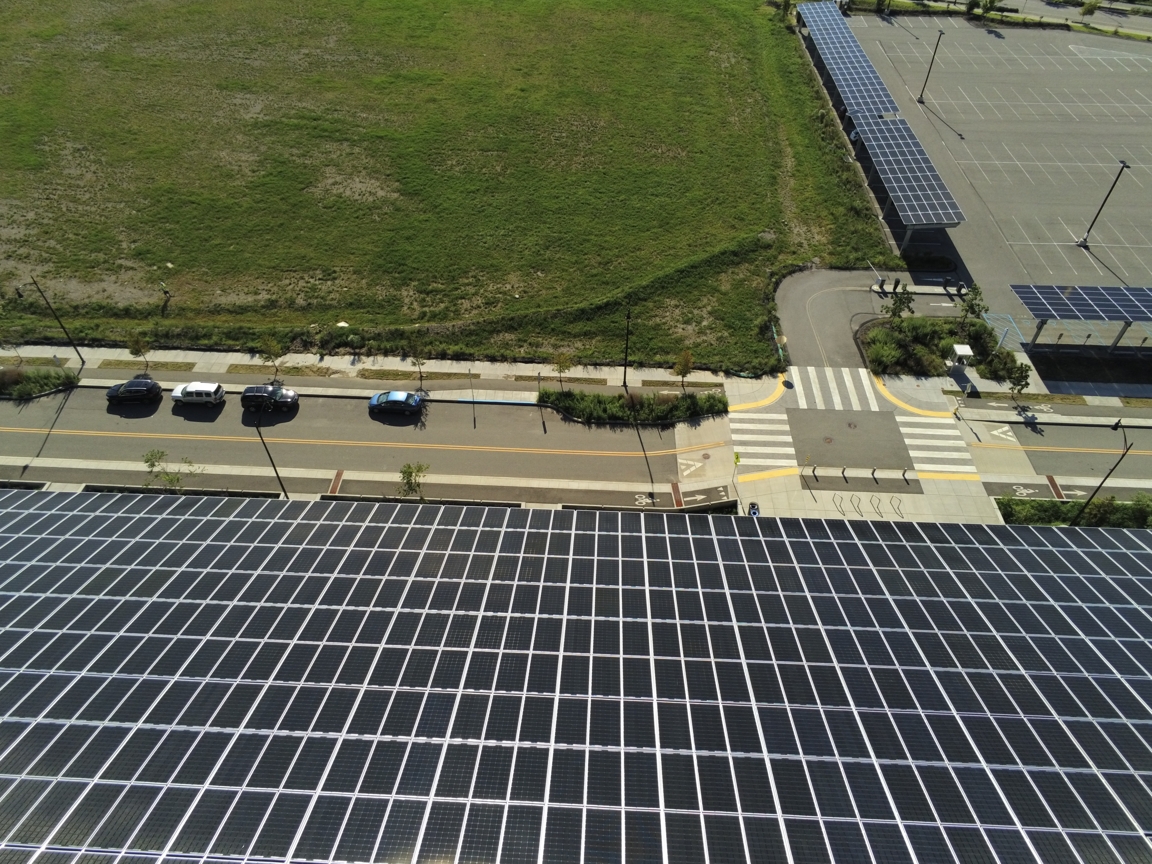}{0.45,0.2}{3,1.7} &
    \spyimg{0.19\textwidth}{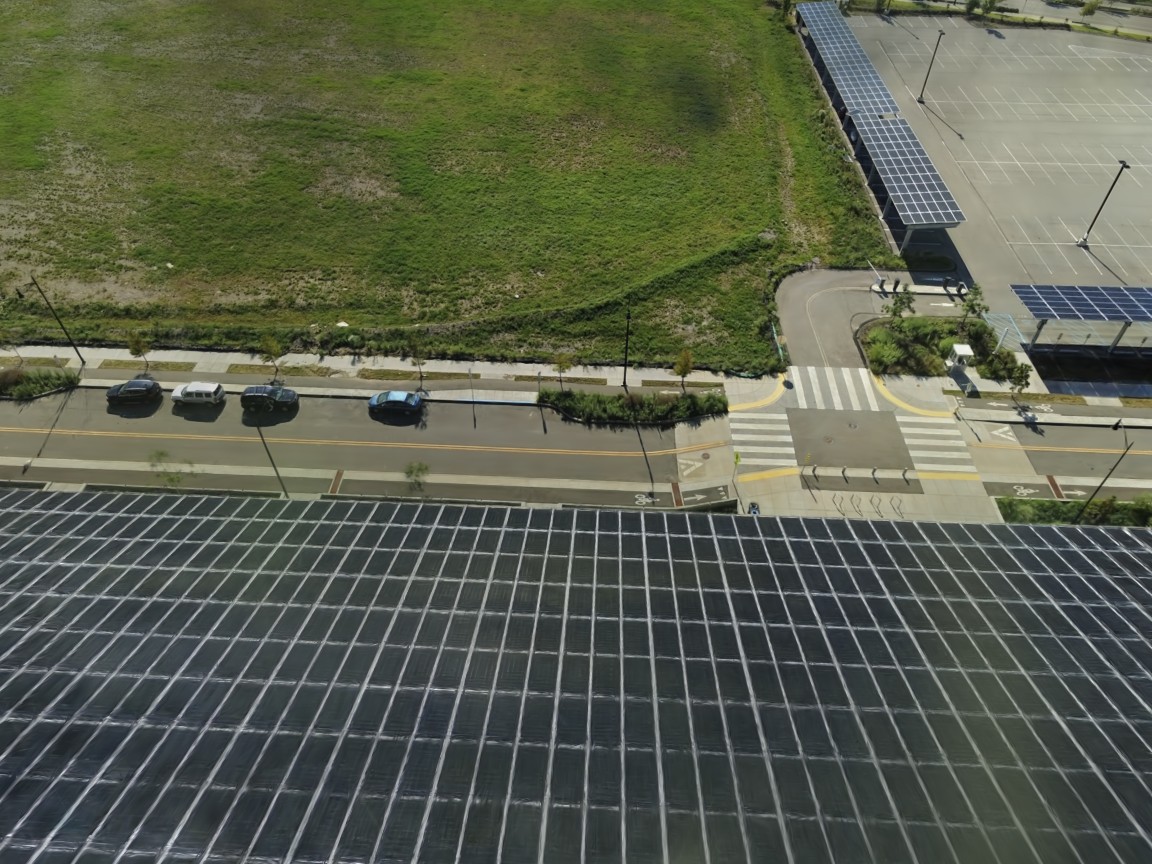}{0.45,0.2}{3,1.7} &
    \spyimg{0.19\textwidth}{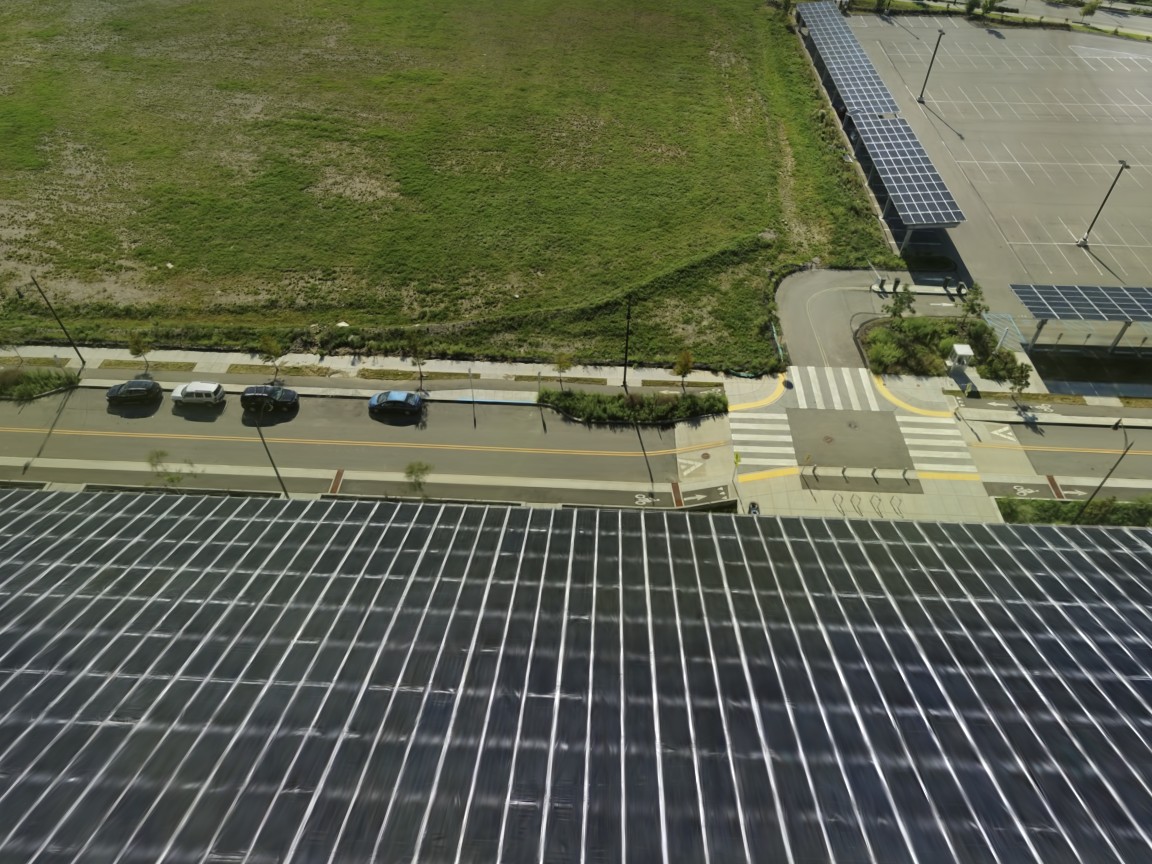}{0.45,0.2}{3,1.7} &
    \spyimg{0.19\textwidth}{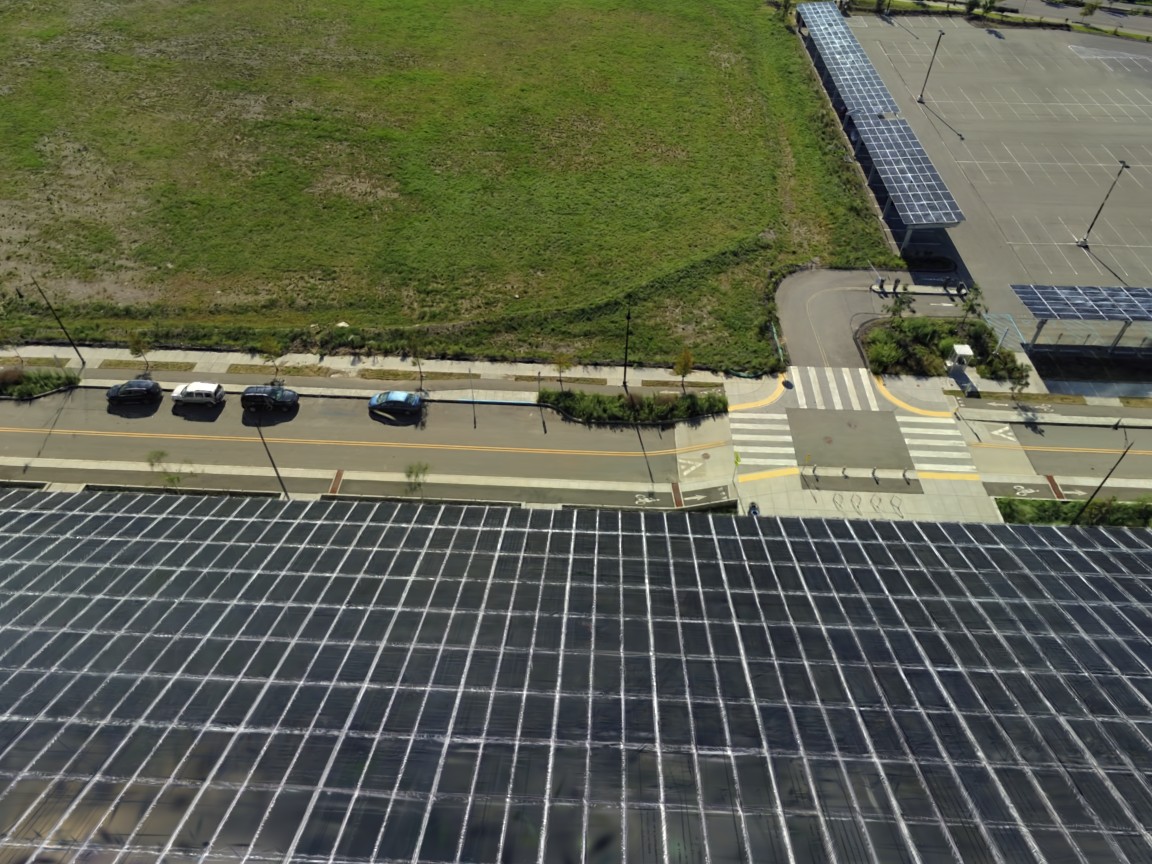}{0.45,0.2}{3,1.7} &
    \spyimg{0.19\textwidth}{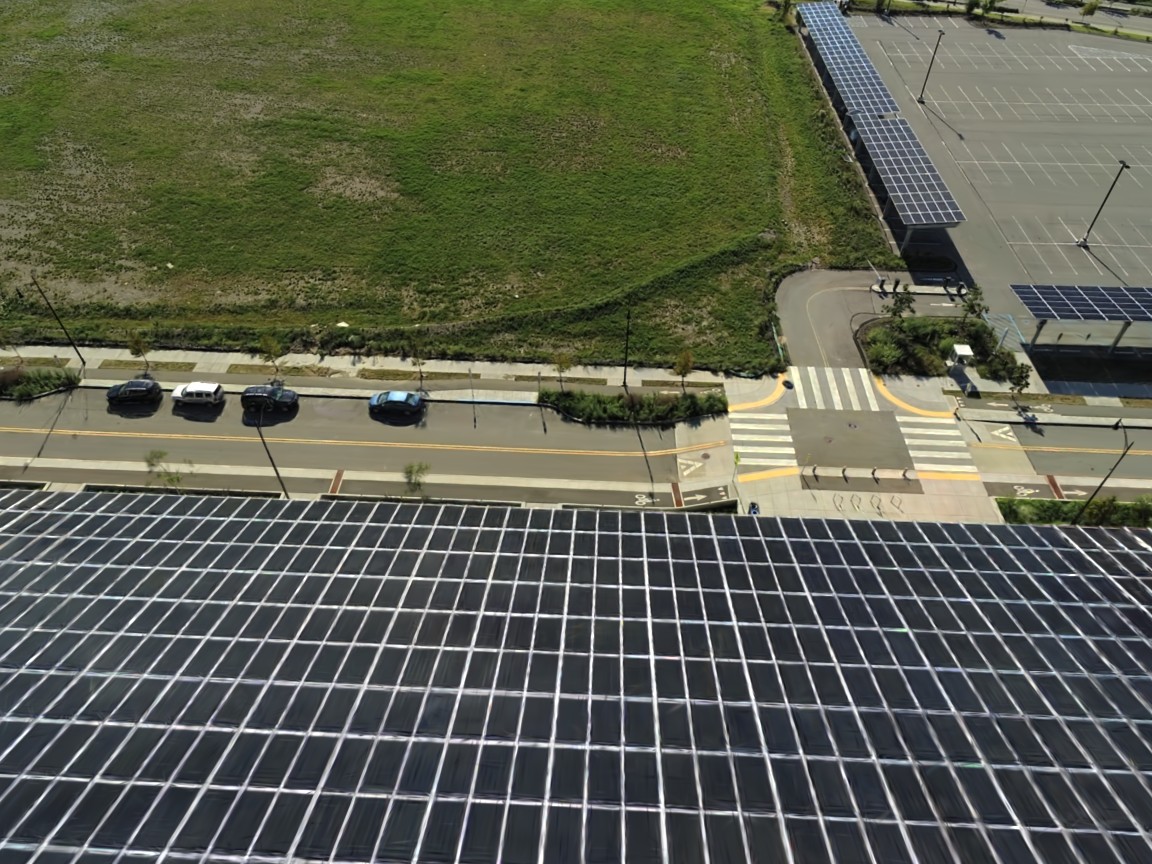}{0.45,0.2}{3,1.7} \\

    \spyimg{0.19\textwidth}{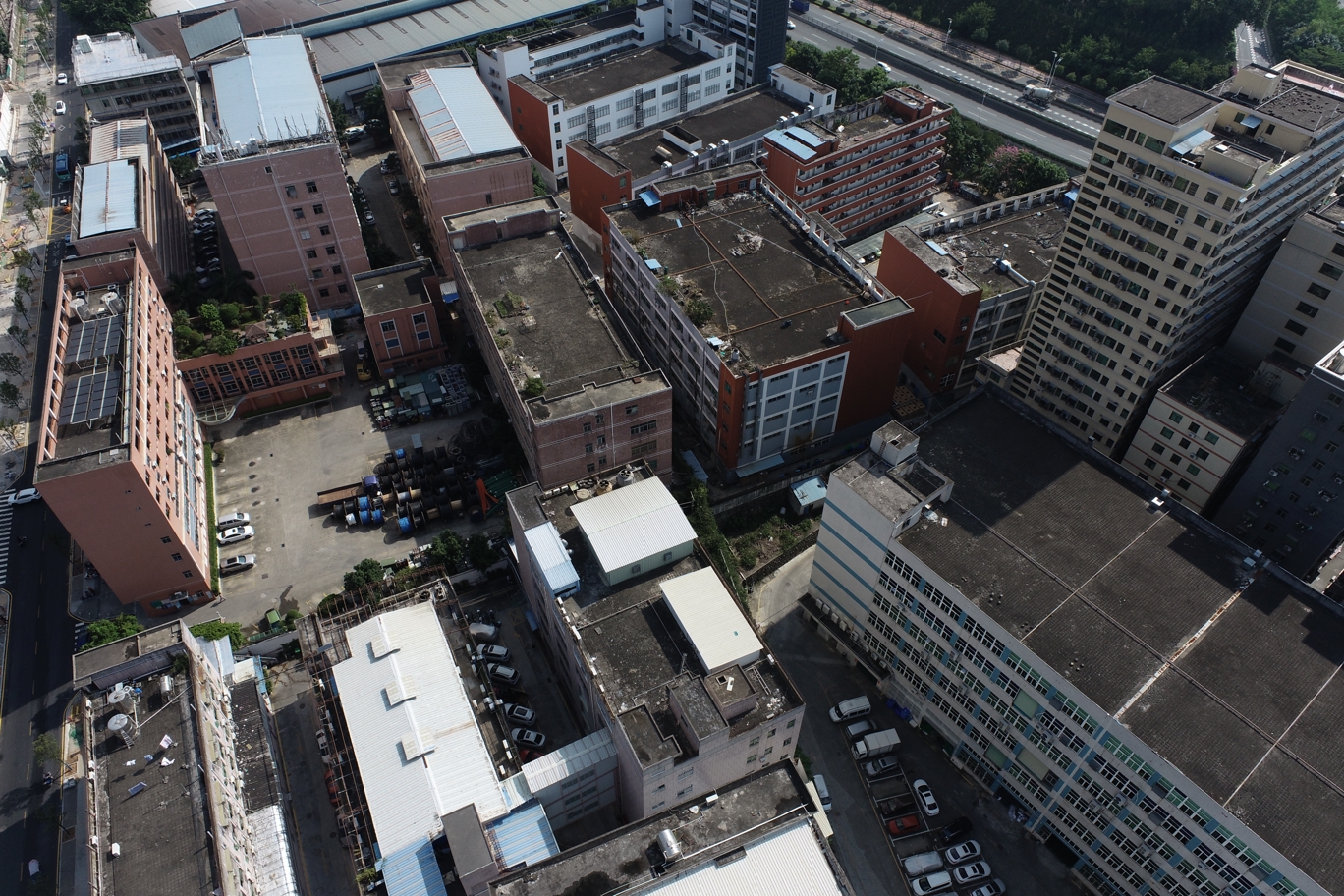}{2.8,1.4}{1.3,0.8} &
    \spyimg{0.19\textwidth}{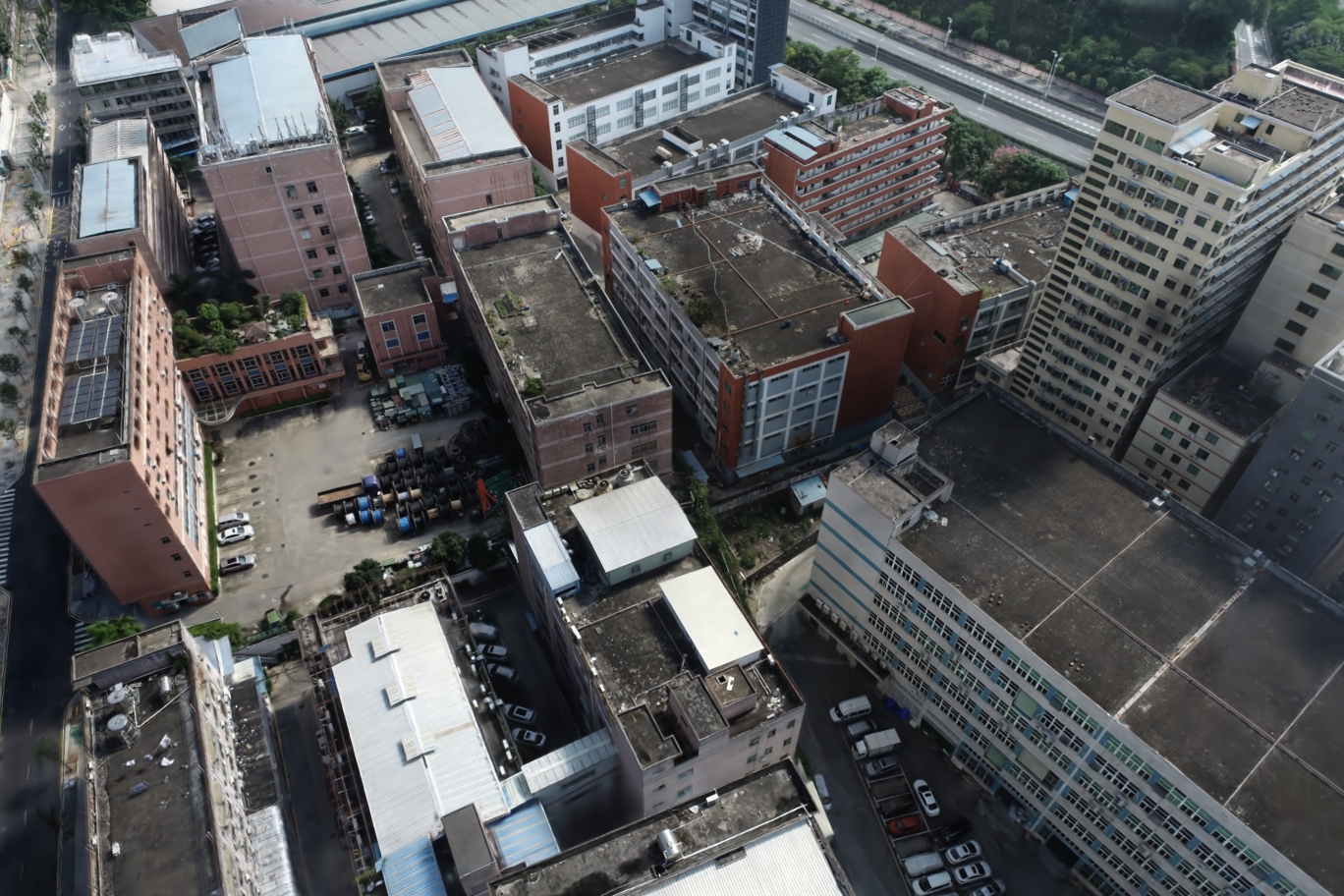}{2.8,1.4}{1.3,0.8} &
    \spyimg{0.19\textwidth}{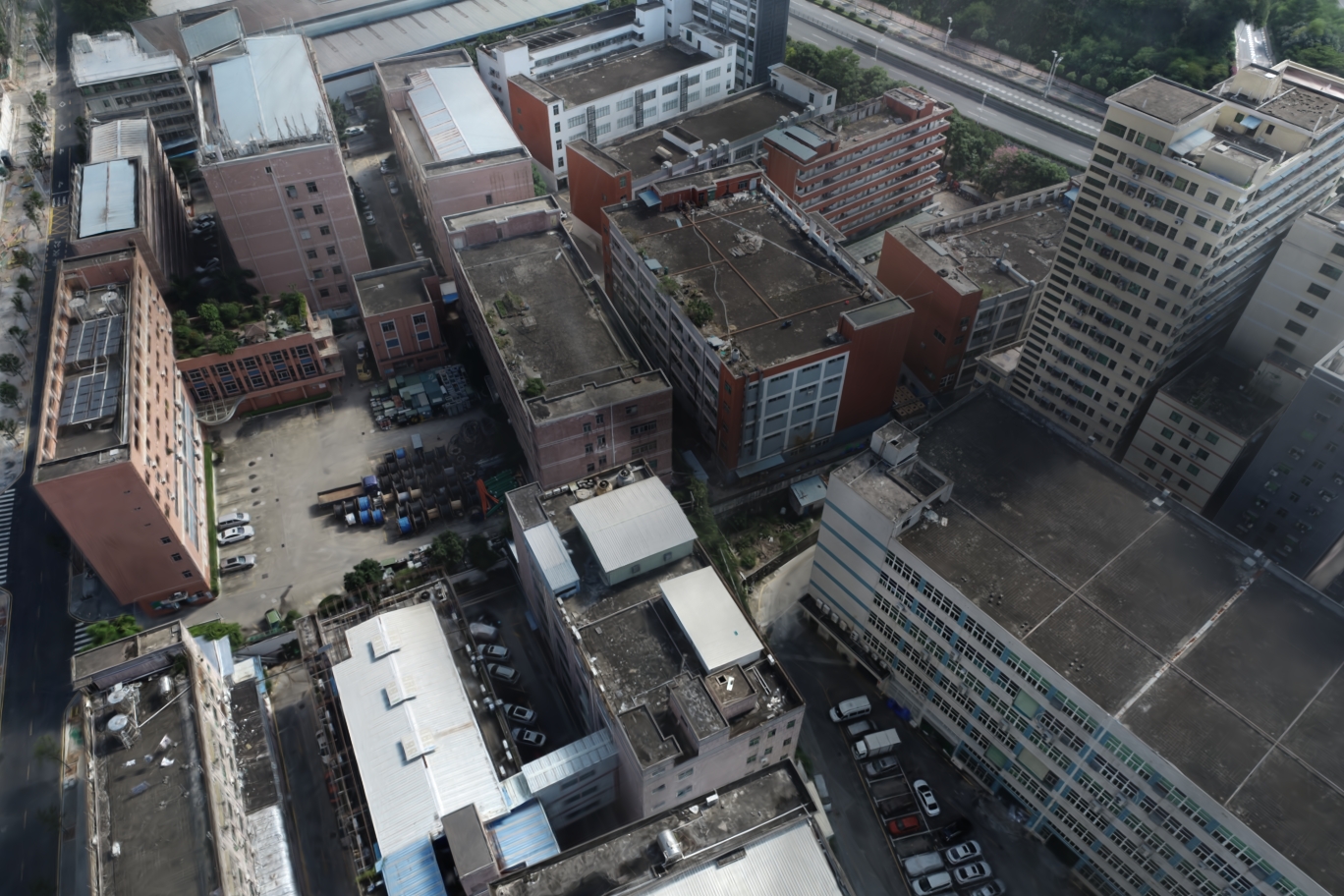}{2.8,1.4}{1.3,0.8} &
    \spyimg{0.19\textwidth}{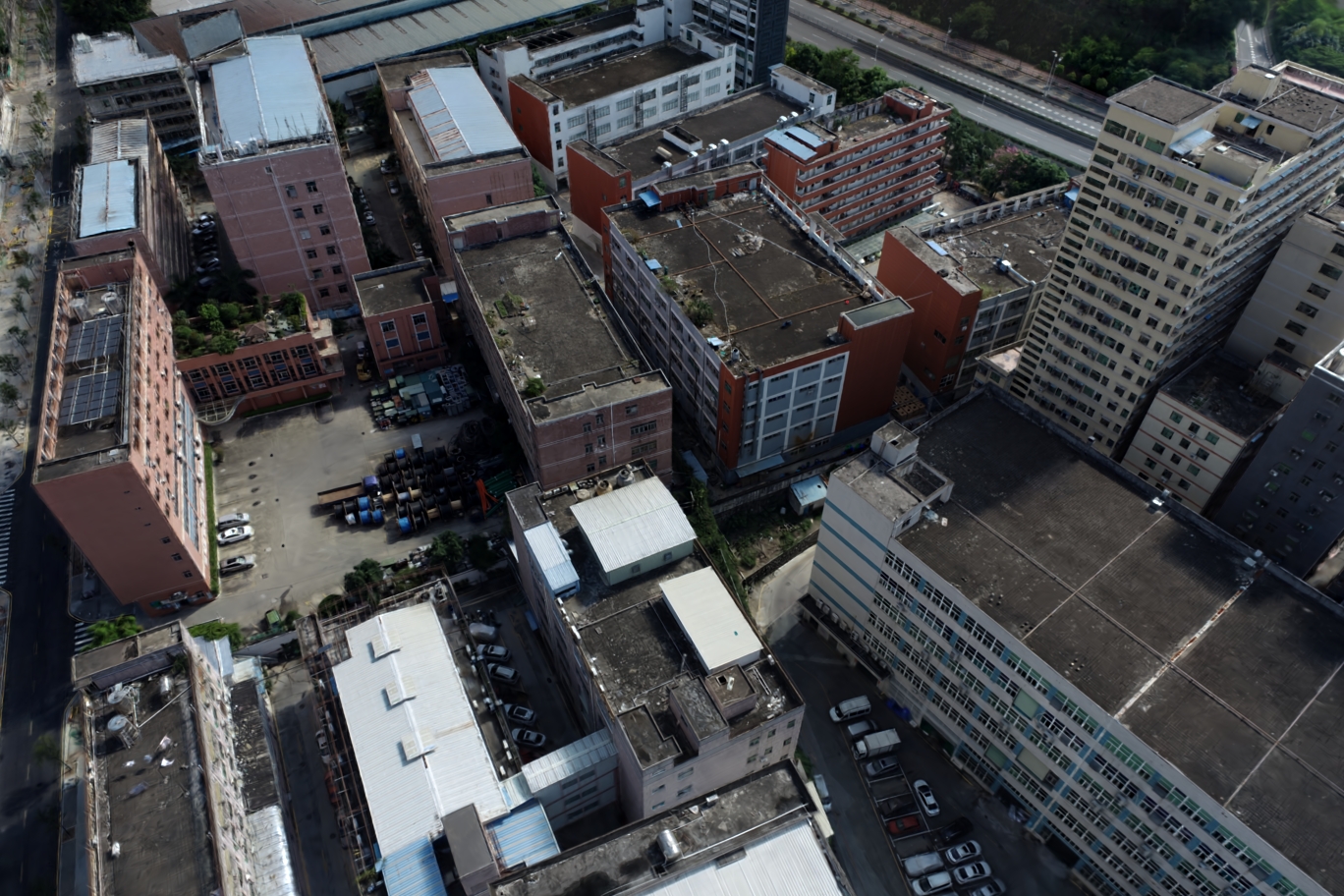}{2.8,1.4}{1.3,0.8} &
    \spyimg{0.19\textwidth}{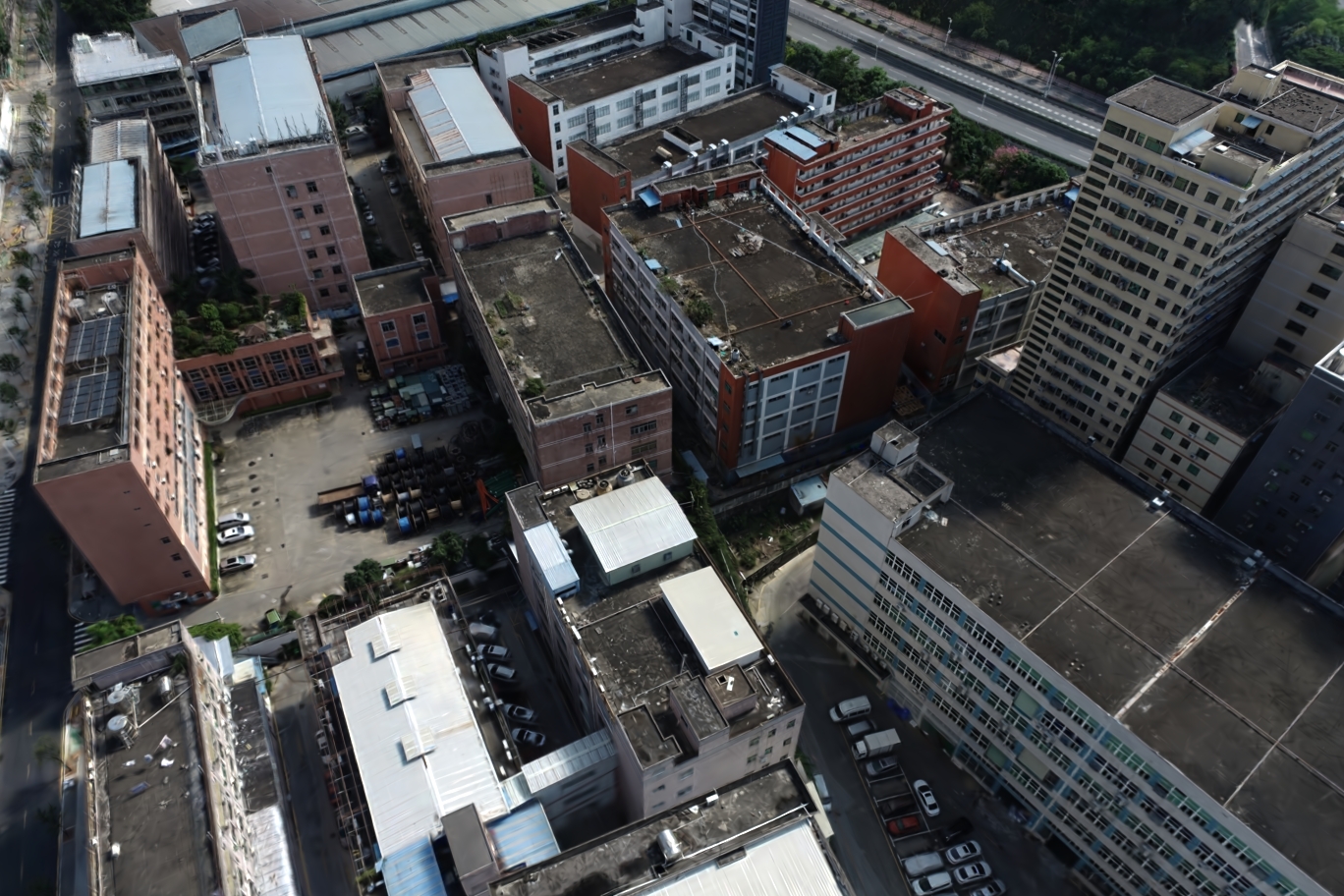}{2.8,1.4}{1.3,0.8} \\

    \spyimgrebuttal{0.19\textwidth}{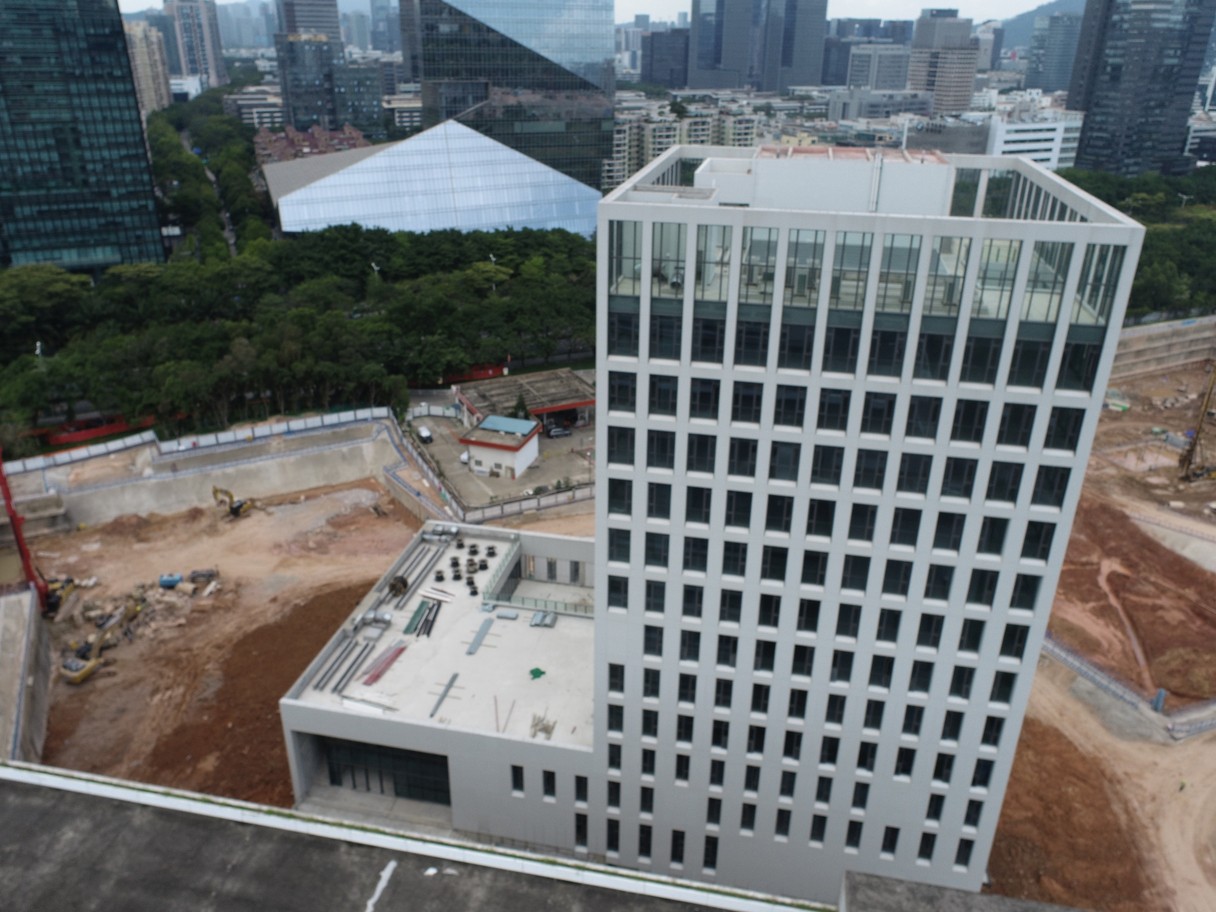}{1.5,1.95}{3,0.75} &
    \spyimgrebuttal{0.19\textwidth}{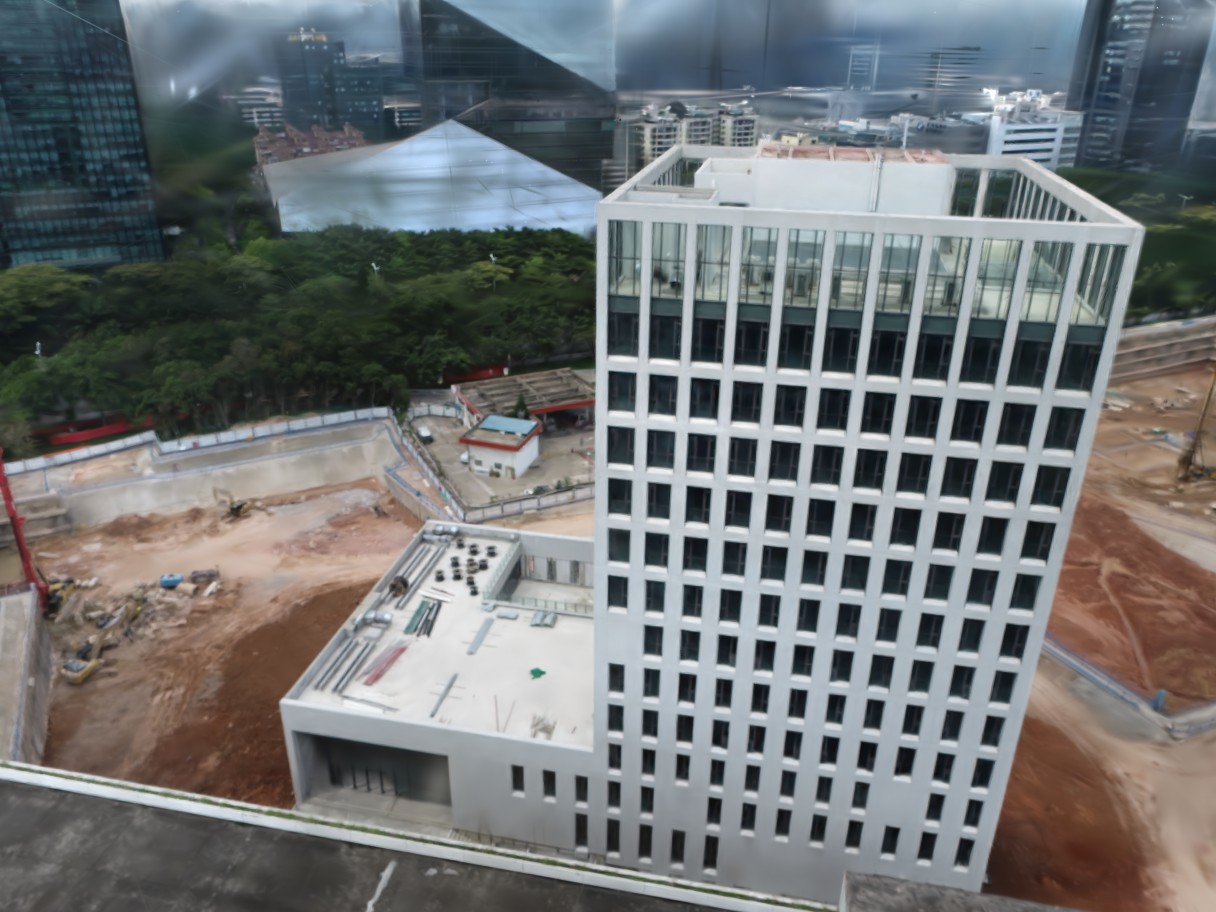}{1.5,1.95}{3,0.75} &
    \spyimgrebuttal{0.19\textwidth}{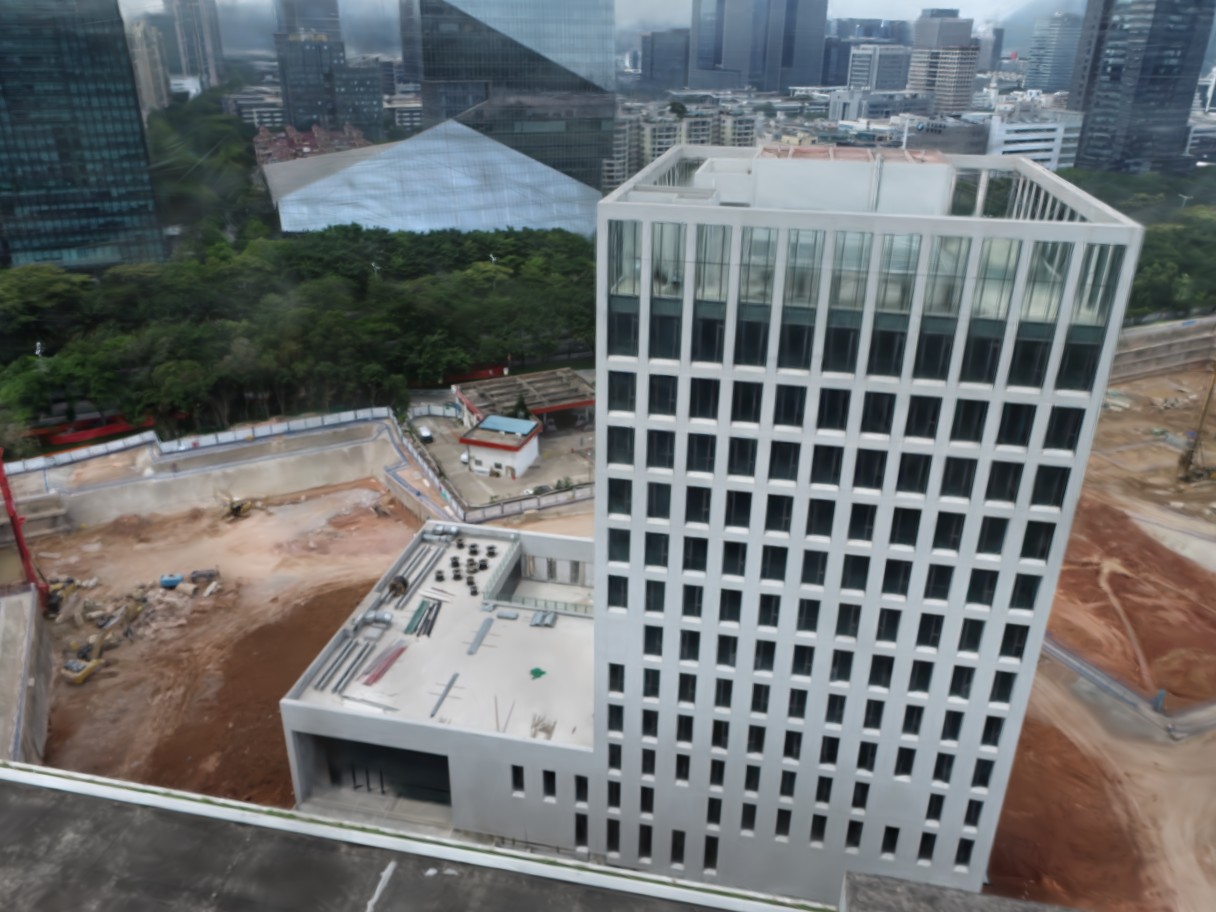}{1.5,1.95}{3,0.75} &
    \spyimgrebuttal{0.19\textwidth}{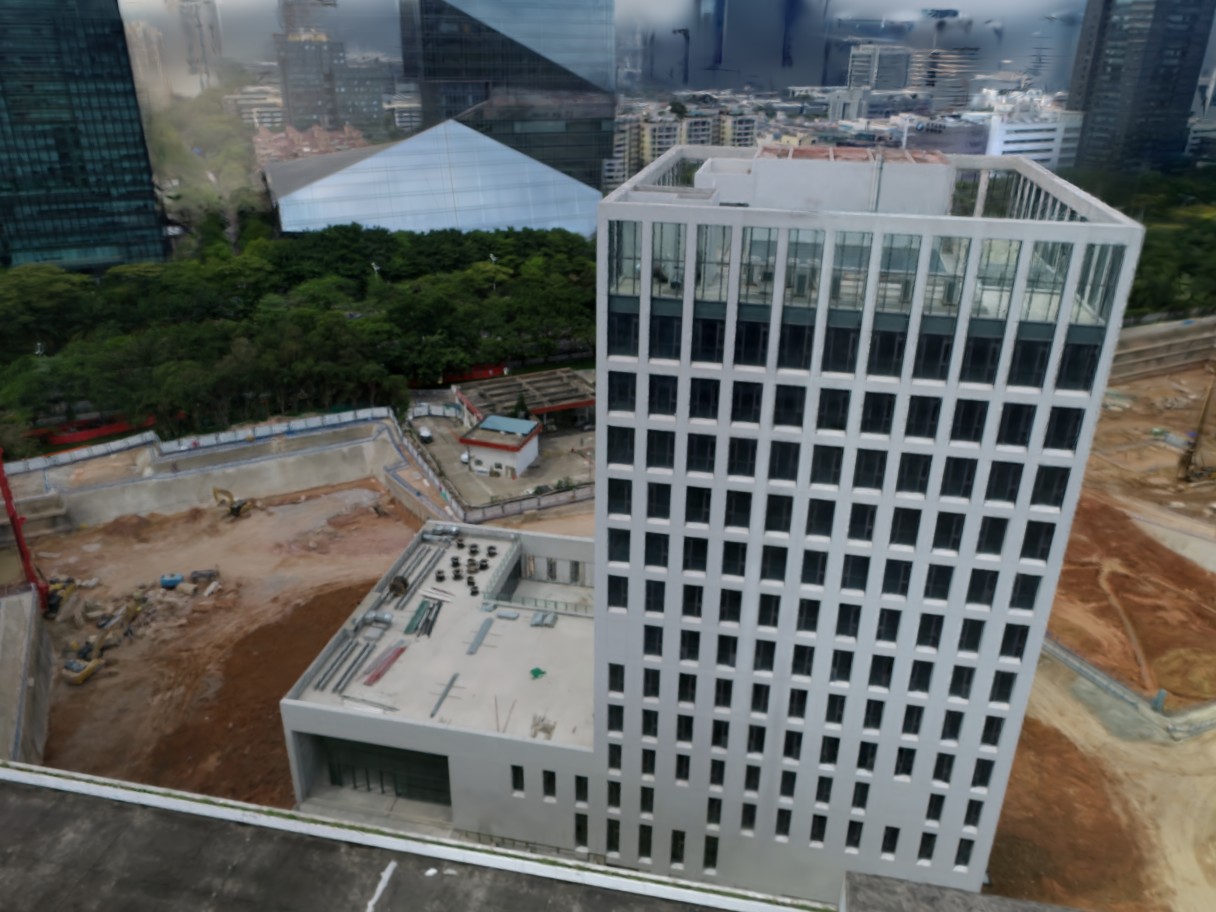}{1.5,1.95}{3,0.75} &
    \spyimgrebuttal{0.19\textwidth}{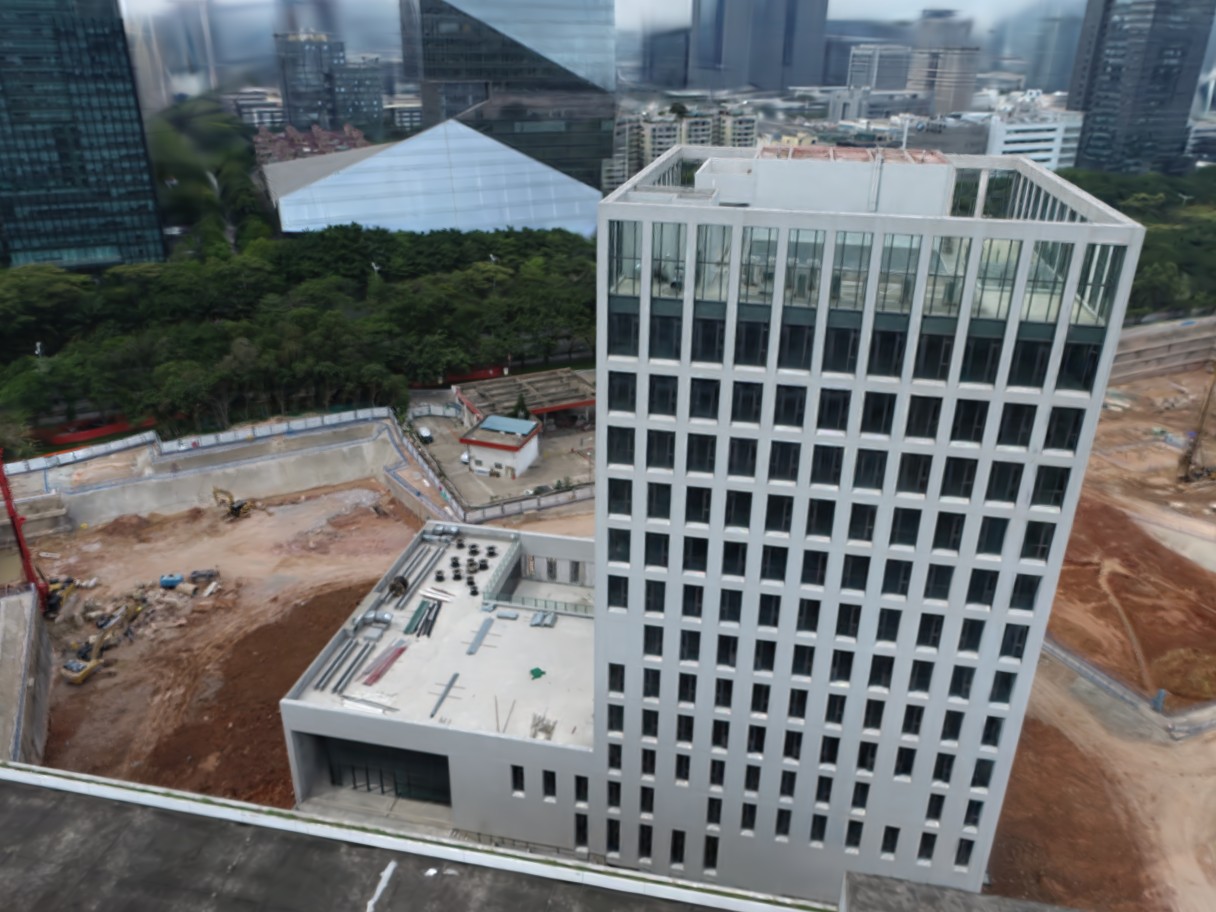}{1.5,1.95}{3,0.75} \\

    \spyimgrebuttal{0.19\textwidth}{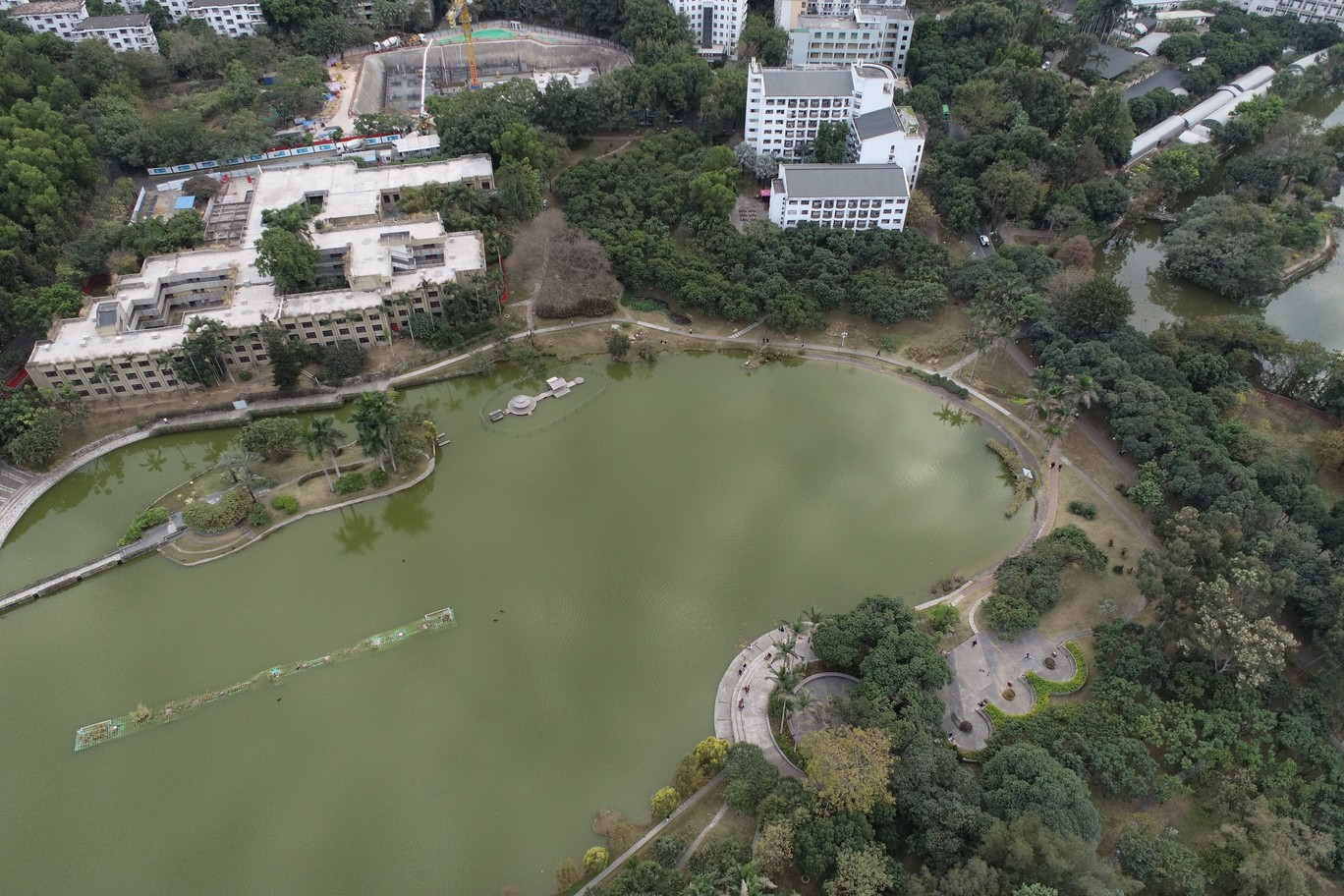}{1.3,1.25}{3,0.7} &
    \spyimgrebuttal{0.19\textwidth}{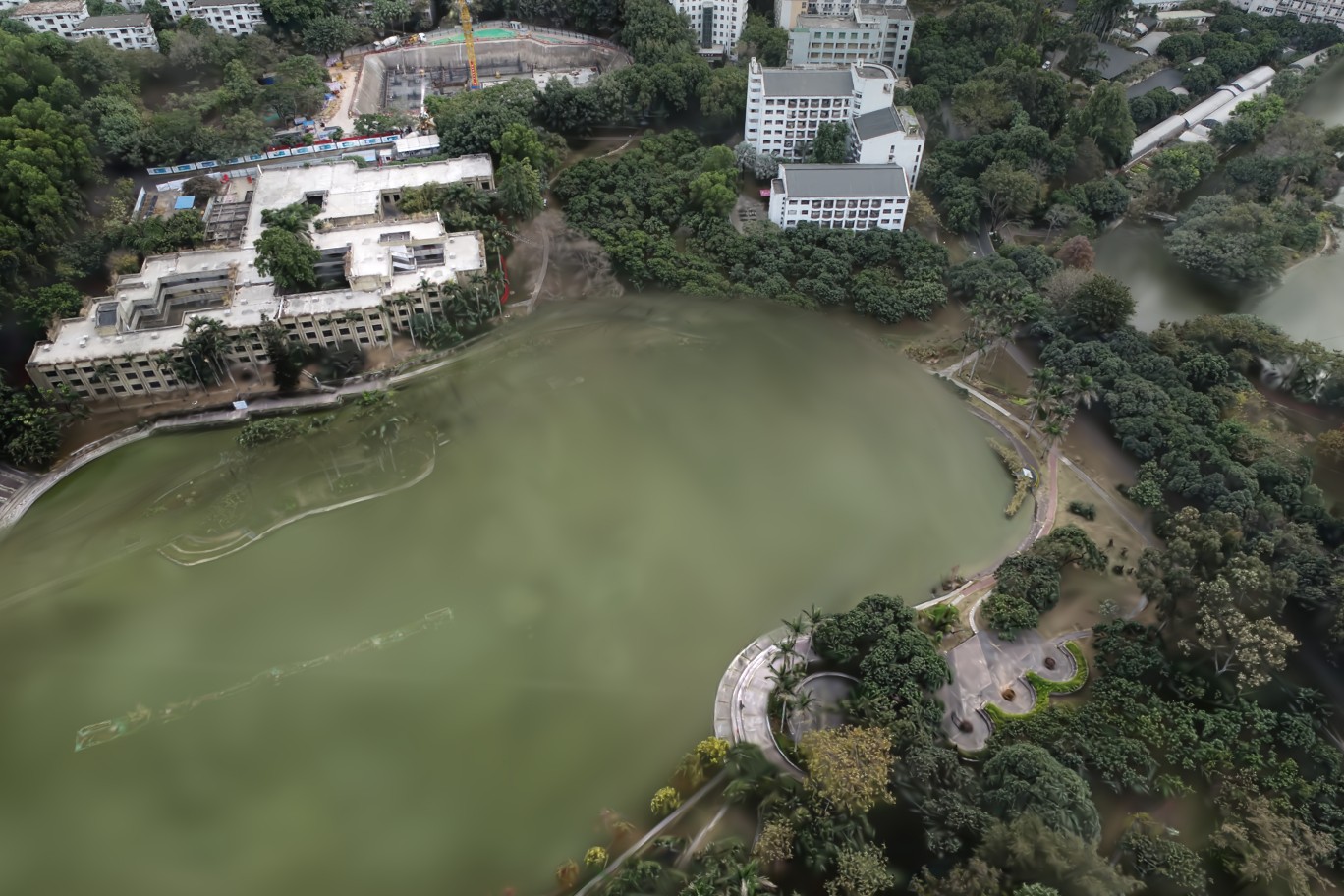}{1.3,1.25}{3,0.7} &
    \spyimgrebuttal{0.19\textwidth}{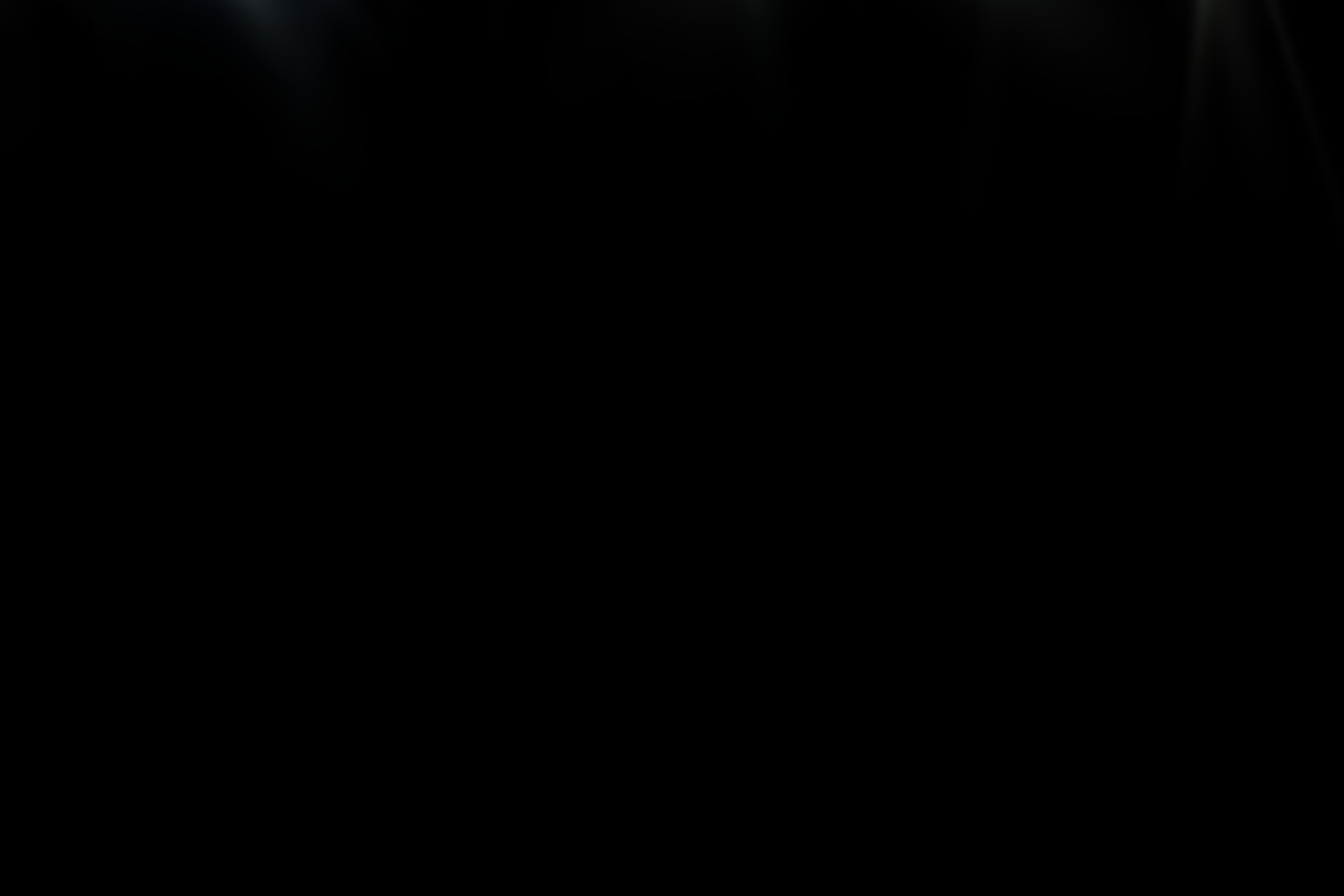}{1.3,1.25}{3,0.7} &
    \spyimgrebuttal{0.19\textwidth}{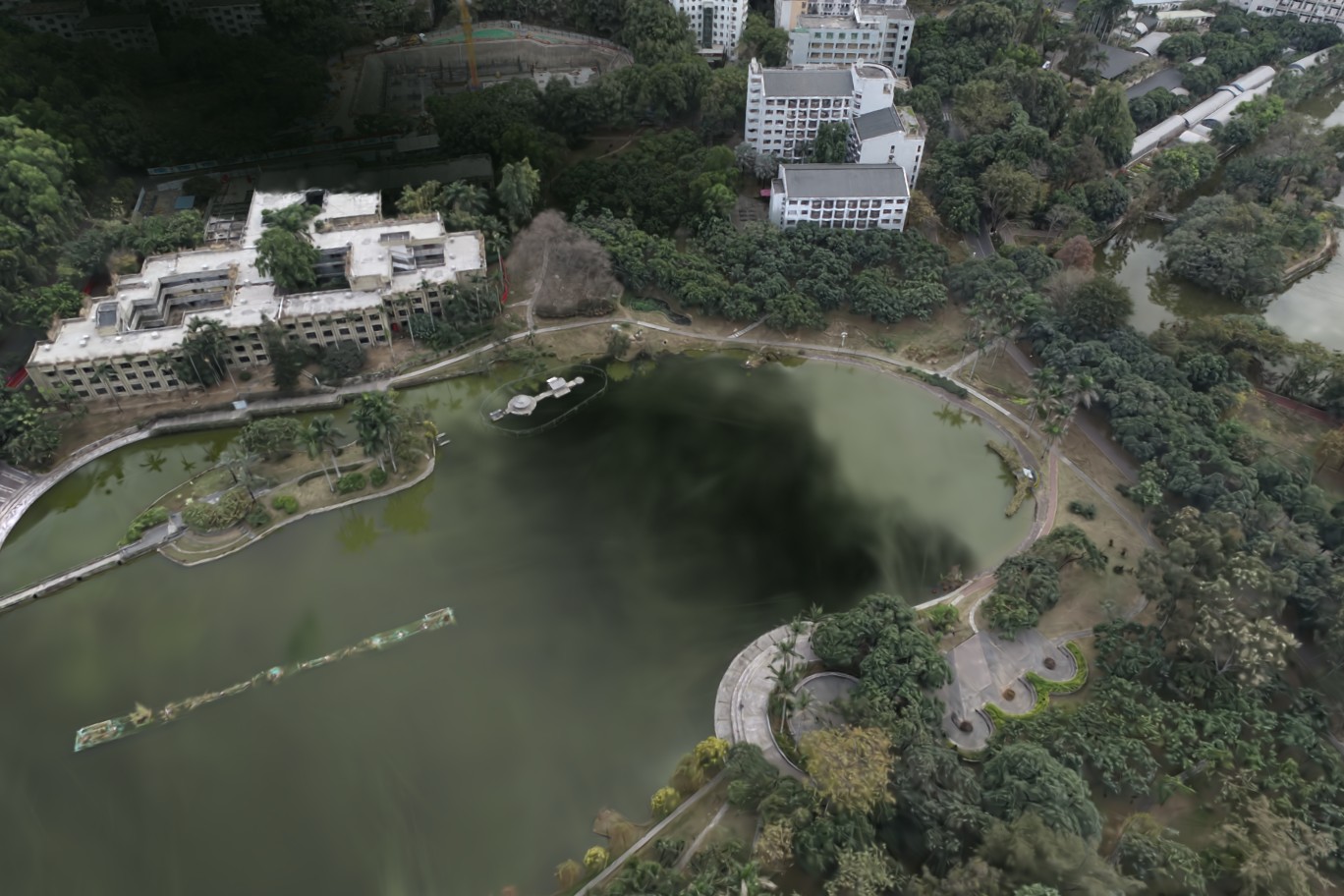}{1.3,1.25}{3,0.7} &
    \spyimgrebuttal{0.19\textwidth}{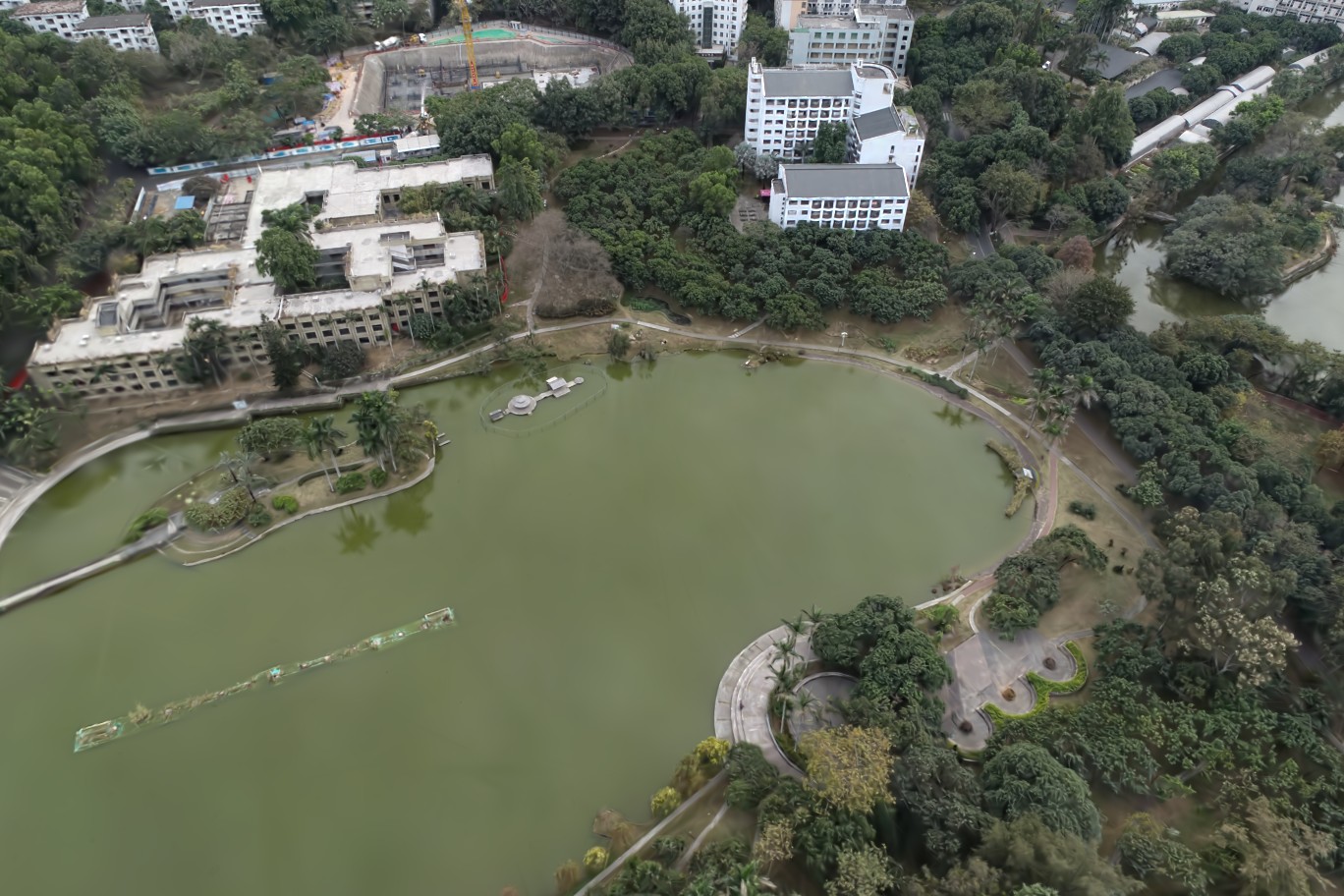}{1.3,1.25}{3,0.7} \\

\end{tabular}
    \caption{\tabletitle{Visualization results on \textit{Building}, \textit{Residence}, \textit{Sci-Art} and \textit{Campus} of ours and previous work.} All methods, except for 3DGS, render in LOD mode.
    The LOD mode of CityGaussian encountered a bug in the \textit{Campus}, resulting in a completely black rendered image.
    }
    \label{fig:additional-qualitative-comparisons}
\end{figure*}

\subsection{Quantitative Comparison of Detail Levels}

\Cref{tab:compare-all-levels} compares the performance of our three detail levels. The results show that all three levels achieve high reconstruction quality. The lower levels exhibit higher numerical values than the higher levels because they are trained and evaluated using downsampled images. Additionally, comparing the \#G across levels further confirms that our LOD strategy effectively controls resource consumption.
\begin{table*}[htbp]
	\begin{center}
		\resizebox{0.8\linewidth}{!}{
			\begin{tabular}{l|rrrr|rrrr|rrrr}
				\toprule
				Scene   &   \multicolumn{4}{c|}{\emph{Rubble}}  & \multicolumn{4}{c|}{\emph{\ourcampusname{}}} & \multicolumn{4}{c}{\emph{BigCity}} \\
				\midrule
				Metrics &  \tabmetricheadwithoutfps{} &
				\tabmetricheadwithoutfps{} &
				\tabmetricheadwithoutfps{} \\
				\midrule
                    
                    1st level & 
                    0.870 & 28.34 & 0.153 & 3.71 &
                    0.889 & 26.57 & 0.114 & 5.41 &
                    0.925 & 27.48 & 0.098 & 10.38 \\
                    
                    2nd level & 
                    0.825 & 27.16 & 0.224 & 7.13 &
                    0.835 & 25.74 & 0.197 & 13.99 &
                    0.855 & 26.10 & 0.198 & 30.47 \\
                    
                    3rd level & 
                    0.826 & 27.29 & 0.228 & 13.52 &
                    0.822 & 25.85 & 0.232 & 25.58 &
                    0.847 & 26.62 & 0.219 & 75.15 \\
				\bottomrule
			\end{tabular}
		}
		\caption{\tabletitle{Quantitative evaluation of all the levels of our method, evaluated using the same downsampling factor as
during training.}}
		\label{tab:compare-all-levels}
	\end{center}
	\centering
\end{table*}

\subsection{Additional ablations}

\Cref{tab:extra-ablation-compare} presents the results of ablation studies on the anti-aliasing, AbsGS, and tile-based culling components in our method.

\noindent\textbf{Anti-aliasing}. The 1th row of \Cref{tab:extra-ablation-compare} reports the impact of anti-aliasing techniques. As shown in \Cref{fig:ablation-aa}, it effectively prevents jagged edges from appearing in areas with low detail levels in images. However, this comes at the cost of requiring more Gaussians. In the \textit{BigCity} scene, since the number of Gaussians has already reached the upper limit without anti-aliasing, enabling anti-aliasing does not allow for additional Gaussians, leading to a slight degradation in metrics. Nevertheless, this feature remains beneficial as it significantly enhances the visual experience.

\noindent\textbf{AbsGS}. The 2nd row of \Cref{tab:extra-ablation-compare} provides the results obtained without AbsGS, highlighting its contribution to quality metrics and enhanced detail restoration, as shown in \Cref{fig:ablation-absgs}. While it does increase the number of Gaussians in certain scenarios, the increase remains within an acceptable range, ensuring that real-time rendering can still be achieved within the constraints of 24GB of VRAM.

\noindent\textbf{Tile-based culling}. The 3rd row of \Cref{tab:extra-ablation-compare} presents the results without tile-based culling, illustrating its role in rendering efficiency. Tile-based culling noticeable improves rendering speed without negatively impacting rendering quality. This is because it skips over redundant Gaussians with minimal contribution, ensuring efficiency while maintaining the desired quality.

\begin{table*}[htbp]
	\begin{center}
		\resizebox{\linewidth}{!}{
			\begin{tabular}{l|ccccc|ccccc|ccccc}
				\toprule
				Scene   &   \multicolumn{5}{c|}{\emph{Rubble}}  & \multicolumn{5}{c|}{\emph{\ourcampusname{}}} &   \multicolumn{5}{c}{\emph{BigCity}} \\
				\midrule
				Metrics & \tabmetrichead{} &
				\tabmetrichead{} &
				\tabmetrichead{} \\
				\midrule
                    w/o anti-aliasing & \textbf{0.817} & 26.85 & \textbf{0.237} & \textbf{3.02} & \underline{103.5} & \textbf{0.817} & 25.69 & \textbf{0.233} & \textbf{4.92} & \underline{67.8} & \textbf{0.847} & \textbf{26.52} & \textbf{0.213} & \textbf{6.68} & \underline{73.6} \\
                    w/o absgrad & 0.795 & 26.67 & 0.275 & \underline{3.15} & \textbf{109.8} & 0.808 & 25.53 & 0.251 & 6.78 & \textbf{68.3} & 0.831 & 26.28 & 0.244 & \underline{6.74} & \textbf{78.1} \\
                    w/o tile-based cull. & \underline{0.814} & \textbf{27.03} & \underline{0.245} & 3.60 & 83.0 & \underline{0.816} & \textbf{25.71} & \underline{0.240} & \underline{6.65} & 54.6 & \underline{0.838} & \underline{26.41} & \underline{0.231} & 6.84 & 65.8 \\
                    full & \underline{0.814} & \textbf{27.03} & \underline{0.245} & 3.60 & 99.7 & \underline{0.816} & \textbf{25.71} & \underline{0.240} & \underline{6.65} & 63.9 & \underline{0.838} & \underline{26.41} & \underline{0.231} & 6.84 & 73.0 \\
                \bottomrule
			\end{tabular}
		}
		\caption{\tabletitle{Additional qualitative ablations.}}
		\label{tab:extra-ablation-compare}
		\vspace{-3mm}
	\end{center}
	\centering
\end{table*}

\begin{figure*}[htbp]
    \centering
    \begin{tabular}{c@{\raggedright\hspace{1.2pt}}c@{\raggedright\hspace{1.2pt}}c@{\raggedright\hspace{1.2pt}}}
        & {\small Anti-Aliasing} & {\small AbsGS}\\
       \rotatebox[origin=c]{90}{Without} & 
       \includeablationimage{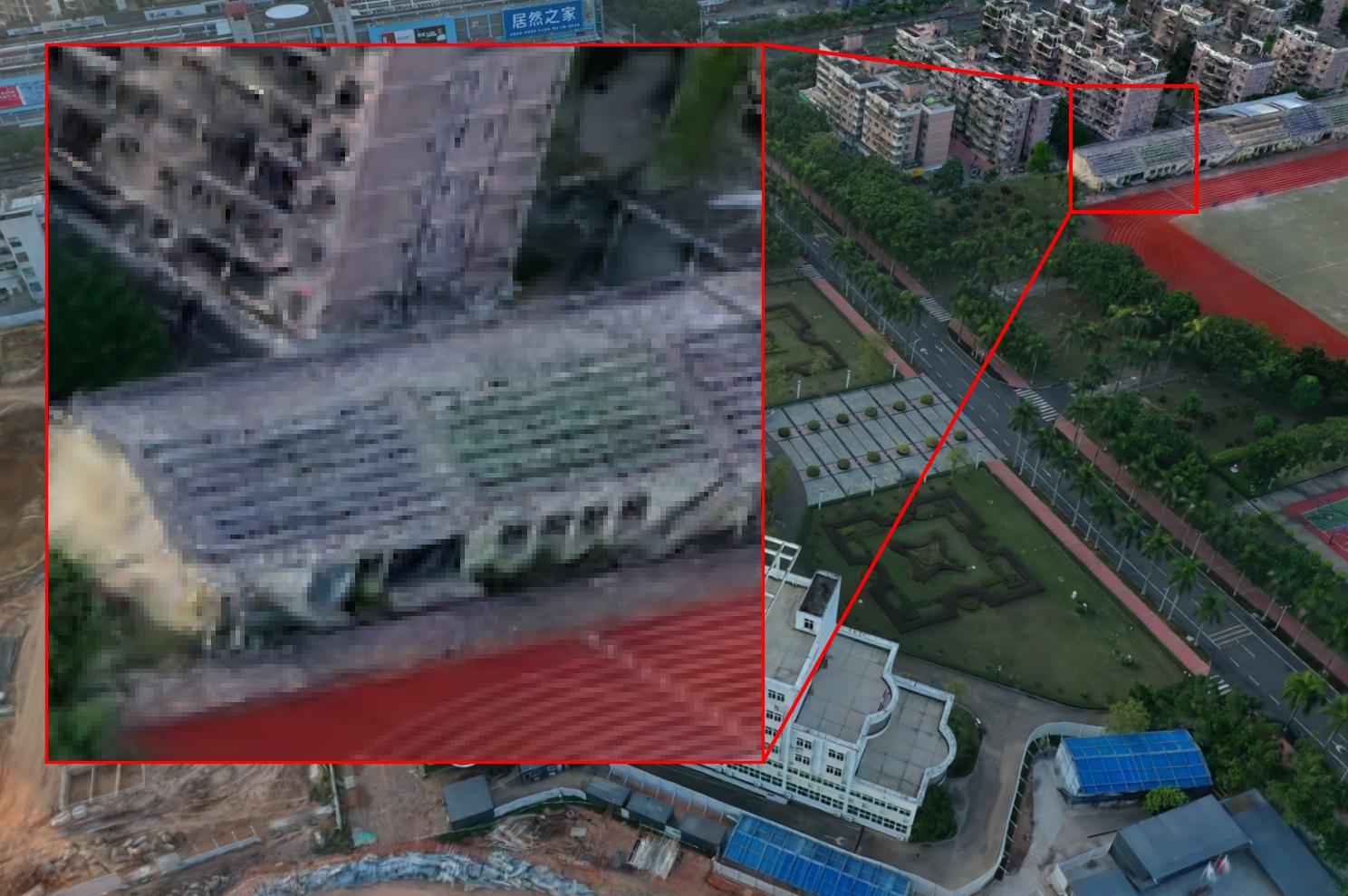} & 
       \includeablationimage{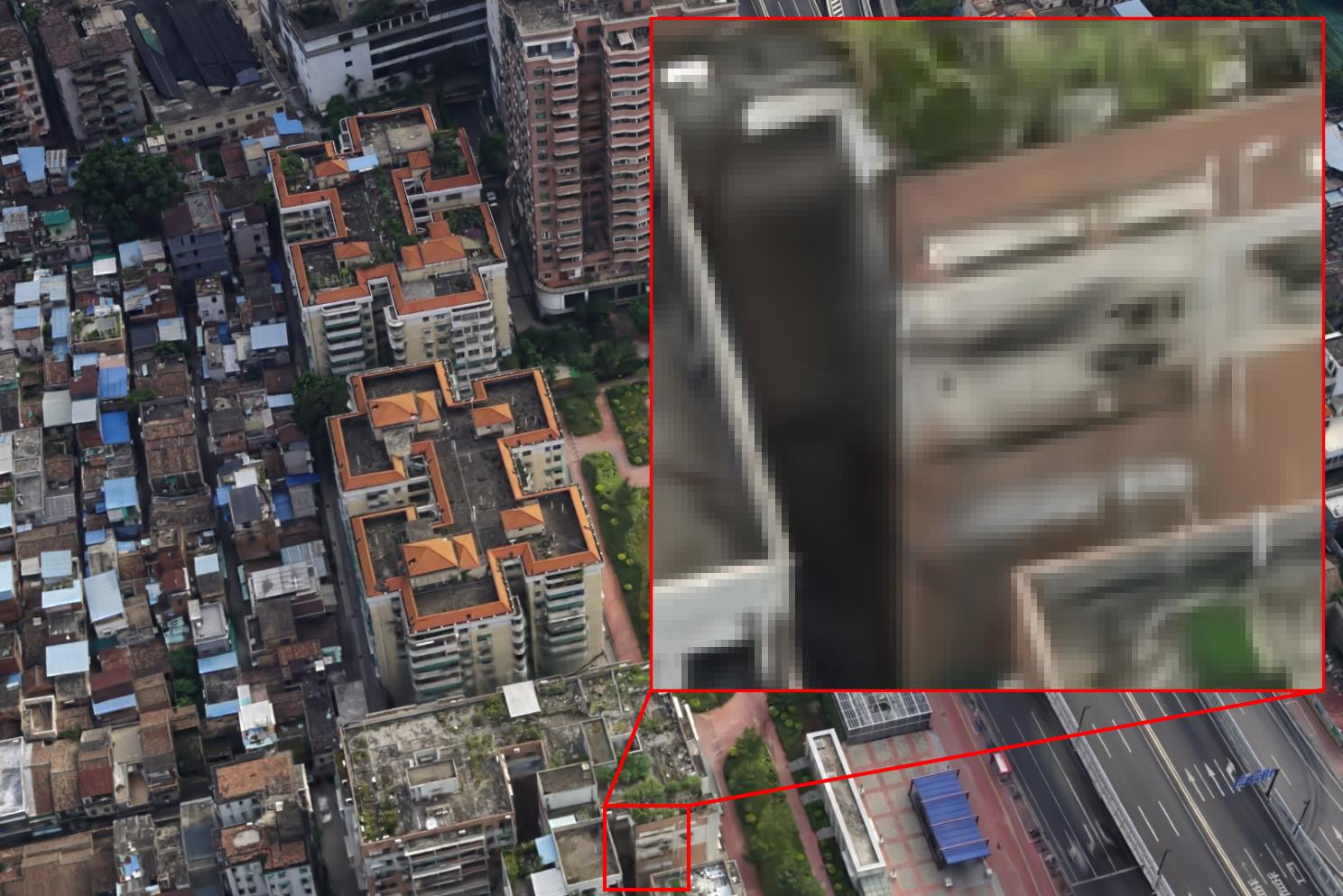} \\
       
       \rule{0pt}{33pt} 
       
       \rotatebox[origin=c]{90}{With} & 
       \includeablationimage{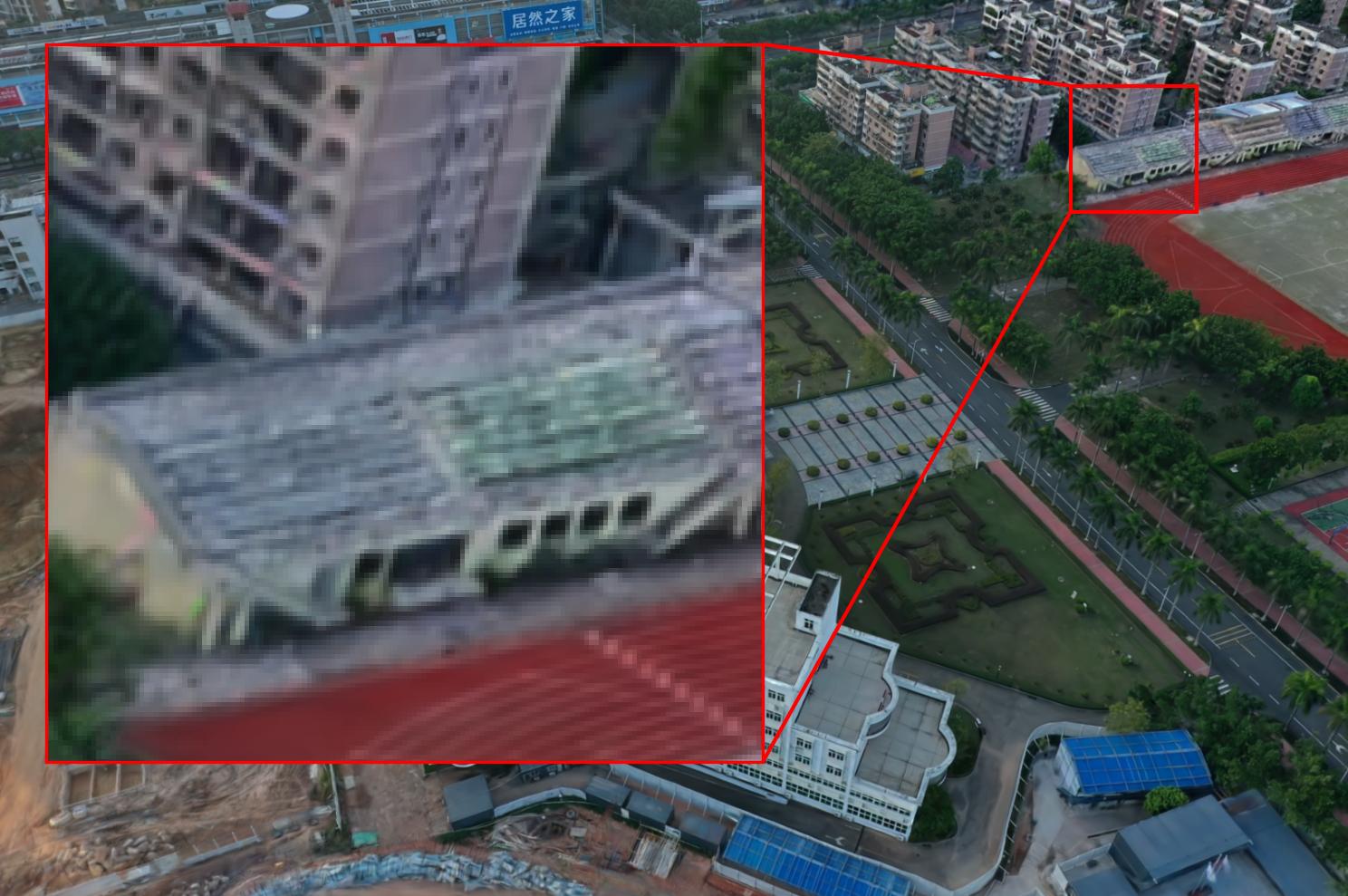} & 
       \includeablationimage{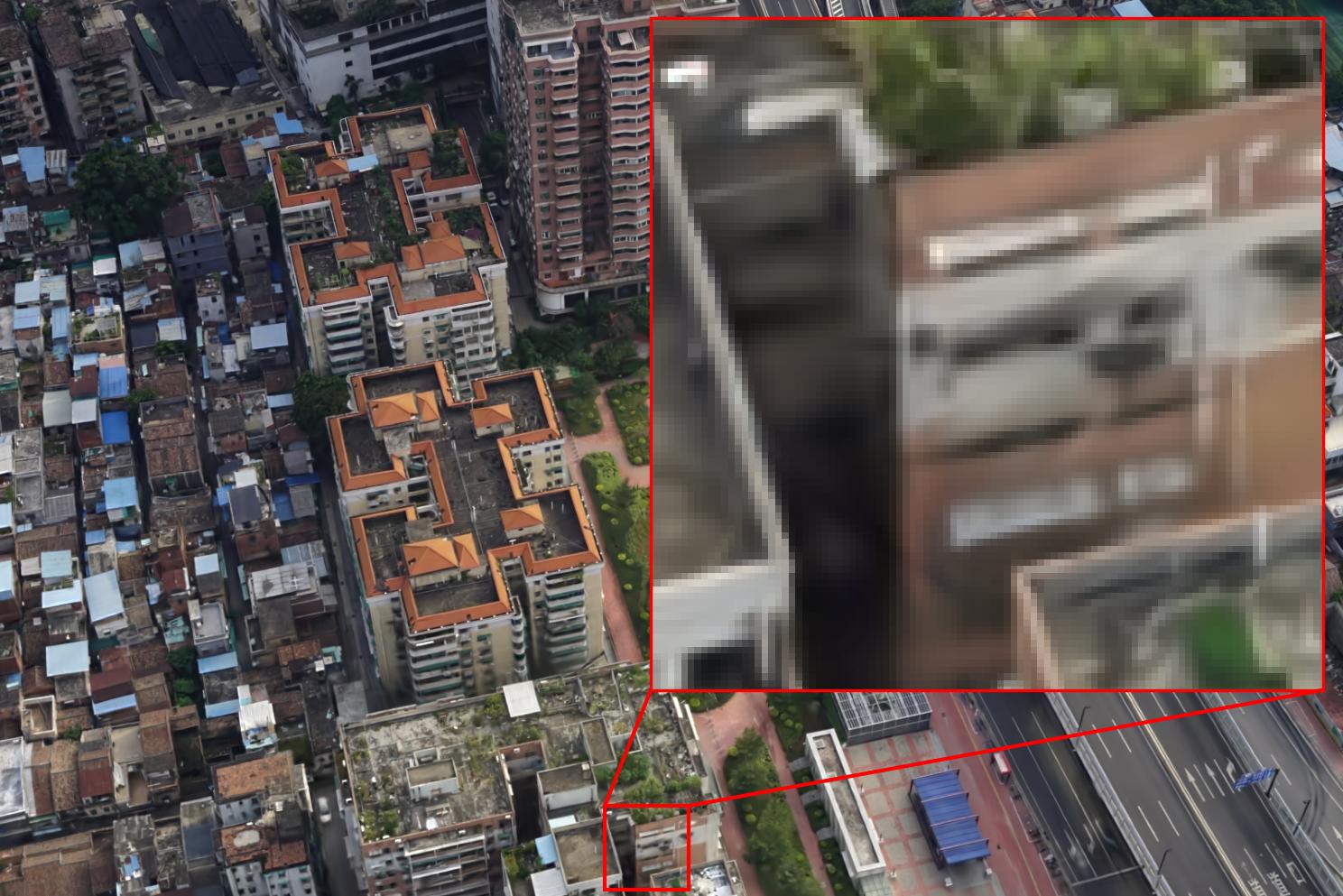} \\

       & \ablationsubfigurelabel{ablation-aa} & \ablationsubfigurelabel{ablation-absgs} \\
    \end{tabular}
    \caption{\tabletitle{Visualization results of ablation on Anti-Aliasing and AbsGS.}}
    \label{fig:additional-ablation-summary}
\end{figure*}

\subsection{Comparison of Gaussian Embedding Lengths}
We evaluate the impact of the length of Gaussian embedding on the \ourcampusname{} scene. As shown in \Cref{tab:embedding-length-exp}, reducing the length leads to a slight drop in metrics, but it is acceptable if the goal is to reduce memory consumption.
\begin{table*}[htbp]
	\begin{center}
		\resizebox{0.23\linewidth}{!}{
                \setlength{\tabcolsep}{3pt}
			\begin{tabular}{l|ccc}
				\toprule
				    Length &  SSIM & PSNR & LPIPS \\
				\midrule
                    4 & \underline{0.816} & 25.58 & \underline{0.237} \\
                    8 & 0.815 & \underline{25.62} & 0.240 \\
                    16 & \textbf{0.822} & \textbf{25.85} & \textbf{0.232} \\
                    \bottomrule
			\end{tabular}
		}
		\caption{\tabletitle{The impact of the length of $\gsembedding{}$.}}
		\label{tab:embedding-length-exp}
            \vspace{-3mm}
	\end{center}
	\centering
\end{table*}

\subsection{Comparison of LOD Selection Parameters}
We evaluate the impact of our LOD parameters on the \ourcampusname{} scene.
\Cref{tab:ablation-lod-part-size} presents the impact of different rendering-time partition sizes on metrics. It can be observed that while smaller partition sizes effectively reduce the number of Gaussians and lower resource consumption, they also lead to a certain degree of metric degradation. Additionally, a larger number of partitions incurs higher overhead due to the LOD selection. In contrast, larger partition sizes exhibit the opposite behavior. This suggests that an optimal partition size must strike a balance between efficiency and rendering quality. \Cref{tab:ablation-lod-dist} illustrates the impact of different distance thresholds for detail levels. Increasing the distance thresholds generally improves rendering quality but also leads to higher resource consumption and reduced rendering speed. Therefore, selecting an appropriate distance threshold requires a trade-off between efficiency and quality.

\begin{table*}[t!]
	\begin{center}
		\resizebox{0.45\linewidth}{!}{
			\begin{tabular}{l|cccccc}
				\toprule
				Part. Size &  SSIM & PSNR & LPIPS & FPS & \#G ($10^6$) & \#P \\
				\midrule
                    45m & 0.803  & 25.27  & 0.256  & 35.9  & \textbf{2.11} & 860 \\
                    90m & 0.808  & 25.46  & 0.250  & 61.7  & \underline{3.06} & 228 \\
                    135m & \underline{0.810}  & \underline{25.50}  & \underline{0.247}  & \textbf{64.4}  & 3.68 & 140 \\
                    180m & \textbf{0.813}  & \textbf{25.62}  & \textbf{0.243}  & \underline{63.1}  & 5.19 & 64 \\
                    \bottomrule
			\end{tabular}
		}
		\caption{\tabletitle{Qualitative ablations of different rendering-time partition size on the \ourcampusname{} scene.} \#P represents the number of partitions.}
		\label{tab:ablation-lod-part-size}
		\vspace{-3mm}
	\end{center}
	\centering
\end{table*}

\begin{table*}[htbp]
	\begin{center}
		\resizebox{0.45\linewidth}{!}{
			\begin{tabular}{l|ccccc}
				\toprule
				Distances &  SSIM & PSNR & LPIPS & FPS & \#G ($10^6$) \\
				\midrule
                    (45m, 90m, $\infty$) & 0.789 & 25.00 & 0.271 & \textbf{66.5} & \textbf{2.40} \\
                    (90m, 180m, $\infty$) & 0.808  & 25.46 & 0.250 & \underline{61.7} & \underline{3.06} \\
                    (135m, 270m, $\infty$) & \underline{0.815} & \underline{25.65} & \underline{0.242} & 58.8 & 3.62 \\
                    (180m, 360m, $\infty$) & \textbf{0.818} & \textbf{25.73} & \textbf{0.238} & 56.2 & 4.13 \\
                    \bottomrule
			\end{tabular}
		}
		\caption{\tabletitle{Qualitative ablations of distance thresholds on the \ourcampusname{} scene with a partition size of 90m.} The distance values represent the maximum distances at which the 3rd, 2nd, and 1st LOD levels are used.}
		\label{tab:ablation-lod-dist}
		\vspace{-3mm}
	\end{center}
	\centering
\end{table*}

\subsection{Evaluation of Similarity Regularization}
Similarity regularization enhances the generalization ability of the appearance transformation module. When performing out-of-domain inference, such as predicting the unobserved regions of an image using its embedding, this regularization effectively mitigates abrupt color changes and suppresses artifacts, as illustrated in the first row of \Cref{fig:sim-reg-vis}.
\newcommand{\spyimgcrop}[4]{%
	\begin{tikzpicture}[spy using outlines={yellow,magnification=3,size=1.5cm, connect spies}]
		\node[anchor=south west,inner sep=0] at (0,0) {\includegraphics[trim={6cm 0 6cm 0},clip,width=#1]{#2}};
		\spy on (#3) in node [left] at (#4);
	\end{tikzpicture}%
}
\begin{figure*}[htbp]
    \begin{center}
    \begin{subfigure}{0.26\linewidth}
        \raisebox{0.5\height}{
        \includegraphics[trim={5cm 0 5cm 0},clip,width=\linewidth]{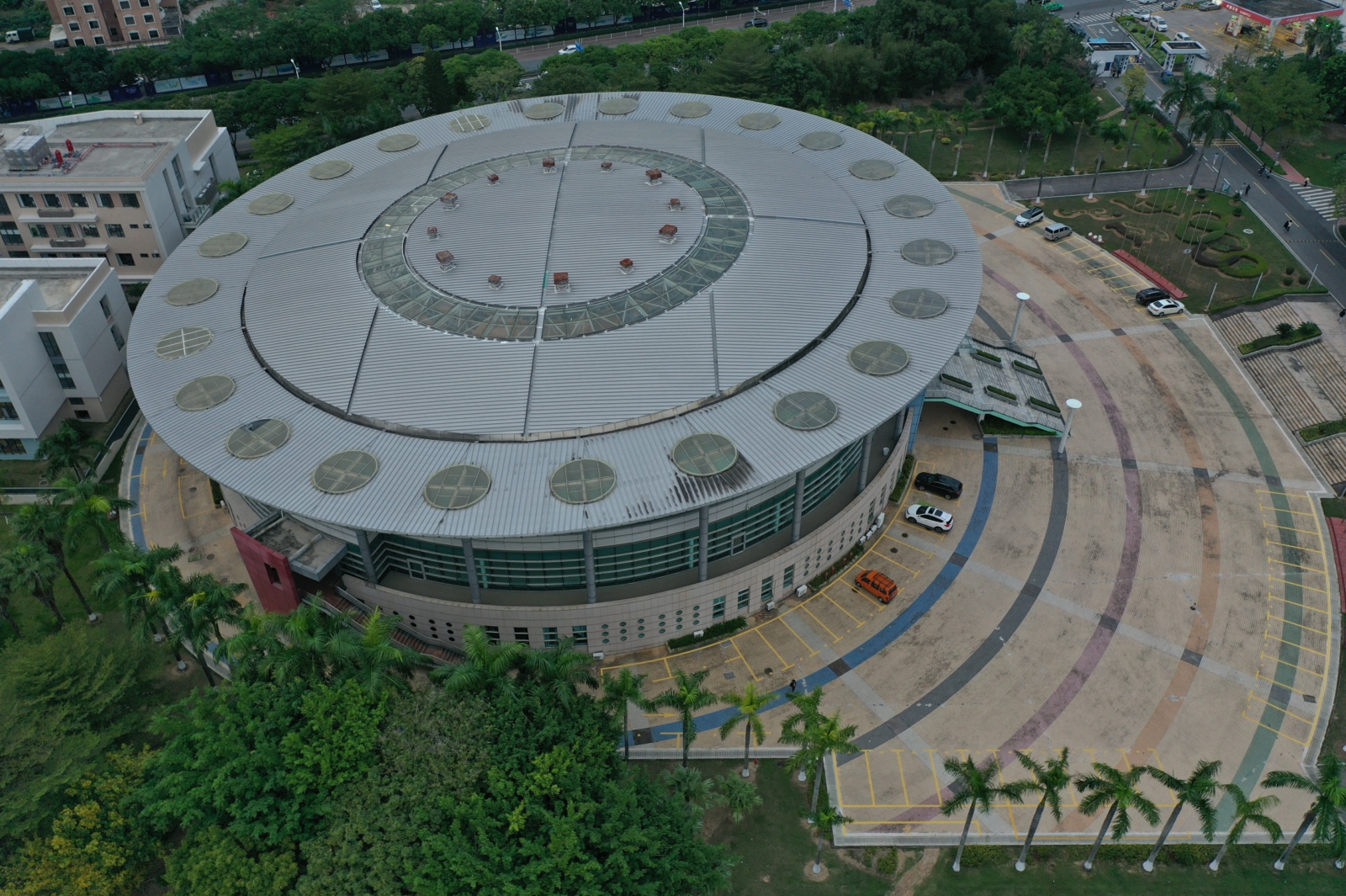}
        }
        \caption{The image provides $\imageembedding$}
        \label{fig:provide-imageembedding}
    \end{subfigure}
    \hspace{0.02\linewidth}
    \raisebox{8.5\height}{$\xrightarrow[]{\imageembedding{}}$}
    \hspace{0.02\linewidth}
    \begin{subfigure}{0.3\linewidth}
        \raisebox{0.2cm}{\spyimgcrop{\linewidth}{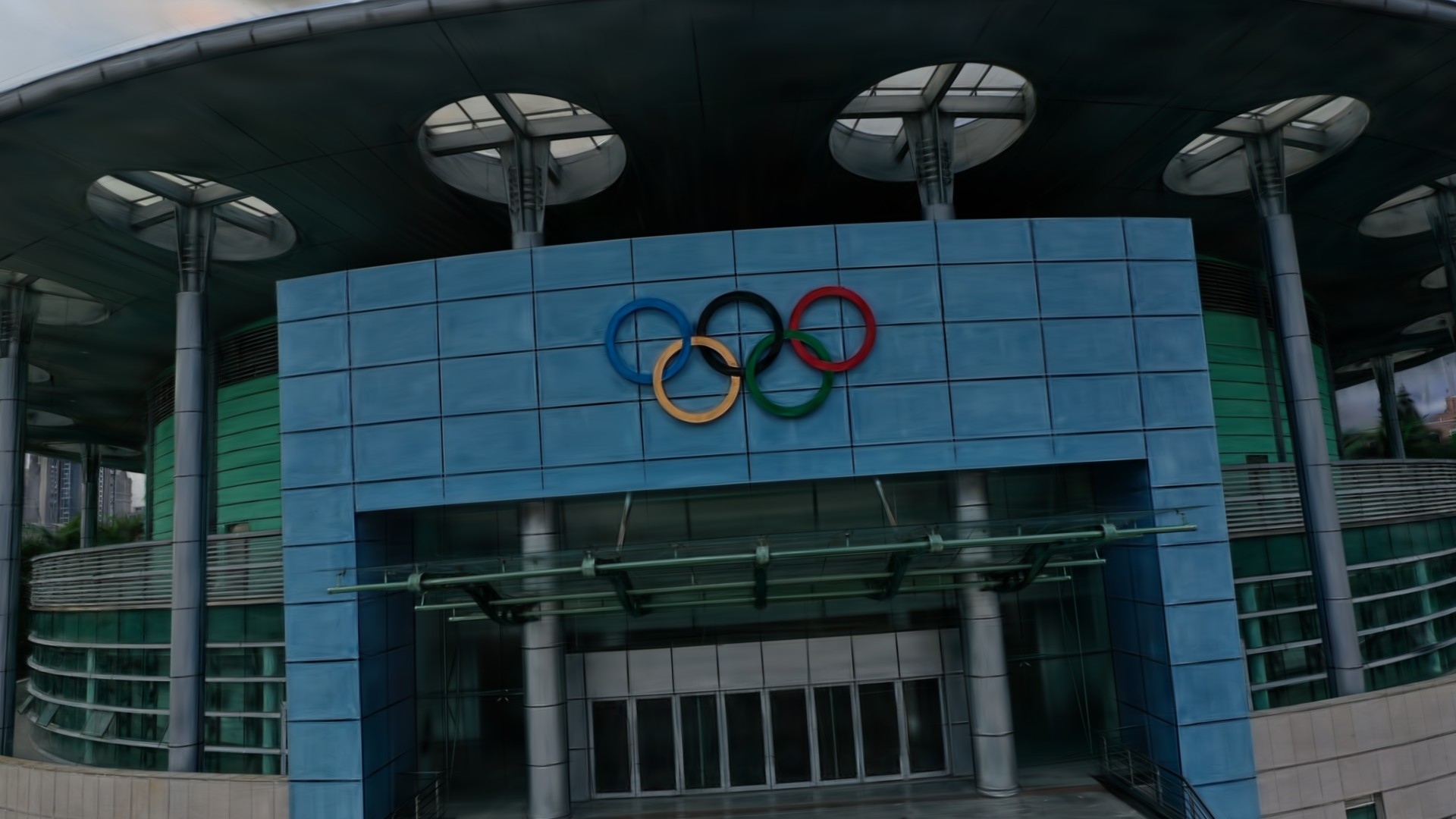}{1.9,1.85}{4.5,0.8}}
        \includegraphics[width=\linewidth]{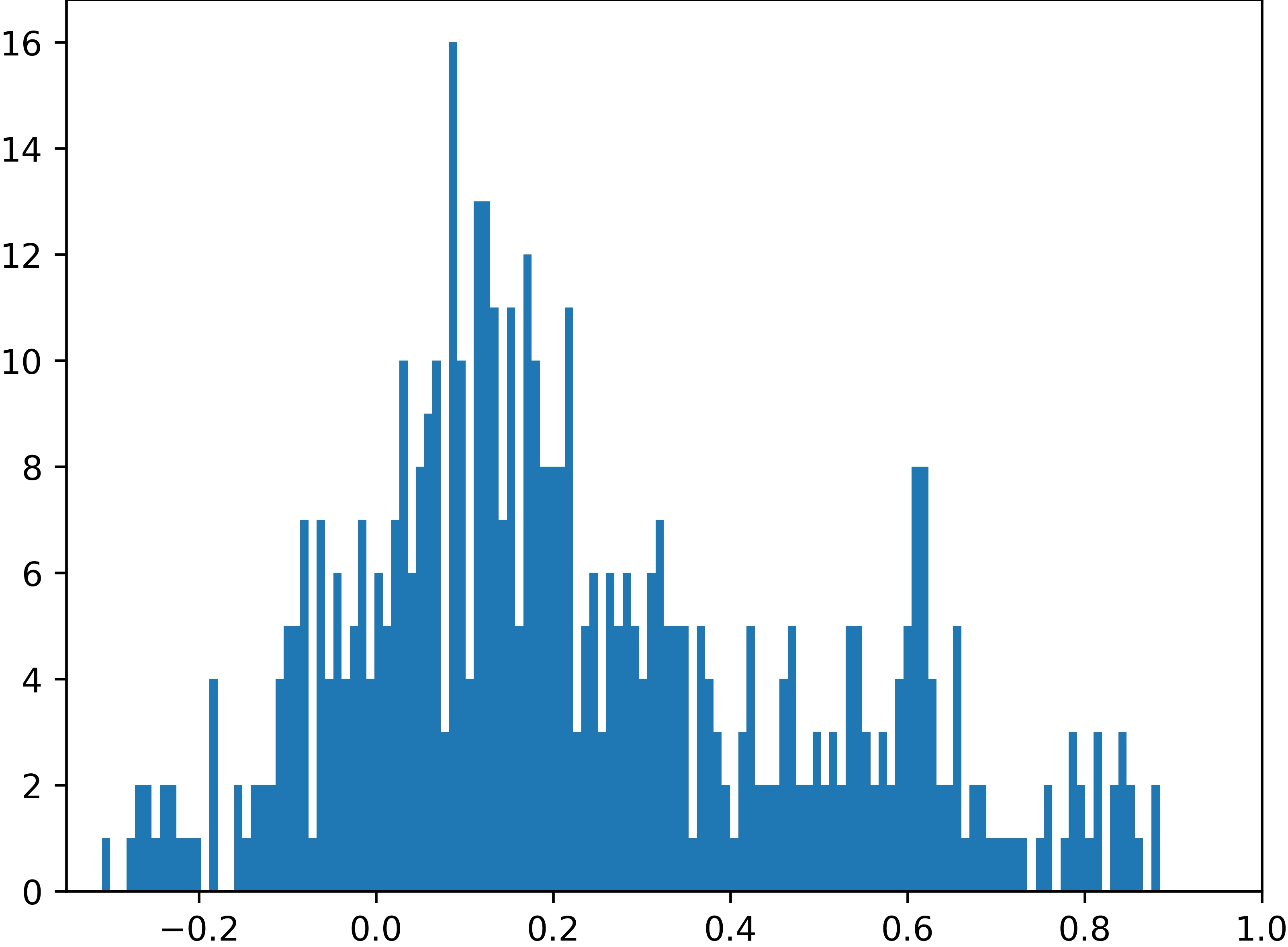}
        \caption{w/o similarity reg.}
        \label{fig:wo-simreg}
    \end{subfigure}
    \begin{subfigure}{0.3\linewidth}
        \raisebox{0.2cm}{\spyimgcrop{\linewidth}{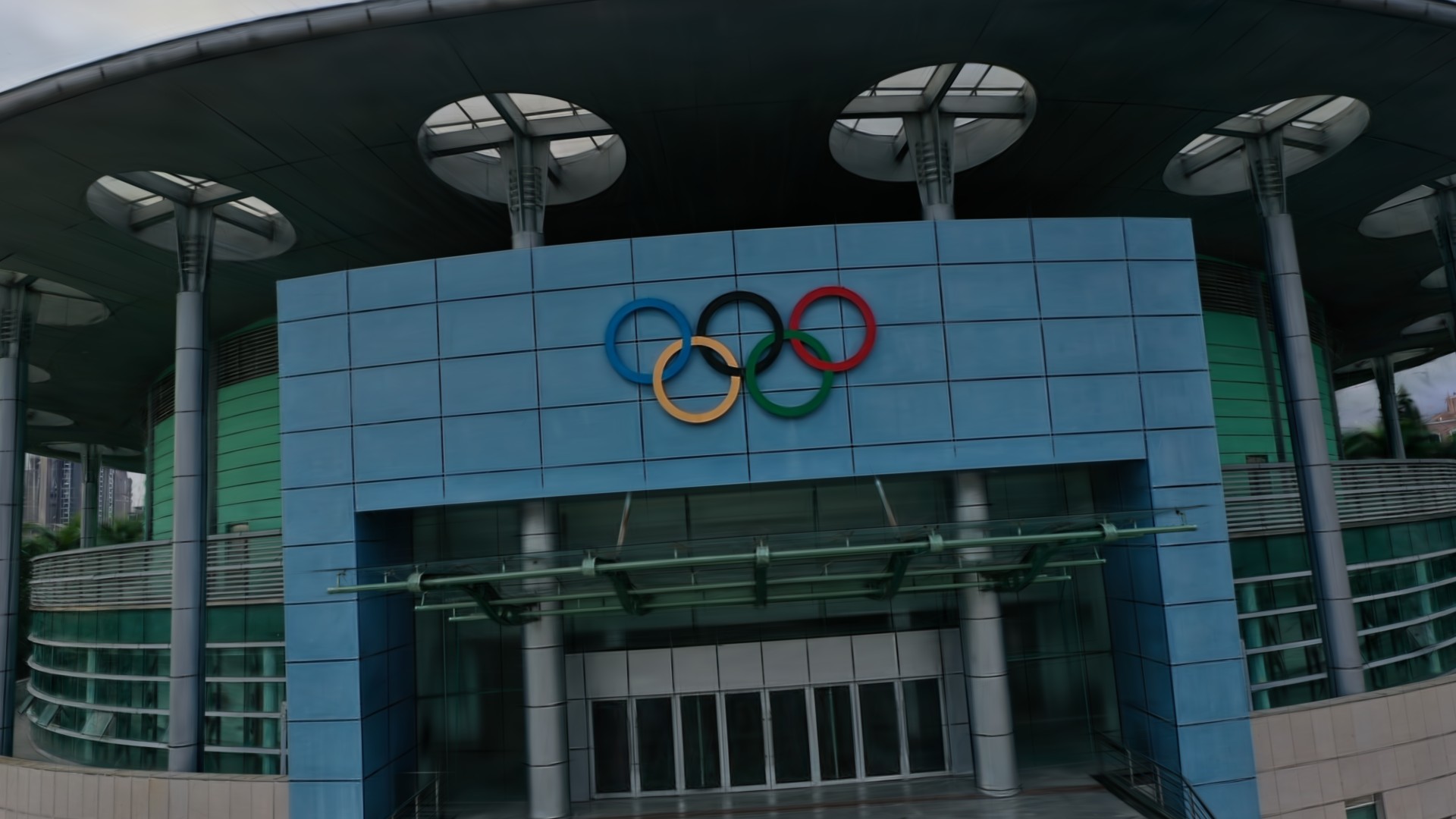}{1.9,1.85}{4.5,0.8}}
        \includegraphics[width=\linewidth]{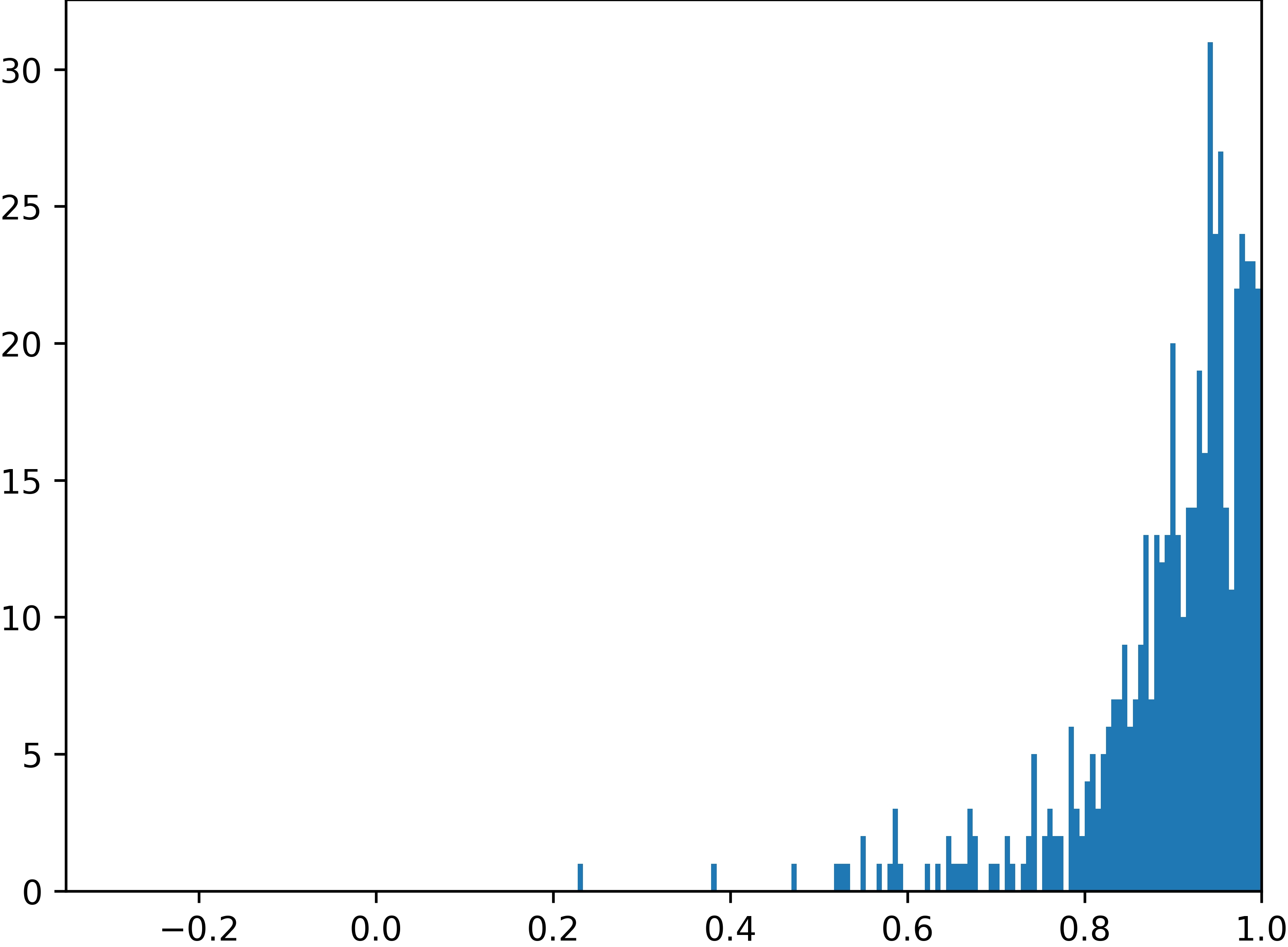}
        \caption{full}
        \label{fig:with-simreg}
    \end{subfigure}
    \caption{\tabletitle{A visual comparison of results with and without similarity regularization.}}
    \label{fig:sim-reg-vis}
    \end{center}
\end{figure*}

The second row of \Cref{fig:sim-reg-vis} presents a statistical analysis of the similarity among Gaussians within a small local region, where 513 Gaussians are selected, and the similarities between 512 of them and a central reference Gaussian are computed to generate a histogram. In the absence of similarity regularization, most Gaussians exhibit low similarity, clustering around 0.1. Such low similarity results in significant differences in appearance transformations among Gaussians, leading to visible artifacts. In contrast, with similarity regularization applied, the similarity values among Gaussians predominantly exceed 0.8. This high degree of similarity ensures more consistent appearance adjustments across Gaussians, effectively preventing the emergence of artifacts.

\end{document}